\documentclass[]{fairmeta}

\usepackage[utf8]{inputenc}
\usepackage{amsmath}
\usepackage{amssymb}
\usepackage{soul}
\usepackage{pifont}
\usepackage{fontawesome5}
\usepackage{array}
\usepackage{colortbl}
\usepackage{longtable}
\usepackage{float}
\usepackage{wrapfig}

\definecolor{linkblue}{HTML}{1A5FB4}
\hypersetup{colorlinks, linkcolor=linkblue, citecolor=linkblue, urlcolor=linkblue}

\newcommand{\featureyes}{{\color{red}\ding{52}}}
\newcommand{\featureno}{{\color{black}\ding{56}}}

\definecolor{calloutfill}{HTML}{DCE7F4}
\definecolor{calloutlight}{HTML}{F6F8FB}
\definecolor{calloutframe}{HTML}{6C829D}
\tcbset{callout/.style={enhanced,colframe=calloutframe,boxrule=0.8pt,arc=4pt,left=5pt,right=5pt,top=3pt,bottom=3pt}}
\newtcolorbox{findingbox}{callout,colback=calloutfill,before skip=6pt,after skip=6pt}
\newcounter{finding}

\title{ARE-bench: Benchmarking Agents' Research Ability in Real-World Researcher Sessions and Workspaces}

\affiliation[]{Shanghai Artificial Intelligence Laboratory, Fudan University, 
East China Normal University, Shanghai Jiao Tong University, Renmin University of China, Peking University}
\contribution[]{\large \textbf{Core Contributor}:  Jiahang Lin$^{\dagger}$, Wanying Ren$^{\dagger}$, Wanghan Xu$^{\dagger}$, Huayu Sha, Yang Shu, Zhonghang Lu, Chengjun Pan, Yi Dong \\[6pt]
\textbf{Contributor}: Shichun Liu, Yuyang Hu, Shihan Dou, Honglin Guo, Huijie Lv , Pujun Zheng, Zhe Sun, Yuankai Ying, Yiding Wang,  Jiahao Chen, Zhixiong Yang\\[6pt]
\textbf{Advisor}: Dingbo Yuan, Yicheng Zou, Zhecan Wang, Guoxiu He, Xiaoyu Xing ,Qi Zhang$^{\text{\faEnvelope}}$, Tao Gui$^{\text{\faEnvelope}}$, Xuanjing Huang

\vspace{3pt}
{\small
\textsuperscript{$\dagger$} Project leaders.
\quad
\faEnvelope\ Corresponding authors.
\par}
\vspace{6pt}

}

\abstract{
Research agents achieve increasingly strong benchmark scores,
yet these scores do not necessarily translate into reliable delivery in ongoing research projects.
Many research-agent benchmarks emphasize standalone tasks and relatively self-contained workflows in clean or curated workspaces. However, researchers ask agents to complete one step within an ongoing project, where requests depend on earlier session decisions and workspaces contain unfinished files, outdated versions, and failed attempts. We introduce ARE-bench to evaluate whether research agents advance ongoing projects while respecting prior decisions and existing workspace state, grounded in \textbf{real session context} and \textbf{in-progress workspaces}. ARE-bench includes 75 tasks built on agent sessions and full workspaces from recruited researchers' own projects, spanning seven research stages, 33 requirements, and 150 sub-requirements, with 1,173 binary checklist items in total.
Each task is anchored at a real user request, preserves its requirements, and summarizes the earlier requests still in effect as background.
The complete workspace is replayed to that boundary without cleanup, preserving artifacts from earlier turns.
Each checklist is compiled from an intent-aligned rubric and credits any valid solution.
Across seven frontier models, the strongest passes every checklist item on only 66.2\% of tasks, and Design and Execution rank among the hardest stages for every model.
Three models pass about 69\% of tasks when only delivery contracts and general norms are checked, but just 28.4\% to 40.0\% once historical and material or workspace requirements are also checked.
Evaluating delivery without these contextual requirements can therefore overestimate research agents and mask differences between them.
}

\date{\today}
\metadata[\faGlobe\ Project Page]{\url{https://arebench.com}}

\begin{document}

\maketitle

\begin{figure}[!h]
\centering
\includegraphics[width=\textwidth]{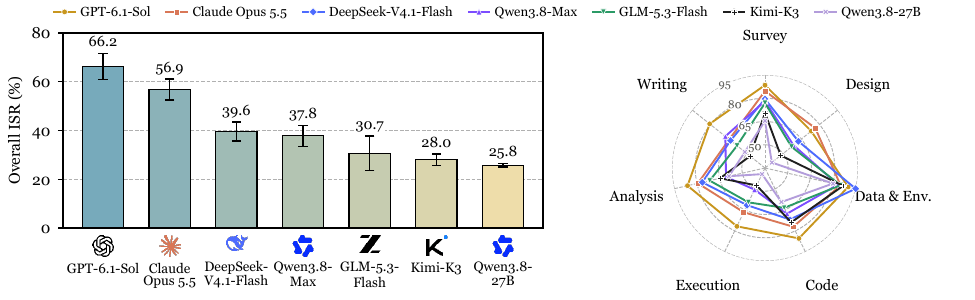}
\caption{ISR of seven models on 75 tasks, averaged over three rounds. Left: overall ISR, with error bars for the sample standard deviation across rounds. Right: ISR by research stage, with the radial axis from 35\% to 95\%.}
\label{fig:teaser}
\end{figure}

\section{Introduction}

\begin{figure}[t]
\centering
\includegraphics[width=\textwidth]{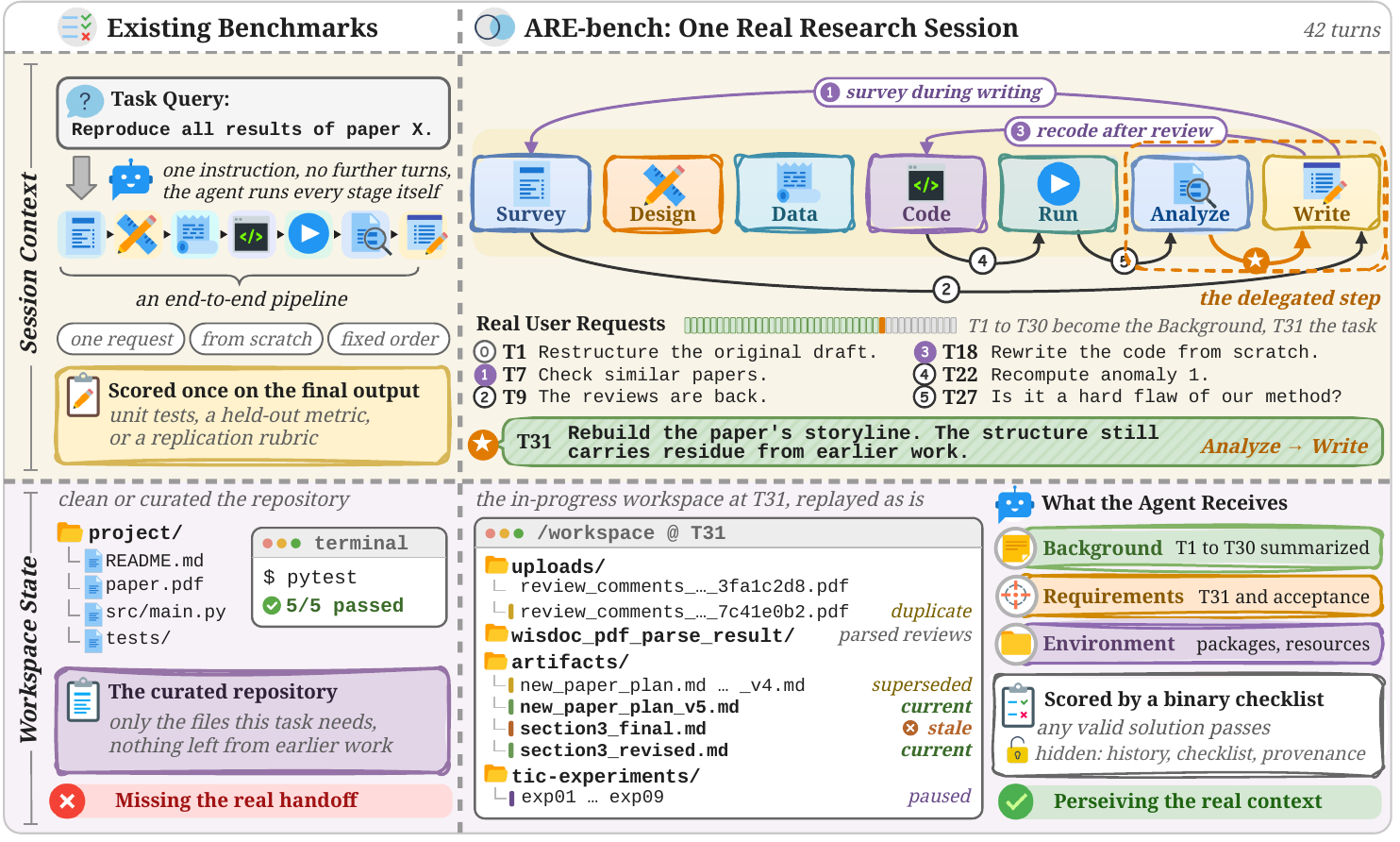}
\caption{An illustrative contrast between a standalone research task and one real research session in ARE-bench. Top: the standalone task has an end-to-end objective, while the real session moves back and forth between research stages. The task is cut at request T31, with still-effective requests from T1 to T30 summarized in Background. Bottom: a clean or curated starting repository is contrasted with the replayed in-progress workspace, including superseded drafts, misleadingly named files, and paused runs.}
\label{fig:compare}
\end{figure}

High research-agent benchmark scores can give an incomplete picture of reliable delivery in real research projects.
For example, Qwen reports a score of 93.0\% for Qwen3.8-Max on PaperBench Code-Dev \citep{qwen-2026-qwen38-model-card}.
This variant assesses code development without checking execution or whether the paper's empirical results are reproduced \citep{starace-etal-2025-paperbench}.
Even a near-ceiling score on this assessment therefore leaves important aspects of research delivery untested.

Many established research-agent benchmarks emphasize standalone objectives, such as Kaggle competitions \citep{chan-etal-2024-mlebench}, research engineering \citep{wijk-etal-2024-rebench}, or full paper replication \citep{starace-etal-2025-paperbench}.
In ongoing projects, researchers also ask agents to handle an individual request, such as repairing a script after a failed run or adding a table to a paper. Success then depends on both the requested change and the prior session decisions and workspace artifacts that the change must respect.
Standalone objectives and curated starting states can leave this ability to advance an existing project underexamined, even when they rigorously assess their intended outcomes.
We therefore evaluate whether research agents advance ongoing projects while respecting prior decisions and existing workspace state, along two dimensions, \textbf{real session context} and \textbf{in-progress workspaces}.

First, \textbf{real session context} supports evaluating whether agents carry earlier decisions into the current delivery.
Real research sessions switch between task types \citep{slinko-etal-2026-taskflows}, returning to literature review after experiments or adding experiments during paper writing (Figure~\ref{fig:compare}, top).
A request to fix a failed experiment may still require preserving methodological choices established during an earlier design discussion. Summarizing still-effective earlier requests as Background makes these constraints available for evaluating whether the agent integrates them with the current request.

Second, \textbf{in-progress workspaces} support evaluating whether agents change existing artifacts while preserving their scientific meaning.
Clean or curated starting states simplify this setting, whereas ongoing projects can contain temporary files, outdated versions, and failed attempts that require checking which artifacts and conventions remain applicable (Figure~\ref{fig:compare}, bottom).
For example, in our CylindFormer repair task, two executions both resume training with nonzero loss and parameter updates, yet pass 20/20 and 16/20 checklist items, respectively. The latter changes coordinate handling, overlap labels, and matching losses required by the reference paper in the workspace (Figure~\ref{fig:repair-divergence}). Restoring execution alone therefore does not establish that a repair preserves the research method.

Recent benchmarks also incorporate real developer sessions \citep{lv-etal-2026-realclawbench,jha-etal-2026-reap} or fine-grained research tasks \citep{wang-etal-2026-aarribench}. Our focus is on jointly evaluating the still-effective requirements of a real research request and the in-progress workspace in which that request arose.

We introduce ARE-bench to evaluate this ability.
ARE-bench includes 75 tasks spanning seven research stages (Survey, Design, Data \& Env., Code, Execution, Analysis, and Writing), 33 requirements, and 150 sub-requirements, with 1,173 binary checklist items in total.
We collect agent sessions and full workspaces from recruited researchers' own projects.
For session context, we select a real user request as the task boundary, retain its requirements, and use the Background section to summarize the earlier requests still in effect, covering 10.3 earlier requests on average.
For in-progress workspaces, we replay the complete workspace to that boundary without cleaning, preserving 41.1 files per task on average, with agent-written files from earlier turns in 65.3\% of tasks. The actor receives only the instruction and workspace, and an intent-aligned rubric compiles into the checklist that credits any valid solution.

We evaluate seven frontier models with three rollouts per task by instance success rate (ISR), the fraction of tasks whose checklist items all pass.
GPT-6.1-Sol and Claude Opus 5.5 reach 66.2\% and 56.9\% ISR, while the other five models stay below 40\% and Qwen3.8-27B reaches only 25.8\% (Figure~\ref{fig:teaser}, left).
Checklist item success rate (CSR), which gives partial credit, spans only 13.1 points across models, whereas ISR spans 40.4 points, so models satisfy most items but differ in meeting all of them.
For the same actor deliveries, DeepSeek-V4.1-Flash, GLM-5.3-Flash, and Kimi-K3 each reach about 69\% ISR when only delivery contracts and general norms are checked. Including historical and material or workspace requirements lowers their ISR to 40.0\%, 31.6\%, and 28.4\%. Under the full checklist, the gap between DeepSeek-V4.1-Flash and Kimi-K3 grows from 0.44 to 11.56 points, showing that contextual requirements reveal delivery gaps masked by narrower checks.
For session context, real sessions move between research stages, but stage ability is uneven across models. Kimi-K3 ranks third on Code but last on Writing, Qwen3.8-Max ranks fourth overall but second on Writing, and Claude Opus 5.5 and DeepSeek-V4.1-Flash lead Design and Data \& Env., so an evaluation covering a single stage would rank models differently.
Design and Execution rank among the three hardest stages for every model, where the best and worst models differ by 35.8 and 37.4 points, compared with 15.0 points on the easiest stage, Data \& Env. (Figure~\ref{fig:teaser}, right).
Claude Opus 5.5 also costs 2.6 times as much per run as GPT-6.1-Sol yet trails it by 9.3 points in ISR.

Our contributions are:

\begin{itemize}
    \item We introduce ARE-bench to evaluate whether research agents advance ongoing projects while respecting prior decisions and existing workspace state, with 75 tasks drawn from real researcher sessions and paired with replayed in-progress workspaces.
    \item We map each checklist item to one of 207 sub-requirements across seven stages, covering 150 of them and all 33 requirements, and aggregate these labels into a requirement profile for each task.
    \item We evaluate seven frontier models and find that the strongest reaches only 66.2\% ISR, that requirements carried over from the session and workspace expose failures and model differences hidden by checks limited to delivery contracts and general norms, and that stage rankings diverge from overall rankings.
\end{itemize}

\FloatBarrier
\section{Related Work}

\paragraph{Research-agent benchmarks.}
Research-agent benchmarks cover scientific work at different scopes, with success defined by the outcomes each task requires. For specialized research activities, RE-Bench and MLE-bench assess research-engineering and competition performance \citep{wijk-etal-2024-rebench,chan-etal-2024-mlebench}; ScienceAgentBench and SWE-bench Science evaluate scientific programming and software tasks \citep{chen-etal-2024-scienceagentbench,xu-etal-2026-swebscience}, while SUPER and ResearchEnvBench target repository execution and runtime preparation \citep{bogin-etal-2024-super,wang-etal-2026-researchenvbench}. AARRI-Bench evaluates granular research scenarios involving drafts, logs, and other research artifacts \citep{wang-etal-2026-aarribench}. These settings tie acceptance to a specified engineering or research deliverable. Broader coverage includes AstaBench's heterogeneous scientific suite, including literature-query tasks drawn from user logs, with member-specific metrics \citep{bragg-etal-2025-astabench}, alongside benchmarks organized around reproduction or experimental outcomes. PaperBench and CORE-Bench assess paper replication and computational reproduction \citep{starace-etal-2025-paperbench,siegel-etal-2024-corebench}; ResearchGym evaluates research outcomes against baselines, while EXP-Bench assesses experiment execution and conclusions \citep{garikaparthi-etal-2026-researchgym,kon-etal-2025-expbench}. For scientific-output evaluation, ResearchClawBench anchors rubric-based judgments to a hidden reference study \citep{xu-etal-2026-researchclawbench}. Together, these benchmarks move from specialized research activities to broader scientific workflows while retaining task-specific acceptance targets. ARE-bench takes the current research delegation as its acceptance boundary, compiling requested deliverables and applicable inherited scientific constraints into a checklist. Compared with full-experiment or reference-study targets, this scope supports diagnosing unmet requirements and accepting alternative valid implementations without requiring completion of the surrounding study.

\paragraph{Stateful and interaction-derived benchmarks.}
An evaluation's starting point determines which prior work an agent must continue. Existing benchmarks provide several forms of task state: some Terminal-Bench tasks start with task-specific repository changes \citep{merrill-etal-2026-terminalbench}; OSWorld configures application and file states, while $\tau$-bench evaluates tool-mediated conversations against database goals \citep{xie-etal-2024-osworld,yao-etal-2024-taubench}. LongDS-Bench accumulates analytical state across notebook-derived dialogues \citep{xu-etal-2026-longdsbench}. For usage-derived tasks, the additional question is how a recorded request and its surrounding work become the evaluation input. SWE-bench pairs issue reports with repository snapshots \citep{jimenez-etal-2023-swebench}. At session entry, K-Bench samples real first-turn scientific requests, while SWE-Together starts from a coding session's initial request \citep{brueckner-etal-2026-kbench,wu-etal-2026-swetogether}. Other benchmarks reconstruct the starting point: RealClawBench rewrites requests into standalone instructions with sanitized standalone workspaces \citep{lv-etal-2026-realclawbench}, and REAP backs out accepted solution changes from a current monorepo \citep{jha-etal-2026-reap}. DuMateBench most directly addresses historical continuation by selecting a local objective, retaining visible history, and restoring recoverable pre-cutoff files; it combines deterministic checks with artifact rubrics \citep{niu-etal-2026-dumatebench}. ARE-bench focuses on subsequent requests in real research projects, pairing recovered in-progress workspaces with Background summaries of still-valid prior context, current Requirements, and Environment. Compared with DuMateBench's general-workflow setting, its research-specific requirements and mapping of checklist items to experiment execution, result analysis, and writing support diagnosis of unmet work by research activity.

\paragraph{Positioning.}
Table~\ref{tab:positioning} summarizes these comparisons. The table separates a genuine work request from its provenance in a recorded session, then distinguishes inherited workspace progress from the broader research-workflow scope. ARE-bench combines all four binary properties: each task is a real research request selected from a recorded session, starts from an in-progress workspace, and belongs to a benchmark covering the full research lifecycle. Its rubric scoring translates the active delegation into deliverable checks, linking completion to requested outcomes and locating unmet requirements by research activity.

\begin{table*}[t]
\centering
\scriptsize
\setlength{\tabcolsep}{3.5pt}
\resizebox{\textwidth}{!}{%
\begin{tabular}{lccccc}
\toprule
Benchmark & \textbf{Real request} & \textbf{Session-derived} & \textbf{In-progress workspace} & \textbf{Research workflow} & \textbf{Scoring} \\
\midrule
SWE-bench \citep{jimenez-etal-2023-swebench} & \featureyes & \featureno & \featureno & \featureno & Tests \\
Terminal-Bench (2.0) \citep{merrill-etal-2026-terminalbench} & \featureyes & \featureno & \featureyes & \featureno & Varies \\
OSWorld \citep{xie-etal-2024-osworld} & \featureyes & \featureno & \featureyes & \featureno & Varies \\
$\tau$-bench \citep{yao-etal-2024-taubench} & \featureyes & \featureno & \featureyes & \featureno & Varies \\
MLE-bench \citep{chan-etal-2024-mlebench} & \featureno & \featureno & \featureno & \featureno & Task metric \\
ScienceAgentBench \citep{chen-etal-2024-scienceagentbench} & \featureno & \featureno & \featureyes & \featureno & Varies \\
PaperBench \citep{starace-etal-2025-paperbench} & \featureno & \featureno & \featureno & \featureno & Rubric \\
RE-Bench \citep{wijk-etal-2024-rebench} & \featureno & \featureno & \featureno & \featureno & Task metric \\
CORE-Bench \citep{siegel-etal-2024-corebench} & \featureno & \featureno & \featureyes & \featureno & Answer check \\
AARRI-Bench \citep{wang-etal-2026-aarribench} & \featureyes & \featureno & \featureyes & \featureyes & Varies \\
ResearchGym \citep{garikaparthi-etal-2026-researchgym} & \featureno & \featureno & \featureno & \featureno & Task metric \\
EXP-Bench \citep{kon-etal-2025-expbench} & \featureno & \featureno & \featureno & \featureno & Rubric + Answer check + Tests \\
AstaBench \citep{bragg-etal-2025-astabench} & \featureyes & \featureyes & \featureno & \featureyes & Varies \\
SWE-bench Science \citep{xu-etal-2026-swebscience} & \featureyes & \featureno & \featureno & \featureno & Tests \\
SUPER \citep{bogin-etal-2024-super} & \featureno & \featureno & \featureyes & \featureno & Varies \\
ResearchEnvBench \citep{wang-etal-2026-researchenvbench} & \featureno & \featureno & \featureno & \featureno & Tests \\
ResearchClawBench \citep{xu-etal-2026-researchclawbench} & \featureno & \featureno & \featureyes & \featureyes & Rubric \\
K-Bench \citep{brueckner-etal-2026-kbench} & \featureyes & \featureyes & \featureyes & \featureno & Rubric \\
DuMateBench \citep{niu-etal-2026-dumatebench} & \featureyes & \featureyes & \featureyes & \featureno & Varies \\
SWE-Together \citep{wu-etal-2026-swetogether} & \featureyes & \featureyes & \featureno & \featureno & Rubric \\
RealClawBench \citep{lv-etal-2026-realclawbench} & \featureyes & \featureyes & \featureno & \featureno & Tests \\
REAP/Harvest \citep{jha-etal-2026-reap} & \featureyes & \featureyes & \featureno & \featureno & Tests \\
LongDS-Bench \citep{xu-etal-2026-longdsbench} & \featureno & \featureno & \featureno & \featureno & Answer check \\
\midrule
\textbf{ARE-bench} & \textbf{\featureyes} & \textbf{\featureyes} & \textbf{\featureyes} & \textbf{\featureyes} & \textbf{Rubric} \\
\bottomrule
\end{tabular}
}
\caption{Comparison of benchmark designs for research-project continuation. \textbf{Real request} denotes a documented genuine work need, regardless of source. \textbf{Session-derived} denotes a task or request whose provenance is an actual recorded work session; earlier conversation transfer is optional. \textbf{In-progress workspace} denotes task-relevant files or application state containing prior activity or intermediate progress available when evaluation starts; a base repository, raw data, dependency environment, or empty scaffold alone does not qualify. \textbf{Research workflow} denotes suite-level coverage of Survey, Design, Data \& Environment, Code, Execution, Analysis, and Writing. \textbf{Scoring} names the primary acceptance basis: Tests, State check, Answer check, Task metric, Rubric, a stable combination written with `+`, or Varies when mechanisms differ across tasks or sub-benchmarks. A tick marks a documented feature and a cross means it is not established under these definitions.}
\label{tab:positioning}
\end{table*}

\section{ARE-bench}

\subsection{Task Formulation}

Each task is anchored at a user request in an ongoing research session. At cut point $c$, the raw task is
\begin{equation}
    \tau_c=(q_c,H_{<c},W_c),
\end{equation}
where $q_c$ is the current request, $H_{<c}$ is the history of earlier user requests, and $W_c$ is the workspace immediately before the agent responds. Construction maps $\tau_c$ to the released task
\begin{equation}
    b_c=(I_c,W_c,R_c),
\end{equation}
where $I_c$ is the instruction and $R_c$ is the rubric and provenance added during construction. The benchmark is $B=\{b_c\}$.

We define the instruction as three sections,
\begin{equation}
    I_c=\bigl(I_c^{\mathrm{bg}},I_c^{\mathrm{req}},I_c^{\mathrm{env}}\bigr),
\end{equation}
with Background derived from the request history as
\begin{equation}
    I_c^{\mathrm{bg}}=\operatorname{Compress}(H_{<c}),
\end{equation}
where $\operatorname{Compress}$ retains context relevant to $q_c$ that remains valid at cut point $c$ and rewrites it as a state summary. Requirements $I_c^{\mathrm{req}}$ preserves $q_c$ and specifies delivery paths and acceptance conditions. Environment $I_c^{\mathrm{env}}$ describes available resources and execution limits.

The actor receives only $I_c$ and $W_c$ and produces a delivery $y_c$. The rubric compiles into binary checklist items aligned with user intent. Let $K_c$ be the set of verifiable checklist items, excluding non-triggered conditional ones, and $r_{c,k}\in\{0,1\}$ indicate whether checklist item $k$ is satisfied.

\subsection{Benchmark Construction}

\begin{figure}[!htb]
\centering
\includegraphics[width=0.90\textwidth]{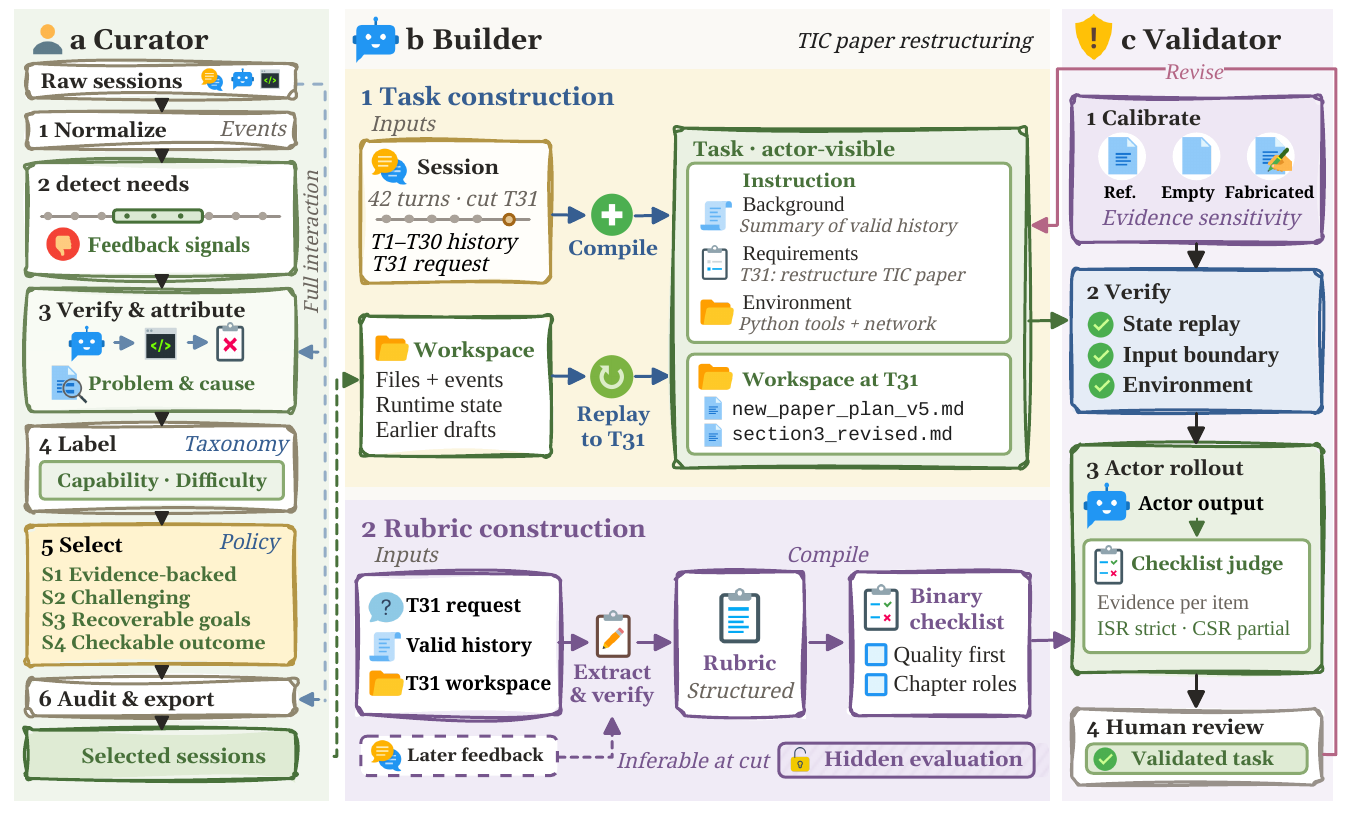}
\caption{Overview of ARE-bench construction, illustrated by a paper restructuring task cut at request T31 of a 42-turn session. (a) The Curator normalizes raw researcher sessions, verifies requirements against the full interaction, labels them with the requirement taxonomy, and selects sessions by rules S1 to S4. (b) The Builder compiles the target request and still-valid history into the actor-visible instruction, replays the workspace to that request, and compiles a hidden rubric into a binary checklist. (c) The Validator calibrates the judge, verifies the task package, runs actors, and sends each task to human review, which returns failing tasks to the Builder.}
\label{fig:overview}
\end{figure}

Figure~\ref{fig:overview} shows the three construction roles. The Curator selects researcher sessions, the Builder turns a request in a selected session into a task, and the Validator tests each task before release.

\subsubsection{Curator}

The Curator filters raw researcher sessions in six steps. It normalizes each session into ordered turns and events that keep their source positions. Reading user turns alone, it splits the session into episodes, recovers goals and requirements with their revisions, and flags feedback signals such as rejections or requests to redo work. It then verifies each requirement against the full interaction, including agent replies, tool results, and produced files, and attributes each confirmed problem to its cause. Verified requirements are labeled with the research requirement taxonomy of Appendix~\ref{app:requirement-taxonomy} and rated for difficulty, and each label cites quoted evidence. After schema and evidence prechecks, an episode is selected only if it satisfies four rules:

\begin{enumerate}
    \item \textbf{S1 Evidence-backed}: at least one primary requirement label is supported by evidence.
    \item \textbf{S2 Challenging}: the task has medium or high difficulty.
    \item \textbf{S3 Recoverable goals}: the goals and requirements can be recovered coherently from the session.
    \item \textbf{S4 Checkable outcome}: completion criteria can be checked from the record, whether or not the original agent succeeded.
\end{enumerate}

Episodes with an unresolved rule are held for review, and other failing episodes are rejected. A final audit checks every quote against the source session and exports each session that contains a selected episode.

\subsubsection{Builder}

\paragraph{Task construction.}
The Builder picks a target request $q_c$ at a turn where a core requirement of the selected episode is carried out. Background summarizes earlier requests still in effect as the state at cut point $c$. It takes facts from earlier agent turns but omits their unexecuted plans, which would reveal solutions. Requirements retains $q_c$ with its delivery paths and explicit acceptance conditions, and Environment states the available tools and network access. The workspace is reconstructed by replaying the platform event stream $\mathcal{A}$ up to the cut point,
\begin{equation}
    W_c=\operatorname{Replay}(\mathcal{A},c),
\end{equation}
and cross-checked against the transcript and the final disk state. A cut point is discarded if it falls inside a file write, if replay disagrees with the transcript, or if a required dependency cannot be recovered. The environment contains the platform baseline and the packages available at the cut point, and it keeps network access when the source environment had it.

\paragraph{Rubric construction.}
The rubric follows the user's intended outcome rather than the original agent's output. The Builder extracts intent from $q_c$, the still-valid history, and the workspace at the cut point. Later user feedback often reveals requirements the user held without stating, so the Builder uses it to sharpen such requirements and keeps an item only if a competent researcher could infer it at the cut point. Requirements raised only in later turns and personal preferences are excluded. Each item checks one deliverable or acceptance condition with a binary decision rule and an evidence requirement. Items do not prescribe implementation paths, so any valid alternative passes. Items stated by the user appear in Requirements, while items a competent researcher would infer stay out of the instruction so that it does not become an answer checklist. Each item also receives one sub-requirement label from the taxonomy, and a task's requirement profile is the union of the parent requirements of its items. The rubric and a provenance package holding raw wording, source turns, path mappings, replay logs, and anonymization decisions form the hidden evaluation set. Appendix~\ref{app:training-repair-case} walks through a complete task.

\subsubsection{Validator}

The Validator tests each task in four steps and returns failing tasks to the Builder for revision. Calibration scores three deliveries with the checklist judge. The original agent's output, when available, tests whether each verdict cites evidence present in the delivery. It calibrates the judge rather than serving as an answer key, because later turns often show that it fell short. An empty delivery should score near zero, and a fabricated delivery that looks complete but contains false content should score clearly below actor runs. Verification then confirms three properties. The replayed state matches the audited state for every file the task uses, the instruction and workspace contain no information from after the cut point, and the released environment installs and imports the required dependencies. In the actor rollout, an agent runs the task and the judge scores each item with cited evidence, yielding ISR and CSR. When every item passes, the Builder rechecks rubric coverage and strictness but does not add uninferable items to lower the score. In human review, the task author checks every failed item. A task proceeds to peer review only when each failure is a valid item that the actor genuinely missed. Otherwise the instruction or rubric is revised and the task is rerun.

\subsection{Benchmark Statistics}

Table~\ref{tab:stats} summarizes ARE-bench. Each of the 75 tasks compiles into 15.6 checklist items on average. Three properties distinguish the task distribution. First, every task starts from an in-progress workspace: the replayed workspace holds 41.1 files on average, 65.3\% of tasks contain agent-written files from earlier turns, and the Background section compresses 36.0 earlier user turns into a state summary that cites on average 10.3 earlier requests still in effect. Second, each task is a single local step with 2.5 deliverables on average. Third, the checklist items span all seven research stages.

These tasks contain 1,173 checklist items covering 150 sub-requirements and all 33 requirements. Table~\ref{tab:requirement-coverage} summarizes this coverage across the seven research stages, and Table~\ref{tab:requirement-mapping} in Appendix~\ref{app:requirement-taxonomy} presents the corresponding requirement hierarchy. A task can contribute to multiple stages, so the task counts across rows overlap.

\begin{table}[t]
\centering
\small
\begin{tabular}{lrrrr}
\toprule
Research stage & Requirements & Sub-requirements & Checklist items & Tasks \\
\midrule
Survey & 2 & 20 & 111 & 26 \\
Design & 3 & 25 & 268 & 41 \\
Data \& Env. & 5 & 16 & 45 & 20 \\
Code & 5 & 23 & 217 & 32 \\
Execution & 7 & 26 & 173 & 33 \\
Analysis & 4 & 13 & 92 & 34 \\
Writing & 7 & 27 & 267 & 31 \\
\midrule
Total & 33 & 150 & 1,173 & 75 \\
\bottomrule
\end{tabular}
\caption{Observed requirement coverage across the 75 tasks. Tasks count distinct tasks within each stage, and the total counts their union.}
\label{tab:requirement-coverage}
\end{table}

\subsection{Evaluation Protocol}
\label{sec:evaluation-protocol}

For each task, an actor receives the instruction and the restored workspace, works under the declared tool and time limits, and submits a delivery $y_c$ containing the artifacts listed in the instruction. The checklist evaluator reads $y_c$ and the minimum input source needed to verify a claimed change. It then assigns each $r_{c,k}$ using the checklist item's evidence rule. Missing artifacts, unresolvable claims, and unsupported conclusions receive zero for the affected checklist item.

The evaluation preserves both the request context and the project's in-progress state. Background carries forward earlier user requests that remain in effect, Requirements specifies the current deliverables and acceptance conditions, and Environment states the available resources and limits. The restored workspace retains unfinished artifacts, outdated versions, and prior failed attempts, requiring the actor to interpret the current request against existing project materials. Checklist items and provenance remain hidden from the actor and support scoring and audit.

Each model runs three rollouts per task, and we report the mean and sample standard deviation of each metric across the three rounds. Each run records its model identifier, actor harness, task resources, and time limit. A run that reaches its time limit is scored on the artifacts delivered by then.

\subsection{Metrics}

Following the checklist-based definitions of \citet{ding-etal-2026-octobench}, let $N=|B|$ and $n_c=|K_c|$, with $K_c$ and $r_{c,k}$ as defined above. The Instance Success Rate (ISR) is strict: a task is successful only when every verifiable checklist item passes.
\begin{equation}
    \operatorname{ISR}=\frac{1}{N}\sum_{c\in B}
    \mathbf{1}\left[\bigwedge_{k\in K_c}(r_{c,k}=1)\right].
\end{equation}
The Checklist Item Success Rate (CSR) gives partial credit at checklist-item level:
\begin{equation}
    \operatorname{CSR}=\frac{1}{N}\sum_{c\in B}
    \frac{1}{n_c}\sum_{k\in K_c}r_{c,k}.
\end{equation}
ISR captures holistic satisfaction under conjunctions of heterogeneous requirements, whereas CSR diagnoses which requirements are most often violated. We report peak context and estimated API cost per run separately. Tasks with no verifiable checklist items are invalid construction outputs and are reported as construction errors rather than assigned a score.

\section{Experiments}

\label{sec:experiments}

\definecolor{ExpScoreBlue}{HTML}{497F9B}

\definecolor{ExpGapAmber}{HTML}{C58B45}

\definecolor{ExpHeader}{HTML}{F0F4F7}

\newcommand{\ExpScoreCell}[2]{\cellcolor{ExpScoreBlue!#1!white}#2}

\newcommand{\ExpGapCell}[2]{\cellcolor{ExpGapAmber!#1!white}#2}

\par\WFclear

\FloatBarrier

\subsection{Experimental Setup}


We evaluate agents on continuing user requests in real research workspaces under the protocol in Section~\ref{sec:evaluation-protocol}. Actors receive the instruction and restored workspace, and deliveries are scored against task-specific binary checklists.


\textbf{Tasks and evaluated models.} The main evaluation comprises 75 tasks: 67 CPU tasks, 8 GPU tasks. Its 1,173 checklist items cover the seven research stages defined above: Survey, Design, Data \& Env., Code, Execution, Analysis, and Writing. We evaluate the seven models in Table~\ref{tab:experiment-models} using the actor harness recorded for each model in the archive.



\begin{table}[!htbp]
\centering\small
\setlength{\tabcolsep}{3pt}
\renewcommand{\arraystretch}{1.16}
\begin{tabular}{@{}>{\raggedright\arraybackslash}p{0.30\linewidth}>{\raggedright\arraybackslash}p{0.35\linewidth}>{\raggedright\arraybackslash}p{0.25\linewidth}@{}}
\toprule
\rowcolor{ExpHeader}
Evaluated model & Archived model identifier & Actor harness \\
\midrule
GPT-6.1-Sol & \texttt{gpt-6.1-sol} & Codex \\
Claude Opus 5.5 & \texttt{claude-opus-5-5} & Claude Code \\
DeepSeek-V4.1-Flash & \texttt{deepseek-v4.1-flash} & Claude Code \\
Qwen3.8-Max & \texttt{qwen3.8-max} & Claude Code \\
GLM-5.3-Flash & \texttt{glm-5.3-flash} & Claude Code \\
Kimi-K3 & \texttt{kimi-k3} & Claude Code \\
Qwen3.8-27B & \texttt{qwen3.8-27b} & Claude Code \\
\bottomrule
\end{tabular}
\caption{Evaluated models and actor harnesses. Model identifiers follow the main evaluation archive. The specific historical GLM deployment identifiers are retained in individual run records.}
\label{tab:experiment-models}
\end{table}


\textbf{Execution environments and compute resources.} CPU tasks run in E2B sandboxes or Docker containers, while GPU and research tasks use Docker. Table~\ref{tab:experiment-resources} reports the declared task resources and actor time limits recorded for each batch. Resources are configured per task to support the existing workspace and its dependencies. GPU and research batches use Codex 0.118.0 or Claude Code 2.1.283. 



\begin{table}[!htbp]
\centering\small
\setlength{\tabcolsep}{3pt}
\renewcommand{\arraystretch}{1.16}
\begin{tabular}{@{}>{\raggedright\arraybackslash}p{0.14\linewidth}>{\raggedright\arraybackslash}p{0.07\linewidth}>{\raggedright\arraybackslash}p{0.10\linewidth}>{\raggedright\arraybackslash}p{0.21\linewidth}>{\raggedright\arraybackslash}p{0.07\linewidth}>{\raggedright\arraybackslash}p{0.28\linewidth}@{}}
\toprule
\rowcolor{ExpHeader}
Task batch & Tasks & CPUs & Memory (MB) & GPUs & Configured actor time limit \\
\midrule
CPU tasks & 67 & 2--16 & 4,096--49152 & 0 & 30 minutes--6 hours \\
GPU tasks & 8 & 8--32 & 32,768--262,144 & 1--4 & 2, 3, 4, 6, or 48 hours, depending on the task \\

\bottomrule
\end{tabular}
\caption{Declared task resources and configured time limits. CPU, memory, and GPU counts come from the task configuration bound to each selected source. Time limits are calculated from overrides and multipliers in the run configuration and specify the actor execution budget. The physical hosts and GPU models remain to be verified.}
\label{tab:experiment-resources}
\end{table}



\textbf{Rollouts and metrics.} Each model performs three rollouts per task, with each rollout scored independently against the checklist. A GLM-5.3 judge issues the verdicts. Following the preceding definitions, \textbf{Instance Success Rate (ISR)} requires all applicable checklist items for a task to pass. \textbf{Checklist Item Success Rate (CSR)} first computes the fraction of passing items within each task and then averages equally across tasks. The main table computes ISR and CSR separately for each round and reports the three-round mean and sample standard deviation, corresponding to single-attempt pass@1 evaluation. 


\clearpage
The experiments address the following questions.


\begin{tcolorbox}[callout,colback=calloutlight]
\textbf{RQ1: Can current models fully complete user tasks in real research workspaces?}\par Integrating the current request with historical context and workspace requirements\par\smallskip
\textbf{RQ2: Why do agents fail to complete research deliveries in real research workspaces?}\par Correctness bottlenecks and omissions of historical, material, and workspace requirements\par\smallskip
\textbf{RQ3: Which research stages are most difficult?}\par Stage difficulty and the distribution of model capabilities\par\smallskip
\textbf{RQ4: Do larger contexts and higher costs correspond to greater strict success?}\par The relationship between resource use and success\par\smallskip
\textbf{RQ5: How does changing the LLM judge affect binary verdicts and model comparisons?}\par Item agreement and ranking sensitivity\par\smallskip
\end{tcolorbox}

\par\WFclear

\FloatBarrier

\subsection{Main Results}

\par\WFclear

\FloatBarrier

\subsubsection{RQ1: Can current models fully complete user tasks in real research workspaces?}



\begin{table}[!htbp]
\centering
\begin{minipage}[t]{0.56\textwidth}
\vspace{0pt}
\centering\small
\setlength{\tabcolsep}{3pt}
\renewcommand{\arraystretch}{1.16}
\begin{tabular}{@{}>{\raggedright\arraybackslash}p{0.36\linewidth}>{\raggedright\arraybackslash}p{0.29\linewidth}>{\raggedright\arraybackslash}p{0.29\linewidth}@{}}
\toprule
\rowcolor{ExpHeader}
Model & Overall ISR $\uparrow$ & Overall CSR $\uparrow$ \\
\midrule
GPT-6.1-Sol & \ExpScoreCell{44}{66.22 $\pm$ 5.39} & \ExpScoreCell{78}{92.72 $\pm$ 2.08} \\
Claude Opus 5.5 & \ExpScoreCell{35}{56.89 $\pm$ 4.29} & \ExpScoreCell{75}{90.20 $\pm$ 1.83} \\
DeepSeek-V4.1-Flash & \ExpScoreCell{21}{39.56 $\pm$ 3.85} & \ExpScoreCell{73}{89.26 $\pm$ 1.72} \\
Qwen3.8-Max & \ExpScoreCell{20}{37.78 $\pm$ 4.29} & \ExpScoreCell{68}{85.45 $\pm$ 0.90} \\
GLM-5.3-Flash & \ExpScoreCell{16}{30.67 $\pm$ 7.06} & \ExpScoreCell{70}{86.66 $\pm$ 0.68} \\
Kimi-K3 & \ExpScoreCell{14}{28.00 $\pm$ 2.31} & \ExpScoreCell{68}{85.33 $\pm$ 0.50} \\
Qwen3.8-27B & \ExpScoreCell{13}{25.78 $\pm$ 0.77} & \ExpScoreCell{60}{79.66 $\pm$ 0.24} \\
\bottomrule
\end{tabular}
\caption{Overall performance of seven models on 75 tasks. ISR and CSR are percentages, reported as the three-round mean $\pm$ sample standard deviation, with higher values indicating better performance. Darker blue indicates higher percentages on a shared quadratic scale.}
\label{tab:main-results}
\end{minipage}\hfill
\begin{minipage}[t]{0.40\textwidth}
\vspace{0pt}
\centering
\includegraphics[width=\linewidth,trim={6.6 14.5 6.6 5.9},clip]{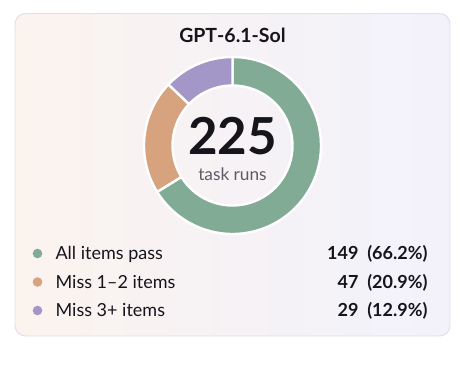}
\captionof{figure}{Breakdown of GPT-6.1-Sol outcomes. The unit is a task run, and a run may fail multiple checklist items.}
\label{fig:failure-breakdown}
\end{minipage}
\end{table}


\textbf{Current models remain unreliable at completing research requests in real workspaces.} As Table~\ref{tab:main-results} shows, even the strongest model, GPT-6.1-Sol, achieves only about 66.2\% strict task success, leaving roughly one-third of runs unable to satisfy all task requirements. Claude Opus 5.5 reaches 56.9\%, all remaining models stay below 40\%, and Qwen3.8-27B achieves only about 25.8\%. Real research requests build on prior work. Background summarizes earlier requests still in effect, and Requirements specifies the current delivery and acceptance conditions, whose project-specific meaning must be interpreted through the restored code, papers, configurations, and logs. Full completion therefore requires models to situate the current request within its project context and identify the requirements that the current changes must continue to satisfy.


CSR exceeds overall ISR for every model, ranging from 79.66\% to 92.72\%. Qwen3.8-Max's higher ISR but lower CSR than GLM-5.3-Flash is consistent with these definitions: ISR measures fully successful deliveries, while CSR measures average checklist completion across tasks, so the metrics need not rank models identically. The per-run failure distribution clarifies this gap: of GPT-6.1-Sol's 76 failed runs, 47 miss only one or two checklist items, accounting for 61.84\%, and this share ranges from 53.09\% to 72.06\% across the seven models. Many failures therefore occur after the model has made progress and satisfied most checklist items. \textbf{Full delivery requires models to jointly address the current request and project requirements still in effect.} Even after resolving the most prominent issue in the current turn, overlooking prior agreements in Background or methodological requirements carried by the workspace can leave critical gaps. RQ2 examines how these gaps manifest in archived deliveries. Figure~\ref{fig:failure-breakdown} breaks down the strongest model's failures.



\par\WFclear

\FloatBarrier

\subsubsection{RQ2: Why do agents fail to complete research deliveries in real research workspaces?}

\par\WFclear

\FloatBarrier

\paragraph{RQ2.1: Which checklist items most often block full delivery?}



\begin{table}[!htbp]
\centering\small
\setlength{\tabcolsep}{3pt}
\renewcommand{\arraystretch}{1.16}
\begin{tabular}{@{}>{\raggedright\arraybackslash}p{0.25\linewidth}>{\raggedright\arraybackslash}p{0.16\linewidth}>{\raggedright\arraybackslash}p{0.16\linewidth}>{\raggedright\arraybackslash}p{0.17\linewidth}>{\raggedright\arraybackslash}p{0.20\linewidth}@{}}
\toprule
\rowcolor{ExpHeader}
Model & Completeness ISR & Correctness ISR & Constraint compliance ISR & Failed runs with correctness errors \\
\midrule
GPT-6.1-Sol & \ExpScoreCell{65}{83.11 $\pm$ 5.39} & \ExpScoreCell{49}{71.11 $\pm$ 5.05} & \ExpScoreCell{80}{93.78 $\pm$ 0.77} & \ExpGapCell{68}{65/76 (85.5\%)} \\
Claude Opus 5.5 & \ExpScoreCell{71}{87.56 $\pm$ 3.08} & \ExpScoreCell{40}{62.67 $\pm$ 3.53} & \ExpScoreCell{73}{89.33 $\pm$ 4.62} & \ExpGapCell{69}{84/97 (86.6\%)} \\
DeepSeek-V4.1-Flash & \ExpScoreCell{64}{82.67 $\pm$ 2.67} & \ExpScoreCell{26}{47.11 $\pm$ 1.54} & \ExpScoreCell{73}{88.89 $\pm$ 5.05} & \ExpGapCell{71}{119/136 (87.5\%)} \\
Qwen3.8-Max & \ExpScoreCell{59}{79.11 $\pm$ 2.04} & \ExpScoreCell{24}{43.56 $\pm$ 8.57} & \ExpScoreCell{67}{84.89 $\pm$ 0.77} & \ExpGapCell{75}{127/140 (90.7\%)} \\
GLM-5.3-Flash & \ExpScoreCell{61}{80.44 $\pm$ 0.77} & \ExpScoreCell{21}{39.11 $\pm$ 3.08} & \ExpScoreCell{63}{81.78 $\pm$ 1.54} & \ExpGapCell{71}{137/156 (87.8\%)} \\
Kimi-K3 & \ExpScoreCell{56}{76.44 $\pm$ 1.54} & \ExpScoreCell{18}{34.22 $\pm$ 2.04} & \ExpScoreCell{65}{83.56 $\pm$ 1.54} & \ExpGapCell{77}{148/162 (91.4\%)} \\
Qwen3.8-27B & \ExpScoreCell{51}{72.00 $\pm$ 1.33} & \ExpScoreCell{16}{31.11 $\pm$ 0.77} & \ExpScoreCell{59}{79.11 $\pm$ 2.04} & \ExpGapCell{79}{155/167 (92.8\%)} \\
\bottomrule
\end{tabular}
\caption{ISR by completeness, correctness, and constraint compliance. Each dimension requires all its applicable checklist items to pass. The final column reports the count and share of failed runs containing correctness errors. Error categories can overlap. Darker blue indicates higher ISR on a shared quadratic percentage scale; darker orange indicates a greater share of failures with correctness errors.}
\label{tab:dimension-failures}
\end{table}


Table~\ref{tab:dimension-failures} identifies correctness as the lowest-scoring dimension for all seven models. Correctness ISR ranges from 31.11\% to 71.11\%, compared with 72.00\%--87.56\% for completeness and 79.11\%--93.78\% for constraint compliance. The same pattern holds for CSR, with correctness lowest for every model. At the run level, 85.5\%--92.8\% of failures include an unmet correctness item. DeepSeek-V4.1-Flash reaches 93.16\% completeness CSR but only 47.11\% correctness ISR and 39.56\% overall ISR, illustrating the gap between delivery coverage and content reliability.


\textbf{Research requirements must hold throughout implementation and validation, as local functional recovery can still lose methodological meaning.} RQ1 concerns whether models integrate the context needed for a task, while Table~\ref{tab:dimension-failures} shows the difficulty of accurately realizing its requirements in research artifacts. The current request may ask only for training repair or framework adaptation, but the model must also preserve the original method and engineering conventions using Background and the papers, code, and design documents in the workspace. The archived examples in Table~\ref{tab:failure-examples} show the consequences of failing to implement these requirements. The CylindFormer repair retains the KPConv name but omits neighborhood and kernel-point aggregation, while the Graph RAG adaptation implements a local workflow without isolating indexes across datasets. Delivering files or code can satisfy some completeness requirements while deviations between the method and its implementation fail correctness items.



\begin{table}[!htbp]
\centering\small
\setlength{\tabcolsep}{3pt}
\renewcommand{\arraystretch}{1.16}
\begin{tabular}{@{}>{\raggedright\arraybackslash}p{0.17\linewidth}>{\raggedright\arraybackslash}p{0.29\linewidth}>{\raggedright\arraybackslash}p{0.48\linewidth}@{}}
\toprule
\rowcolor{ExpHeader}
Failure manifestation & Effect on research delivery & Archived example \\
\midrule
Task degradation & A simpler implementation replaces the original method, retaining its name while changing the required capability & In the round 1 DeepSeek repair of CylindFormer, the backbone named KPConv applies only Linear, LayerNorm, and ReLU to individual points, without neighborhood or kernel-point aggregation, failing O4 \\
Insufficient investigation and evidence checking & Cited work does not support the mechanism or subject discussed, leaving conclusions without reliable evidence & In the round 1 Qwen3.8-27B revision of a cuproptosis review, ferroptosis-resistance research is described as a cuproptosis mechanism and cancer subtypes are conflated, failing O6 \\
Requirement omission & Existing constraints are not implemented, affecting subsequent use despite local functional progress & In the round 2 DeepSeek adaptation of Graph RAG, datasets share a global index directory instead of isolated indexes, failing O11 \\
Inadequate validation design & Missing comparison rules prevent attribution of improvements to the factor under study & In the round 1 GLM-5.3-Flash plan for 4D spacetime tubes, costs are recorded but executable rules for matching inference budgets across comparison arms are absent, failing O6 \\
Conclusions exceeding the evidence & A limited diagnostic result is extended into an untested causal explanation & In the round 2 GLM-5.3-Flash diagnosis of network-security generalization, adversarial validation is treated as a test of conditional distributions, and a failure to detect differences is attributed to a single cause, failing O13 \\
\bottomrule
\end{tabular}
\caption{Failure manifestations and archived examples. These selected qualitative examples explain why checklist items fail. They do not constitute a uniform classification or frequency analysis of all failures.}
\label{tab:failure-examples}
\end{table}


These examples illustrate the difficulty of full delivery through method preservation, requirement implementation, and evidence checking. Models must interpret Background and Requirements together and verify that the requirements still hold against existing workspace artifacts after making changes. For example, resuming training also requires checking the original method's key computations, and adapting a framework requires checking its behavior across datasets. Reliable continuation means advancing the current work while maintaining the project's existing requirements.

\par\WFclear

\FloatBarrier

\paragraph{RQ2.2: What delivery gaps are revealed by historical and material or workspace requirements?}


Real research requests often briefly describe only the current work, leaving some requirements and constraints implicit in prior agreements and project materials rather than restating them in the current turn. In ARE-bench, Background summarizes historical constraints, while papers, code, configurations, and existing results in the workspace clarify the specific meaning of methods and acceptance conditions. The agent must therefore interpret the current request within the project context and identify requirements that its changes must continue to respect. To examine these gaps, we hold actor deliveries and item-level judge verdicts fixed and change only the checklist items included in ISR. We first check delivery contracts and general professional norms, then add historical and material or workspace requirements, and finally check the complete checklist. Table~\ref{tab:checklist-sources} contrasts three models with similar scores under the first scope.



\begin{table}[!htbp]
\centering\small
\setlength{\tabcolsep}{3pt}
\renewcommand{\arraystretch}{1.16}
\begin{tabular}{@{}>{\raggedright\arraybackslash}p{0.27\linewidth}>{\raggedright\arraybackslash}p{0.22\linewidth}>{\raggedright\arraybackslash}p{0.29\linewidth}>{\raggedright\arraybackslash}p{0.16\linewidth}@{}}
\toprule
\rowcolor{ExpHeader}
Model & Delivery contracts + general norms & Add historical and material or workspace requirements & Full checklist \\
\midrule
DeepSeek-V4.1-Flash & \ExpScoreCell{47}{69.33} & \ExpScoreCell{21}{40.00} & \ExpScoreCell{21}{39.56} \\
GLM-5.3-Flash & \ExpScoreCell{47}{69.33} & \ExpScoreCell{16}{31.56} & \ExpScoreCell{16}{30.67} \\
Kimi-K3 & \ExpScoreCell{48}{69.78} & \ExpScoreCell{15}{28.44} & \ExpScoreCell{14}{28.00} \\
\bottomrule
\end{tabular}
\caption{ISR (\%) as the scope of checklist evaluation expands. Every model covers the same 75 tasks with three rollouts per task, and values are three-round means. Each column requires all applicable items within its scope to pass, reusing the same deliveries and item verdicts. Delivery contracts and general norms correspond to \texttt{contract + norm}. The second scope adds \texttt{inherited + anchored}, while the full checklist also includes \texttt{latent} items discovered through subsequent rework but inferable at the task cut point. Darker blue indicates higher ISR on a shared quadratic percentage scale.}
\label{tab:checklist-sources}
\end{table}


As Table~\ref{tab:checklist-sources} shows, all three models achieve approximately 69\% ISR when only delivery contracts and general norms are checked. Adding historical and material or workspace requirements reduces DeepSeek, GLM, and Kimi to 40.00\%, 31.56\%, and 28.44\%, respectively. Among runs passing the first group of checks, 66, 85, and 93 still fail at least one historical or material or workspace requirement, accounting for 48.5\%, 54.5\%, and 57.4\% of each model's total failed runs. DeepSeek and Kimi differ by only 0.44 percentage points under the first scope, compared with 11.56 points under the full checklist.


\textbf{An important bottleneck in real research delivery is satisfying implicit requirements that remain binding through project context but are not restated in the current turn.} Roughly half of the failed runs for these three models satisfy delivery contracts and general norms yet are blocked by historical or material or workspace requirements. Thus, even a model that performs the action described in the current request may fail to deliver a usable result because it departs from prior research agreements, material-grounded methods, or the computations the project requires. Reliable research continuation requires the agent to integrate these distributed constraints and preserve them throughout modification and validation.


This capability requirement connects back to the distinction in task granularity and starting state established earlier. Compared with a setting that pursues a standalone end-to-end objective from an initial state, ARE-bench starts at a user request within an ongoing real project, retaining historical agreements still in effect in Background and existing artifacts in the cut-point workspace. \textbf{Evaluation concerns both achieving the current objective and maintaining the project's existing research requirements while doing so.} Table~\ref{tab:checklist-sources} shows that this dimension reveals capability differences that are difficult to observe when only delivery contracts and general norms are checked.

\par\WFclear

\FloatBarrier

\subsection{Stage Analysis}

\par\WFclear

\FloatBarrier

\subsubsection{RQ3: Which research stages are most difficult?}



\begin{table}[!htbp]
\centering\small
\setlength{\tabcolsep}{3pt}
\renewcommand{\arraystretch}{1.16}
\resizebox{\linewidth}{!}{%
\begin{tabular}{lccccccc}
\toprule
\rowcolor{ExpHeader}
Model & Survey & Design & Data \& Env. & Code & Execution & Analysis & Writing \\
\midrule
GPT-6.1-Sol & \ExpScoreCell{72}{88.46 $\pm$ 6.66} & \ExpScoreCell{52}{73.17 $\pm$ 4.88} & \ExpScoreCell{74}{90.00 $\pm$ 0.00} & \ExpScoreCell{68}{85.42 $\pm$ 9.55} & \ExpScoreCell{56}{76.77 $\pm$ 7.00} & \ExpScoreCell{69}{86.27 $\pm$ 6.12} & \ExpScoreCell{61}{80.65 $\pm$ 5.59} \\
Claude Opus 5.5 & \ExpScoreCell{67}{84.62 $\pm$ 3.85} & \ExpScoreCell{56}{76.42 $\pm$ 2.82} & \ExpScoreCell{67}{85.00 $\pm$ 5.00} & \ExpScoreCell{57}{77.08 $\pm$ 7.86} & \ExpScoreCell{44}{66.67 $\pm$ 3.03} & \ExpScoreCell{60}{79.41 $\pm$ 2.94} & \ExpScoreCell{42}{64.52 $\pm$ 11.63} \\
DeepSeek-V4.1-Flash & \ExpScoreCell{60}{79.49 $\pm$ 4.44} & \ExpScoreCell{40}{62.60 $\pm$ 2.82} & \ExpScoreCell{82}{95.00 $\pm$ 0.00} & \ExpScoreCell{50}{71.88 $\pm$ 5.41} & \ExpScoreCell{39}{61.62 $\pm$ 4.63} & \ExpScoreCell{56}{76.47 $\pm$ 0.00} & \ExpScoreCell{41}{63.44 $\pm$ 6.72} \\
Qwen3.8-Max & \ExpScoreCell{58}{78.21 $\pm$ 4.44} & \ExpScoreCell{36}{58.54 $\pm$ 7.32} & \ExpScoreCell{65}{83.33 $\pm$ 12.58} & \ExpScoreCell{46}{67.71 $\pm$ 6.51} & \ExpScoreCell{29}{50.51 $\pm$ 4.63} & \ExpScoreCell{38}{60.78 $\pm$ 1.70} & \ExpScoreCell{46}{67.74 $\pm$ 5.59} \\
GLM-5.3-Flash & \ExpScoreCell{57}{76.92 $\pm$ 3.85} & \ExpScoreCell{35}{56.91 $\pm$ 6.14} & \ExpScoreCell{67}{85.00 $\pm$ 5.00} & \ExpScoreCell{41}{63.54 $\pm$ 1.80} & \ExpScoreCell{37}{59.60 $\pm$ 1.75} & \ExpScoreCell{50}{71.57 $\pm$ 6.12} & \ExpScoreCell{36}{58.06 $\pm$ 5.59} \\
Kimi-K3 & \ExpScoreCell{49}{70.51 $\pm$ 4.44} & \ExpScoreCell{27}{47.97 $\pm$ 2.82} & \ExpScoreCell{70}{86.67 $\pm$ 7.64} & \ExpScoreCell{53}{73.96 $\pm$ 4.77} & \ExpScoreCell{26}{47.47 $\pm$ 3.50} & \ExpScoreCell{42}{64.71 $\pm$ 0.00} & \ExpScoreCell{26}{47.31 $\pm$ 3.72} \\
Qwen3.8-27B & \ExpScoreCell{43}{65.38 $\pm$ 6.66} & \ExpScoreCell{22}{40.65 $\pm$ 2.82} & \ExpScoreCell{60}{80.00 $\pm$ 5.00} & \ExpScoreCell{37}{59.38 $\pm$ 5.41} & \ExpScoreCell{21}{39.39 $\pm$ 0.00} & \ExpScoreCell{36}{58.82 $\pm$ 2.94} & \ExpScoreCell{30}{51.61 $\pm$ 0.00} \\
\bottomrule
\end{tabular}}
\caption{ISR across seven research stages. A task passes a stage only if all its applicable checklist items in that stage pass. Tasks without items in the stage are excluded from its denominator. Values are three-round means $\pm$ sample standard deviations, and one task may involve multiple stages. Darker blue indicates higher ISR on a shared quadratic percentage scale.}
\label{tab:stage}
\end{table}



\begin{table}[!htbp]
\centering\small
\setlength{\tabcolsep}{3pt}
\renewcommand{\arraystretch}{1.16}
\begin{tabular}{@{}>{\raggedright\arraybackslash}p{0.19\linewidth}>{\raggedright\arraybackslash}p{0.13\linewidth}>{\raggedright\arraybackslash}p{0.16\linewidth}>{\raggedright\arraybackslash}p{0.23\linewidth}>{\raggedright\arraybackslash}p{0.23\linewidth}@{}}
\toprule
\rowcolor{ExpHeader}
Research stage & Tasks covered & Checklist items & Seven-model mean stage ISR & Best-to-worst model gap \\
\midrule
Survey & 26 & 111 & \ExpScoreCell{57}{77.66} & \ExpGapCell{12}{23.08} \\
Design & 41 & 268 & \ExpScoreCell{37}{59.47} & \ExpGapCell{18}{35.77} \\
Data \& Env. & 20 & 45 & \ExpScoreCell{69}{86.43} & \ExpGapCell{10}{15.00} \\
Code & 32 & 217 & \ExpScoreCell{50}{71.28} & \ExpGapCell{14}{26.04} \\
Execution & 33 & 173 & \ExpScoreCell{35}{57.43} & \ExpGapCell{19}{37.37} \\
Analysis & 34 & 92 & \ExpScoreCell{50}{71.15} & \ExpGapCell{14}{27.45} \\
Writing & 31 & 267 & \ExpScoreCell{39}{61.90} & \ExpGapCell{17}{33.33} \\
\bottomrule
\end{tabular}
\caption{Stage coverage and descriptive difficulty. The seven-model mean weights models equally, and the best-to-worst gap is in percentage points. Task counts indicate stage coverage in the benchmark, while effective denominators follow the scored records in each round. Darker blue indicates higher mean ISR, and darker orange indicates larger model gaps, on a shared quadratic percentage scale.}
\label{tab:stage-coverage}
\end{table}


Tables~\ref{tab:stage} and~\ref{tab:stage-coverage} show that Design and Execution are among the three most difficult stages for every model. The seven-model mean ISR is 86.43\% for Data \& Env. and 71.28\% for Code, falling to 59.47\% for Design and 57.43\% for Execution. Design and Execution also show the largest gaps between models, at 35.77 and 37.37 percentage points, respectively. Stage strengths differ: GPT-6.1-Sol leads five stages, Claude Opus 5.5 leads Design, and DeepSeek-V4.1-Flash leads Data \& Env. Kimi-K3 ranks third on Code at 73.96\%, but its Execution and Writing scores are only 47.47\% and 47.31\%, ranking second-last and last, respectively.


\textbf{Strength in code implementation does not yet ensure effective research work.} Dependency installation and code executability provide relatively direct verification feedback. Design and Execution additionally require judging whether an experiment can answer the research question and taking appropriate action based on existing scripts, logs, and checkpoints. These stage differences indicate that implementation strength must be assessed separately in subsequent execution, validation, and reporting. A single score cannot capture this capability distribution. Stages come from naturally occurring research sessions, so this interpretation concerns the evaluated task distribution. Future ablations can further distinguish the contributions of context and workspace state.

\par\WFclear

\FloatBarrier

\subsection{Session Position and Cut-Point Selection}

\label{sec:session-position}

\par\WFclear

\FloatBarrier

\subsubsection{Supplementary question: Are tasks near the session midpoint more challenging than those at adjacent positions?}



\begin{table}[!htbp]
\centering\small
\setlength{\tabcolsep}{3pt}
\renewcommand{\arraystretch}{1.16}
\begin{tabular}{@{}>{\raggedright\arraybackslash}p{0.72\linewidth}>{\raggedright\arraybackslash}p{0.22\linewidth}@{}}
\toprule
\rowcolor{ExpHeader}
Position relative to the complete session midpoint & CSR (\%) \\
\midrule
Before the midpoint, original segment 2 & \ExpScoreCell{66}{95.92} \\
Near the midpoint, original segment 3 & \ExpScoreCell{8}{91.63} \\
After the midpoint, original segment 4 & \ExpScoreCell{90}{97.66} \\
\bottomrule
\end{tabular}
\caption{CSR at and around the midpoint of the complete session. Each of 10 sessions is divided into five consecutive segments by user-turn count. We show the original segments 2--4, each containing 10 tasks, for 30 tasks in total. GLM-5.3 runs on the instruction and restored workspace at each cut point, and results weight tasks equally. Blue intensity is scaled over the displayed CSR range, with darker cells indicating higher values.}
\label{tab:session-position}
\end{table}


As Table~\ref{tab:session-position} shows, CSR near the midpoint is 91.63\%, lower than the preceding and following segments by 4.29 and 6.03 percentage points, respectively. Within these three adjacent segments, tasks in the middle of the session are more challenging. Models can already use the state summary in Background and artifacts in the restored workspace, but must still determine how earlier decisions constrain the current request and advance the research accordingly.


\textbf{The midpoint of a complete session provides candidate cut points that remain challenging despite the available research state.} This observation gives empirical support to the construction practice of typically seeking candidate requests near the complete user session's midpoint. Candidates come from sessions that pass Curator rules S1 to S4 and must also pass the Builder's replay checks. After selection, earlier requests still in effect are summarized in Background, the current request and delivery and acceptance conditions are retained in Requirements, and the workspace is replayed to the cut point. The experiment therefore examines an in-progress research state consistent with the construction procedure.

\par\WFclear

\FloatBarrier

\subsection{Cost Analysis}

\par\WFclear

\FloatBarrier

\subsubsection{RQ4: Do larger contexts and higher costs correspond to greater strict success?}




\begin{figure}[!htbp]
\centering
\includegraphics[width=\linewidth]{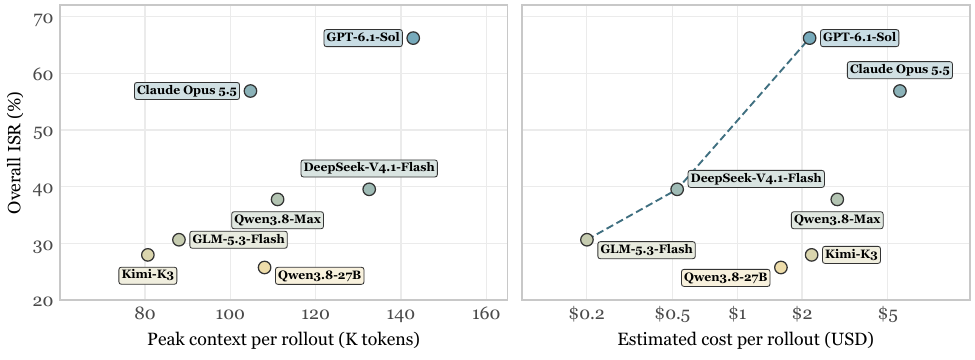}
\caption{Mean peak context, estimated cost, and ISR per run. Peak context is the largest single API call input within a run, and K denotes thousands. Estimates apply the manuscript's frozen prices dated 2026-11-02 to the measured uncached input, cache reads, cache writes, and output tokens of every call, covering model API charges. Cache writes use the ordinary input price, except that Claude Opus 5.5 uses its cache-write price of USD 5 per million tokens. DeepSeek's context and cost use 222 runs with token records, while its ISR uses all 225 scored runs. Other models use their available records.}
\label{fig:cost}
\end{figure}


Figure~\ref{fig:cost} shows that neither peak context nor cost aligns with strict success across models. DeepSeek-V4.1-Flash reaches a peak context of 132.55K, close to the 142.87K of GPT-6.1-Sol, but achieves 39.56\% ISR compared with 66.22\%. Claude Opus 5.5 and Qwen3.8-27B reach similar peak contexts of 104.71K and 108.02K, yet their ISR differs by 31.11 percentage points. Under the frozen prices, GPT-6.1-Sol costs about USD 2.17 per run and Claude Opus 5.5 about USD 5.69, roughly 2.62 times as much for an ISR lower by 9.33 percentage points. GLM-5.3-Flash and DeepSeek-V4.1-Flash offer 30.67\% and 39.56\% ISR at approximately USD 0.20 and USD 0.53, respectively.


\textbf{Larger contexts and higher costs do not automatically produce more reliable research deliveries.} Under this task distribution and these frozen prices, GLM-5.3-Flash, DeepSeek-V4.1-Flash, and GPT-6.1-Sol form the cost-success Pareto frontier, offering choices at different cost levels. The additional resources spent on longer trajectories constitute progress only when they satisfy research requirements. These results suggest that the value of resource use lies in helping models check constraints, complete validation, and resolve remaining errors. Cross-model comparisons reveal this efficiency difference, while the benefit of increasing a single model's budget requires a separate experiment.

\par\WFclear

\FloatBarrier

\subsection{Judge Agreement and Result Sensitivity}

\par\WFclear

\FloatBarrier

\subsubsection{RQ5: How does changing the LLM judge affect binary verdicts and model comparisons?}



\begin{table}[!htbp]
\centering\small
\setlength{\tabcolsep}{3pt}
\renewcommand{\arraystretch}{1.16}
\begin{tabular}{@{}>{\raggedright\arraybackslash}p{0.40\linewidth}>{\raggedright\arraybackslash}p{0.18\linewidth}>{\raggedright\arraybackslash}p{0.18\linewidth}>{\raggedright\arraybackslash}p{0.18\linewidth}@{}}
\toprule
\rowcolor{ExpHeader}
DeepSeek verdict / Qwen verdict & Pass & Fail & Total \\
\midrule
Pass & \ExpScoreCell{90}{8,043} & \ExpScoreCell{9}{67} & \cellcolor{ExpHeader}8,110 \\
Fail & \ExpScoreCell{15}{648} & \ExpScoreCell{19}{1,116} & \cellcolor{ExpHeader}1,764 \\
\cellcolor{ExpHeader}Total & \cellcolor{ExpHeader}8,691 & \cellcolor{ExpHeader}1,183 & \cellcolor{ExpHeader}9,874 \\
\bottomrule
\end{tabular}
\caption{DeepSeek and Qwen binary verdicts on matched inputs and checklist items. Starting from 657 matched scoring pairs, matching checklist-item IDs and content hashes yields 9,874 paired binary verdicts. The judges are \texttt{deepseek-v4.1-flash-1017} and \texttt{qwen3.8-flash}. These results cover a subset of tasks. Darker blue indicates larger joint counts; marginal totals use neutral shading.}
\label{tab:judge-verdicts}
\end{table}


\begin{figure}[!htbp]
\centering
\includegraphics[width=0.75\linewidth]{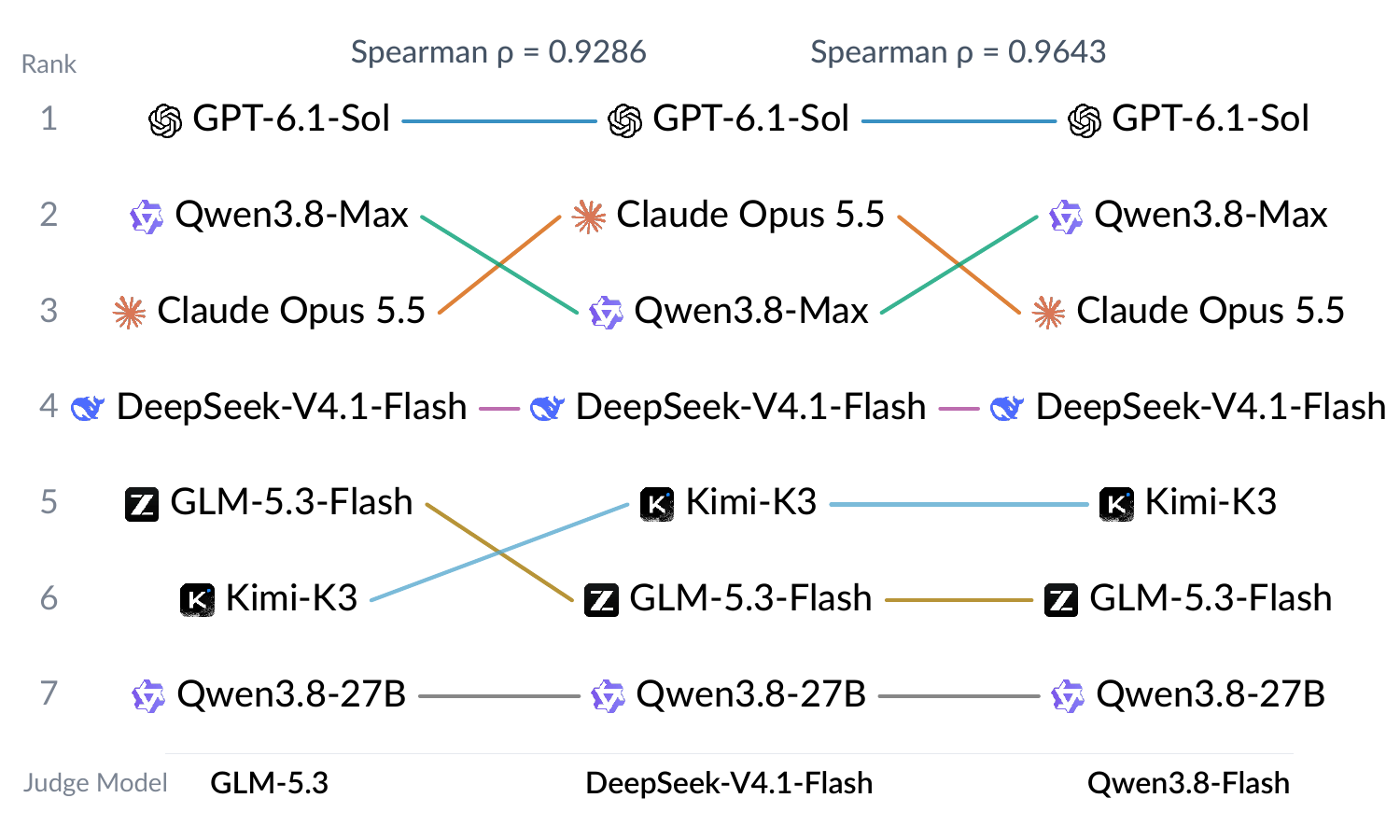}
\caption{Cross-judge CSR ranks on the 20-task common-coverage subset. The columns show the historical default GLM reference, DeepSeek judge, and Qwen judge. Nodes use model logos and short names, and the same actor model is linked by the same color. More stable ranks produce more horizontal lines. The subset contains 147 actor trajectories, 21 per model. Scores first average included rounds within each task and then weight tasks equally. Current DeepSeek and Qwen inputs match, while historical default scores are a reference only. Tasks have different numbers of included rounds, and the subset retains the stage package's source selection and policy of deferring low-scoring sources.}
\label{fig:judge-ranks}
\end{figure}

On the same 20-task common-coverage subset, we compare cross-judge ranking agreement for ISR and CSR. The mean Spearman correlation across the three judge pairs is 0.9524 for CSR, exceeding 0.9028 for ISR, so Figure~\ref{fig:judge-ranks} uses CSR ranks. Correlations preserve ISR ties rather than arbitrarily ordering models with equal scores.




Table~\ref{tab:judge-verdicts} contains 9,159 agreeing binary verdicts, giving 92.76\% agreement and Cohen's $\kappa$ of 0.717 after adjustment for chance agreement using marginal pass rates. Of 715 disagreements, 648, or 90.63\%, are DeepSeek fail and Qwen pass, indicating a directional difference in acceptance. In Figure~\ref{fig:judge-ranks}, GPT-6.1-Sol ranks first under all three scoring sources. Spearman rank correlation is 0.9643 between DeepSeek and Qwen, and 0.9286 and 0.9643 between the historical reference and the two judges, respectively. Rank changes mainly involve adjacent models.


\textbf{Changing the judge preserves most binary verdicts and the model ranking.} Disagreements concentrate in one direction and mainly reorder adjacent models. Independent human annotations are the next step for attributing disagreements and completing the evidence on judge reliability.

\par\WFclear

\FloatBarrier

\subsection{Case Study}

\par\WFclear

\FloatBarrier

\subsubsection{Case question: Does resuming training mean that a research repair is complete?}



\begin{table}[!htbp]
\centering\small
\setlength{\tabcolsep}{3pt}
\renewcommand{\arraystretch}{1.16}
\begin{tabular}{@{}>{\raggedright\arraybackslash}p{0.20\linewidth}>{\raggedright\arraybackslash}p{0.37\linewidth}>{\raggedright\arraybackslash}p{0.37\linewidth}@{}}
\toprule
\rowcolor{ExpHeader}
Evaluation aspect & Successful execution, 20/20 & Execution with unmet items, 16/20 \\
\midrule
Training progress & Sustained nonzero loss and parameter updates & Also sustained training, nonzero loss, and parameter updates \\
O7, fusion coordinates & Projection indexes agree with the actual feature grid & Inconsistent coordinates clip some samples to boundary columns \\
O9, overlap labels & Separate 2D and 3D overlap checks retain positive, negative, and ignored regions & A single projection ratio replaces independent 2D and 3D overlap tests \\
O11, coarse matching loss & L2 distance, overlap weights, and specified settings are retained & Squared distance replaces L2 without corresponding boundary-parameter changes, and weights are altered \\
O12, fine matching loss & Bidirectional loss and the coarse/fine loss combination are retained & The changed distance and weight expressions are reused \\
Training and delivery items & Archived review accepts checkpoints, limited validation, and the report & The other 16 items pass, including training recovery, checkpoints, validation, and the report \\
\bottomrule
\end{tabular}
\caption{Two executions of the CylindFormer training-repair task. The task ID is \texttt{cylindformer-training-repair}. Both executions start at the same repair boundary and must resume training according to the reference paper. Scores follow the archived review. Appendix~\ref{app:training-repair-case} contains the complete inputs, restored workspace, and 20-item checklist.}
\label{tab:repair-comparison}
\end{table}


\begin{wrapfigure}{r}{0.4\textwidth}
\centering
\captionsetup{font=footnotesize,skip=4pt}
\vspace{-0.4cm}
\includegraphics[width=\linewidth]{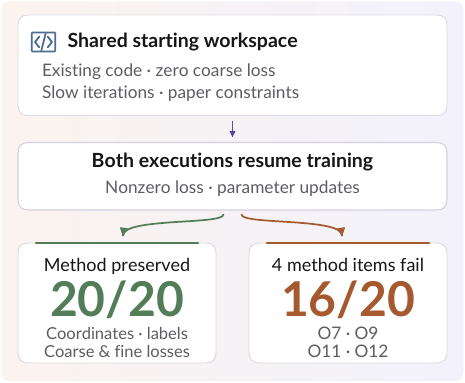}
\caption{Divergent executions of the same training-repair request. The comparison concerns method consistency during recovery. Differences in final registration accuracy still require separate measurement.}
\label{fig:repair-divergence}
\end{wrapfigure}

In Table~\ref{tab:repair-comparison}, both executions resume image-to-point-cloud registration training with nonzero loss and parameter updates, but pass 20/20 and 16/20 checklist items, respectively. The execution with unmet items has four methodological deviations in projection and feature-grid coordinates, 2D/3D overlap labels, coarse matching loss, and fine matching loss. The successful execution resumes training while preserving the paper's method. Figure~\ref{fig:repair-divergence} connects their shared execution progress to their different methodological validity.




\textbf{Program recovery and faithful repair of the research method are distinct outcomes that require separate verification.} Nonzero loss and training progress provide evidence of execution recovery, while checking coordinates, labels, and objectives establishes whether the repair preserves methodological meaning. This comparison grounds RQ2's correctness bottleneck in specific failures and shows how restored workspaces and retained research constraints jointly support evaluation of requests within ongoing research.


\par\WFclear

\FloatBarrier

\section{Conclusion}

ARE-bench evaluates the next research action inside an evolving project. Every task starts from the replayed mid-session workspace with unfinished files and failed attempts preserved, while its task brief gives the actor a self-contained and auditable specification. The ISR and CSR checklist metrics expose strict completion and partial compliance, while separate cost metrics expose efficiency trade-offs. Replay checks and anonymization make the benchmark suitable for comparing research agents without treating an original trajectory as a gold answer. The release protocol supports filling the result schema as model runs become available while keeping the benchmark definition fixed.

\section*{Limitations}

The source sessions may overrepresent particular projects, tools, or researcher populations. Cut-point sampling from existing sessions does not guarantee coverage of all research workflows. Replay and dependency recovery can lose context that was available to the original researcher, and anonymization can remove clues needed for faithful interpretation. Rubric writing remains a human judgment process, and a checklist evaluator can still miss evidence or over-credit a plausible claim. GPT-6.1-Sol runs with the Codex harness and the other models with Claude Code, so cross-model differences reflect the model and harness together. The benchmark therefore publishes provenance records for audit, but its scores should be interpreted as evidence about the released task distribution rather than as a universal measure of scientific ability.

\section*{Ethics Statement}

Session data must be collected under an appropriate authorization or consent arrangement. Before release, identifiers, private paths, credentials, unpublished data, and organization-specific metadata are removed or replaced with stable aliases. The provenance package is access-controlled and is not included in actor inputs. We release only the minimum workspace content needed to reproduce each task and document residual disclosure risks, withdrawal procedures, and the scope of permitted reuse.

\clearpage
\beginappendix

\section{Task and Rubric Example}

A task contains an instruction with Background, Requirements, and Environment sections. A representative requirements block asks the agent to update an existing analysis, save the result to an absolute path, preserve a named preprocessing choice, and report the command used to reproduce the artifact. Its rubric compiles into one binary checklist item for each deliverable, required transformation, preservation constraint, and reproducibility condition. The actor never sees the rubric text.

\section{Research Requirement Taxonomy}
\label{app:requirement-taxonomy}

We use a research requirement taxonomy to characterize the kinds of work researchers request from agents. The taxonomy serves as a reference for benchmark construction and organizes research work into 33 requirements and 207 sub-requirements, grouped into seven research stages. Requirements describe broader areas of work, such as literature review, experiment design, and code testing. Sub-requirements specify individual user needs using action--object descriptions, such as searching for relevant literature or writing unit tests. Table~\ref{tab:requirement-mapping} presents the portion of this taxonomy represented in our benchmark: 150 sub-requirements covering all 33 requirements.

The table organizes research requirements into three levels: research stage, requirement, and sub-requirement. Two additional attributes characterize each sub-requirement. \emph{Capability primitives} are drawn from the cognitive process dimension of the revised Bloom's taxonomy \citep{anderson-krathwohl-2001-taxonomy,krathwohl-2002-revision} and describe the cognitive operations involved, such as differentiating under Analyze, checking under Evaluate, and planning under Create. \emph{Work object} identifies the material these operations act on, such as code, data, or research documents.

Each checklist item is associated with one sub-requirement, which identifies its parent requirement and research stage. A task can therefore involve several requirements and stages through its checklist items.

\begingroup
\footnotesize
\setlength{\tabcolsep}{4pt}
\renewcommand{\arraystretch}{1.20}
\setlength{\extrarowheight}{1pt}
\newlength{\reqwidth}
\setlength{\reqwidth}{\dimexpr\textwidth-8\tabcolsep\relax}
\newcommand{\reqseparator}{%
  \noalign{\begingroup\color{black!7}%
    \hrule width\dimexpr.09\reqwidth+\tabcolsep\relax
      height\dimexpr\aboverulesep+\cmidrulewidth+\belowrulesep\relax
    \kern-\dimexpr\aboverulesep+\cmidrulewidth+\belowrulesep\relax
  \endgroup}%
  \cmidrule(l){2-5}}
\makeatletter
\newcommand{\reqrowstrut}[1]{%
  \rule[-\dimexpr#1-\ht\@arstrutbox\relax]{0pt}{#1}}
\newcommand{\reqgrouplabel}[5]{%
  \raisebox{\dimexpr\ht\@arstrutbox-#3\relax}[0pt][0pt]{%
    \vbox to#2{\hsize=#1\linewidth=\hsize
      \parindent=0pt\parskip=0pt #4\vfil\noindent#5\par\vfil}}}
\makeatother

\setbox0=\hbox{
\begin{tabular}{@{}>{\columncolor{black!7}[0pt][\tabcolsep]}p{.09\reqwidth}p{.15\reqwidth}p{.28\reqwidth}p{.33\reqwidth}>{\columncolor{white}[\tabcolsep][0pt]}p{.15\reqwidth}@{}}\raggedright\arraybackslash  & \raggedright\arraybackslash  & \raggedright\arraybackslash Search for relevant literature & \raggedright\arraybackslash Apply: implement; Analyze: differentiate & \raggedright\arraybackslash Evidence \\\end{tabular}}
\expandafter\edef\csname reqrowheight0\endcsname{\the\dimexpr\ht0+\dp0\relax}
\setbox0=\hbox{\begin{tabular}{@{}>{\columncolor{black!7}[0pt][\tabcolsep]}p{.09\reqwidth}p{.15\reqwidth}p{.28\reqwidth}p{.33\reqwidth}>{\columncolor{white}[\tabcolsep][0pt]}p{.15\reqwidth}@{}}\raggedright\arraybackslash  & \raggedright\arraybackslash  & \raggedright\arraybackslash Verify bibliographic information & \raggedright\arraybackslash Evaluate: critique & \raggedright\arraybackslash Evidence \\\end{tabular}}
\expandafter\edef\csname reqrowheight1\endcsname{\the\dimexpr\ht0+\dp0\relax}
\setbox0=\hbox{\begin{tabular}{@{}>{\columncolor{black!7}[0pt][\tabcolsep]}p{.09\reqwidth}p{.15\reqwidth}p{.28\reqwidth}p{.33\reqwidth}>{\columncolor{white}[\tabcolsep][0pt]}p{.15\reqwidth}@{}}\raggedright\arraybackslash  & \raggedright\arraybackslash  & \raggedright\arraybackslash Extract key points from literature & \raggedright\arraybackslash Analyze: differentiate; Understand: summarize & \raggedright\arraybackslash Evidence \\\end{tabular}}
\expandafter\edef\csname reqrowheight2\endcsname{\the\dimexpr\ht0+\dp0\relax}
\setbox0=\hbox{\begin{tabular}{@{}>{\columncolor{black!7}[0pt][\tabcolsep]}p{.09\reqwidth}p{.15\reqwidth}p{.28\reqwidth}p{.33\reqwidth}>{\columncolor{white}[\tabcolsep][0pt]}p{.15\reqwidth}@{}}\raggedright\arraybackslash  & \raggedright\arraybackslash  & \raggedright\arraybackslash Verify literature content & \raggedright\arraybackslash Evaluate: check & \raggedright\arraybackslash Evidence \\\end{tabular}}
\expandafter\edef\csname reqrowheight3\endcsname{\the\dimexpr\ht0+\dp0\relax}
\setbox0=\hbox{\begin{tabular}{@{}>{\columncolor{black!7}[0pt][\tabcolsep]}p{.09\reqwidth}p{.15\reqwidth}p{.28\reqwidth}p{.33\reqwidth}>{\columncolor{white}[\tabcolsep][0pt]}p{.15\reqwidth}@{}}\raggedright\arraybackslash  & \raggedright\arraybackslash  & \raggedright\arraybackslash Assess literature quality & \raggedright\arraybackslash Evaluate: critique & \raggedright\arraybackslash Evidence \\\end{tabular}}
\expandafter\edef\csname reqrowheight4\endcsname{\the\dimexpr\ht0+\dp0\relax}
\setbox0=\hbox{\begin{tabular}{@{}>{\columncolor{black!7}[0pt][\tabcolsep]}p{.09\reqwidth}p{.15\reqwidth}p{.28\reqwidth}p{.33\reqwidth}>{\columncolor{white}[\tabcolsep][0pt]}p{.15\reqwidth}@{}}\raggedright\arraybackslash  & \raggedright\arraybackslash  & \raggedright\arraybackslash Establish evidence traceability & \raggedright\arraybackslash Analyze: organize; Evaluate: check & \raggedright\arraybackslash Evidence \\\end{tabular}}
\expandafter\edef\csname reqrowheight5\endcsname{\the\dimexpr\ht0+\dp0\relax}
\setbox0=\hbox{\begin{tabular}{@{}>{\columncolor{black!7}[0pt][\tabcolsep]}p{.09\reqwidth}p{.15\reqwidth}p{.28\reqwidth}p{.33\reqwidth}>{\columncolor{white}[\tabcolsep][0pt]}p{.15\reqwidth}@{}}\raggedright\arraybackslash  & \raggedright\arraybackslash  & \raggedright\arraybackslash Synthesize external evidence & \raggedright\arraybackslash Understand: summarize; Understand: compare; Analyze: organize & \raggedright\arraybackslash Evidence \\\end{tabular}}
\expandafter\edef\csname reqrowheight6\endcsname{\the\dimexpr\ht0+\dp0\relax}
\setbox0=\hbox{\begin{tabular}{@{}>{\columncolor{black!7}[0pt][\tabcolsep]}p{.09\reqwidth}p{.15\reqwidth}p{.28\reqwidth}p{.33\reqwidth}>{\columncolor{white}[\tabcolsep][0pt]}p{.15\reqwidth}@{}}\raggedright\arraybackslash  & \raggedright\arraybackslash  & \raggedright\arraybackslash Revise research questions & \raggedright\arraybackslash Apply: implement; Understand: explain; Analyze: organize & \raggedright\arraybackslash Questions \\\end{tabular}}
\expandafter\edef\csname reqrowheight7\endcsname{\the\dimexpr\ht0+\dp0\relax}
\setbox0=\hbox{\begin{tabular}{@{}>{\columncolor{black!7}[0pt][\tabcolsep]}p{.09\reqwidth}p{.15\reqwidth}p{.28\reqwidth}p{.33\reqwidth}>{\columncolor{white}[\tabcolsep][0pt]}p{.15\reqwidth}@{}}\raggedright\arraybackslash  & \raggedright\arraybackslash  & \raggedright\arraybackslash Assess method applicability & \raggedright\arraybackslash Evaluate: critique & \raggedright\arraybackslash Methods \\\end{tabular}}
\expandafter\edef\csname reqrowheight8\endcsname{\the\dimexpr\ht0+\dp0\relax}
\setbox0=\hbox{\begin{tabular}{@{}>{\columncolor{black!7}[0pt][\tabcolsep]}p{.09\reqwidth}p{.15\reqwidth}p{.28\reqwidth}p{.33\reqwidth}>{\columncolor{white}[\tabcolsep][0pt]}p{.15\reqwidth}@{}}\raggedright\arraybackslash  & \raggedright\arraybackslash  & \raggedright\arraybackslash Write a literature review report & \raggedright\arraybackslash Understand: summarize; Understand: explain; Analyze: organize & \raggedright\arraybackslash Documents \\\end{tabular}}
\expandafter\edef\csname reqrowheight9\endcsname{\the\dimexpr\ht0+\dp0\relax}
\setbox0=\hbox{\begin{tabular}{@{}>{\columncolor{black!7}[0pt][\tabcolsep]}p{.09\reqwidth}p{.15\reqwidth}p{.28\reqwidth}p{.33\reqwidth}>{\columncolor{white}[\tabcolsep][0pt]}p{.15\reqwidth}@{}}\raggedright\arraybackslash  & \raggedright\arraybackslash  & \raggedright\arraybackslash Generate candidate research questions & \raggedright\arraybackslash Create: generate; Understand: explain & \raggedright\arraybackslash Questions \\\end{tabular}}
\expandafter\edef\csname reqrowheight10\endcsname{\the\dimexpr\ht0+\dp0\relax}
\setbox0=\hbox{\begin{tabular}{@{}>{\columncolor{black!7}[0pt][\tabcolsep]}p{.09\reqwidth}p{.15\reqwidth}p{.28\reqwidth}p{.33\reqwidth}>{\columncolor{white}[\tabcolsep][0pt]}p{.15\reqwidth}@{}}\raggedright\arraybackslash  & \raggedright\arraybackslash  & \raggedright\arraybackslash Define core concepts & \raggedright\arraybackslash Understand: explain; Understand: exemplify; Analyze: organize & \raggedright\arraybackslash Questions \\\end{tabular}}
\expandafter\edef\csname reqrowheight11\endcsname{\the\dimexpr\ht0+\dp0\relax}
\setbox0=\hbox{\begin{tabular}{@{}>{\columncolor{black!7}[0pt][\tabcolsep]}p{.09\reqwidth}p{.15\reqwidth}p{.28\reqwidth}p{.33\reqwidth}>{\columncolor{white}[\tabcolsep][0pt]}p{.15\reqwidth}@{}}\raggedright\arraybackslash  & \raggedright\arraybackslash  & \raggedright\arraybackslash Define research variables & \raggedright\arraybackslash Understand: explain; Analyze: organize; Create: plan & \raggedright\arraybackslash Questions \\\end{tabular}}
\expandafter\edef\csname reqrowheight12\endcsname{\the\dimexpr\ht0+\dp0\relax}
\setbox0=\hbox{\begin{tabular}{@{}>{\columncolor{black!7}[0pt][\tabcolsep]}p{.09\reqwidth}p{.15\reqwidth}p{.28\reqwidth}p{.33\reqwidth}>{\columncolor{white}[\tabcolsep][0pt]}p{.15\reqwidth}@{}}\raggedright\arraybackslash  & \raggedright\arraybackslash  & \raggedright\arraybackslash Identify problem assumptions & \raggedright\arraybackslash Understand: infer; Analyze: attribute; Analyze: differentiate & \raggedright\arraybackslash Questions \\\end{tabular}}
\expandafter\edef\csname reqrowheight13\endcsname{\the\dimexpr\ht0+\dp0\relax}
\setbox0=\hbox{\begin{tabular}{@{}>{\columncolor{black!7}[0pt][\tabcolsep]}p{.09\reqwidth}p{.15\reqwidth}p{.28\reqwidth}p{.33\reqwidth}>{\columncolor{white}[\tabcolsep][0pt]}p{.15\reqwidth}@{}}\raggedright\arraybackslash  & \raggedright\arraybackslash  & \raggedright\arraybackslash Formulate research hypotheses & \raggedright\arraybackslash Understand: infer; Create: generate & \raggedright\arraybackslash Questions \\\end{tabular}}
\expandafter\edef\csname reqrowheight14\endcsname{\the\dimexpr\ht0+\dp0\relax}
\setbox0=\hbox{\begin{tabular}{@{}>{\columncolor{black!7}[0pt][\tabcolsep]}p{.09\reqwidth}p{.15\reqwidth}p{.28\reqwidth}p{.33\reqwidth}>{\columncolor{white}[\tabcolsep][0pt]}p{.15\reqwidth}@{}}\raggedright\arraybackslash  & \raggedright\arraybackslash  & \raggedright\arraybackslash Assess research novelty & \raggedright\arraybackslash Understand: compare; Analyze: differentiate; Evaluate: critique & \raggedright\arraybackslash Questions \\\end{tabular}}
\expandafter\edef\csname reqrowheight15\endcsname{\the\dimexpr\ht0+\dp0\relax}
\setbox0=\hbox{\begin{tabular}{@{}>{\columncolor{black!7}[0pt][\tabcolsep]}p{.09\reqwidth}p{.15\reqwidth}p{.28\reqwidth}p{.33\reqwidth}>{\columncolor{white}[\tabcolsep][0pt]}p{.15\reqwidth}@{}}\raggedright\arraybackslash  & \raggedright\arraybackslash  & \raggedright\arraybackslash Assess research feasibility & \raggedright\arraybackslash Analyze: differentiate; Evaluate: critique & \raggedright\arraybackslash Questions \\\end{tabular}}
\expandafter\edef\csname reqrowheight16\endcsname{\the\dimexpr\ht0+\dp0\relax}
\setbox0=\hbox{\begin{tabular}{@{}>{\columncolor{black!7}[0pt][\tabcolsep]}p{.09\reqwidth}p{.15\reqwidth}p{.28\reqwidth}p{.33\reqwidth}>{\columncolor{white}[\tabcolsep][0pt]}p{.15\reqwidth}@{}}\raggedright\arraybackslash  & \raggedright\arraybackslash  & \raggedright\arraybackslash Identify research ethics risks & \raggedright\arraybackslash Understand: classify; Evaluate: critique & \raggedright\arraybackslash Questions \\\end{tabular}}
\expandafter\edef\csname reqrowheight17\endcsname{\the\dimexpr\ht0+\dp0\relax}
\setbox0=\hbox{\begin{tabular}{@{}>{\columncolor{black!7}[0pt][\tabcolsep]}p{.09\reqwidth}p{.15\reqwidth}p{.28\reqwidth}p{.33\reqwidth}>{\columncolor{white}[\tabcolsep][0pt]}p{.15\reqwidth}@{}}\raggedright\arraybackslash  & \raggedright\arraybackslash  & \raggedright\arraybackslash Identify research constraints & \raggedright\arraybackslash Remember: recall; Understand: classify; Analyze: differentiate & \raggedright\arraybackslash Projects \\\end{tabular}}
\expandafter\edef\csname reqrowheight18\endcsname{\the\dimexpr\ht0+\dp0\relax}
\setbox0=\hbox{\begin{tabular}{@{}>{\columncolor{black!7}[0pt][\tabcolsep]}p{.09\reqwidth}p{.15\reqwidth}p{.28\reqwidth}p{.33\reqwidth}>{\columncolor{white}[\tabcolsep][0pt]}p{.15\reqwidth}@{}}\raggedright\arraybackslash  & \raggedright\arraybackslash  & \raggedright\arraybackslash Write a problem definition document & \raggedright\arraybackslash Understand: summarize; Understand: explain; Analyze: organize & \raggedright\arraybackslash Documents \\\end{tabular}}
\expandafter\edef\csname reqrowheight19\endcsname{\the\dimexpr\ht0+\dp0\relax}
\setbox0=\hbox{\begin{tabular}{@{}>{\columncolor{black!7}[0pt][\tabcolsep]}p{.09\reqwidth}p{.15\reqwidth}p{.28\reqwidth}p{.33\reqwidth}>{\columncolor{white}[\tabcolsep][0pt]}p{.15\reqwidth}@{}}\raggedright\arraybackslash  & \raggedright\arraybackslash  & \raggedright\arraybackslash Identify resource requirements & \raggedright\arraybackslash Understand: classify; Analyze: differentiate; Analyze: organize & \raggedright\arraybackslash Plans \\\end{tabular}}
\expandafter\edef\csname reqrowheight20\endcsname{\the\dimexpr\ht0+\dp0\relax}
\setbox0=\hbox{\begin{tabular}{@{}>{\columncolor{black!7}[0pt][\tabcolsep]}p{.09\reqwidth}p{.15\reqwidth}p{.28\reqwidth}p{.33\reqwidth}>{\columncolor{white}[\tabcolsep][0pt]}p{.15\reqwidth}@{}}\raggedright\arraybackslash  & \raggedright\arraybackslash  & \raggedright\arraybackslash Estimate data needs and costs & \raggedright\arraybackslash Apply: implement; Apply: execute; Understand: infer & \raggedright\arraybackslash Plans \\\end{tabular}}
\expandafter\edef\csname reqrowheight21\endcsname{\the\dimexpr\ht0+\dp0\relax}
\setbox0=\hbox{\begin{tabular}{@{}>{\columncolor{black!7}[0pt][\tabcolsep]}p{.09\reqwidth}p{.15\reqwidth}p{.28\reqwidth}p{.33\reqwidth}>{\columncolor{white}[\tabcolsep][0pt]}p{.15\reqwidth}@{}}\raggedright\arraybackslash  & \raggedright\arraybackslash  & \raggedright\arraybackslash Estimate compute needs and costs & \raggedright\arraybackslash Apply: implement; Apply: execute; Understand: infer & \raggedright\arraybackslash Plans \\\end{tabular}}
\expandafter\edef\csname reqrowheight22\endcsname{\the\dimexpr\ht0+\dp0\relax}
\setbox0=\hbox{\begin{tabular}{@{}>{\columncolor{black!7}[0pt][\tabcolsep]}p{.09\reqwidth}p{.15\reqwidth}p{.28\reqwidth}p{.33\reqwidth}>{\columncolor{white}[\tabcolsep][0pt]}p{.15\reqwidth}@{}}\raggedright\arraybackslash  & \raggedright\arraybackslash  & \raggedright\arraybackslash Estimate the research timeline & \raggedright\arraybackslash Apply: implement; Understand: infer & \raggedright\arraybackslash Plans \\\end{tabular}}
\expandafter\edef\csname reqrowheight23\endcsname{\the\dimexpr\ht0+\dp0\relax}
\setbox0=\hbox{\begin{tabular}{@{}>{\columncolor{black!7}[0pt][\tabcolsep]}p{.09\reqwidth}p{.15\reqwidth}p{.28\reqwidth}p{.33\reqwidth}>{\columncolor{white}[\tabcolsep][0pt]}p{.15\reqwidth}@{}}\raggedright\arraybackslash  & \raggedright\arraybackslash  & \raggedright\arraybackslash Plan research execution & \raggedright\arraybackslash Create: plan & \raggedright\arraybackslash Plans \\\end{tabular}}
\expandafter\edef\csname reqrowheight24\endcsname{\the\dimexpr\ht0+\dp0\relax}
\setbox0=\hbox{\begin{tabular}{@{}>{\columncolor{black!7}[0pt][\tabcolsep]}p{.09\reqwidth}p{.15\reqwidth}p{.28\reqwidth}p{.33\reqwidth}>{\columncolor{white}[\tabcolsep][0pt]}p{.15\reqwidth}@{}}\raggedright\arraybackslash  & \raggedright\arraybackslash  & \raggedright\arraybackslash Define experimental subjects and units of analysis & \raggedright\arraybackslash Understand: explain; Analyze: organize & \raggedright\arraybackslash Plans \\\end{tabular}}
\expandafter\edef\csname reqrowheight25\endcsname{\the\dimexpr\ht0+\dp0\relax}
\setbox0=\hbox{\begin{tabular}{@{}>{\columncolor{black!7}[0pt][\tabcolsep]}p{.09\reqwidth}p{.15\reqwidth}p{.28\reqwidth}p{.33\reqwidth}>{\columncolor{white}[\tabcolsep][0pt]}p{.15\reqwidth}@{}}\raggedright\arraybackslash  & \raggedright\arraybackslash  & \raggedright\arraybackslash Operationalize experimental variables & \raggedright\arraybackslash Understand: explain; Analyze: organize; Create: plan & \raggedright\arraybackslash Plans \\\end{tabular}}
\expandafter\edef\csname reqrowheight26\endcsname{\the\dimexpr\ht0+\dp0\relax}
\setbox0=\hbox{\begin{tabular}{@{}>{\columncolor{black!7}[0pt][\tabcolsep]}p{.09\reqwidth}p{.15\reqwidth}p{.28\reqwidth}p{.33\reqwidth}>{\columncolor{white}[\tabcolsep][0pt]}p{.15\reqwidth}@{}}\raggedright\arraybackslash  & \raggedright\arraybackslash  & \raggedright\arraybackslash Design experimental controls & \raggedright\arraybackslash Analyze: differentiate; Create: plan & \raggedright\arraybackslash Plans \\\end{tabular}}
\expandafter\edef\csname reqrowheight27\endcsname{\the\dimexpr\ht0+\dp0\relax}
\setbox0=\hbox{\begin{tabular}{@{}>{\columncolor{black!7}[0pt][\tabcolsep]}p{.09\reqwidth}p{.15\reqwidth}p{.28\reqwidth}p{.33\reqwidth}>{\columncolor{white}[\tabcolsep][0pt]}p{.15\reqwidth}@{}}\raggedright\arraybackslash  & \raggedright\arraybackslash  & \raggedright\arraybackslash Design ablation experiments & \raggedright\arraybackslash Analyze: differentiate; Create: plan & \raggedright\arraybackslash Plans \\\end{tabular}}
\expandafter\edef\csname reqrowheight28\endcsname{\the\dimexpr\ht0+\dp0\relax}
\setbox0=\hbox{\begin{tabular}{@{}>{\columncolor{black!7}[0pt][\tabcolsep]}p{.09\reqwidth}p{.15\reqwidth}p{.28\reqwidth}p{.33\reqwidth}>{\columncolor{white}[\tabcolsep][0pt]}p{.15\reqwidth}@{}}\raggedright\arraybackslash  & \raggedright\arraybackslash  & \raggedright\arraybackslash Design repeated experiments & \raggedright\arraybackslash Analyze: organize; Create: plan & \raggedright\arraybackslash Plans \\\end{tabular}}
\expandafter\edef\csname reqrowheight29\endcsname{\the\dimexpr\ht0+\dp0\relax}
\setbox0=\hbox{\begin{tabular}{@{}>{\columncolor{black!7}[0pt][\tabcolsep]}p{.09\reqwidth}p{.15\reqwidth}p{.28\reqwidth}p{.33\reqwidth}>{\columncolor{white}[\tabcolsep][0pt]}p{.15\reqwidth}@{}}\raggedright\arraybackslash  & \raggedright\arraybackslash  & \raggedright\arraybackslash Define evaluation metrics & \raggedright\arraybackslash Analyze: organize; Create: plan & \raggedright\arraybackslash Plans \\\end{tabular}}
\expandafter\edef\csname reqrowheight30\endcsname{\the\dimexpr\ht0+\dp0\relax}
\setbox0=\hbox{\begin{tabular}{@{}>{\columncolor{black!7}[0pt][\tabcolsep]}p{.09\reqwidth}p{.15\reqwidth}p{.28\reqwidth}p{.33\reqwidth}>{\columncolor{white}[\tabcolsep][0pt]}p{.15\reqwidth}@{}}\raggedright\arraybackslash  & \raggedright\arraybackslash  & \raggedright\arraybackslash Define research acceptance criteria & \raggedright\arraybackslash Analyze: organize; Create: plan & \raggedright\arraybackslash Plans \\\end{tabular}}
\expandafter\edef\csname reqrowheight31\endcsname{\the\dimexpr\ht0+\dp0\relax}
\setbox0=\hbox{\begin{tabular}{@{}>{\columncolor{black!7}[0pt][\tabcolsep]}p{.09\reqwidth}p{.15\reqwidth}p{.28\reqwidth}p{.33\reqwidth}>{\columncolor{white}[\tabcolsep][0pt]}p{.15\reqwidth}@{}}\raggedright\arraybackslash  & \raggedright\arraybackslash  & \raggedright\arraybackslash Design sensitivity analyses & \raggedright\arraybackslash Understand: compare; Analyze: differentiate; Create: plan & \raggedright\arraybackslash Plans \\\end{tabular}}
\expandafter\edef\csname reqrowheight32\endcsname{\the\dimexpr\ht0+\dp0\relax}
\setbox0=\hbox{\begin{tabular}{@{}>{\columncolor{black!7}[0pt][\tabcolsep]}p{.09\reqwidth}p{.15\reqwidth}p{.28\reqwidth}p{.33\reqwidth}>{\columncolor{white}[\tabcolsep][0pt]}p{.15\reqwidth}@{}}\raggedright\arraybackslash  & \raggedright\arraybackslash  & \raggedright\arraybackslash Design experimental procedures & \raggedright\arraybackslash Analyze: organize; Create: plan & \raggedright\arraybackslash Plans \\\end{tabular}}
\expandafter\edef\csname reqrowheight33\endcsname{\the\dimexpr\ht0+\dp0\relax}
\setbox0=\hbox{\begin{tabular}{@{}>{\columncolor{black!7}[0pt][\tabcolsep]}p{.09\reqwidth}p{.15\reqwidth}p{.28\reqwidth}p{.33\reqwidth}>{\columncolor{white}[\tabcolsep][0pt]}p{.15\reqwidth}@{}}\raggedright\arraybackslash  & \raggedright\arraybackslash  & \raggedright\arraybackslash Design experiment recording standards & \raggedright\arraybackslash Analyze: organize; Create: plan & \raggedright\arraybackslash Plans \\\end{tabular}}
\expandafter\edef\csname reqrowheight34\endcsname{\the\dimexpr\ht0+\dp0\relax}
\setbox0=\hbox{\begin{tabular}{@{}>{\columncolor{black!7}[0pt][\tabcolsep]}p{.09\reqwidth}p{.15\reqwidth}p{.28\reqwidth}p{.33\reqwidth}>{\columncolor{white}[\tabcolsep][0pt]}p{.15\reqwidth}@{}}\raggedright\arraybackslash  & \raggedright\arraybackslash  & \raggedright\arraybackslash Design experiment stopping rules & \raggedright\arraybackslash Analyze: organize; Evaluate: check; Create: plan & \raggedright\arraybackslash Plans \\\end{tabular}}
\expandafter\edef\csname reqrowheight35\endcsname{\the\dimexpr\ht0+\dp0\relax}
\setbox0=\hbox{\begin{tabular}{@{}>{\columncolor{black!7}[0pt][\tabcolsep]}p{.09\reqwidth}p{.15\reqwidth}p{.28\reqwidth}p{.33\reqwidth}>{\columncolor{white}[\tabcolsep][0pt]}p{.15\reqwidth}@{}}\raggedright\arraybackslash  & \raggedright\arraybackslash  & \raggedright\arraybackslash Identify threats to experimental validity & \raggedright\arraybackslash Understand: classify; Understand: infer; Analyze: differentiate & \raggedright\arraybackslash Plans \\\end{tabular}}
\expandafter\edef\csname reqrowheight36\endcsname{\the\dimexpr\ht0+\dp0\relax}
\setbox0=\hbox{\begin{tabular}{@{}>{\columncolor{black!7}[0pt][\tabcolsep]}p{.09\reqwidth}p{.15\reqwidth}p{.28\reqwidth}p{.33\reqwidth}>{\columncolor{white}[\tabcolsep][0pt]}p{.15\reqwidth}@{}}\raggedright\arraybackslash  & \raggedright\arraybackslash  & \raggedright\arraybackslash Assess experiment design feasibility & \raggedright\arraybackslash Analyze: differentiate; Evaluate: critique & \raggedright\arraybackslash Plans \\\end{tabular}}
\expandafter\edef\csname reqrowheight37\endcsname{\the\dimexpr\ht0+\dp0\relax}
\setbox0=\hbox{\begin{tabular}{@{}>{\columncolor{black!7}[0pt][\tabcolsep]}p{.09\reqwidth}p{.15\reqwidth}p{.28\reqwidth}p{.33\reqwidth}>{\columncolor{white}[\tabcolsep][0pt]}p{.15\reqwidth}@{}}\raggedright\arraybackslash  & \raggedright\arraybackslash  & \raggedright\arraybackslash Verify experiment designs & \raggedright\arraybackslash Evaluate: check; Analyze: organize & \raggedright\arraybackslash Plans \\\end{tabular}}
\expandafter\edef\csname reqrowheight38\endcsname{\the\dimexpr\ht0+\dp0\relax}
\setbox0=\hbox{\begin{tabular}{@{}>{\columncolor{black!7}[0pt][\tabcolsep]}p{.09\reqwidth}p{.15\reqwidth}p{.28\reqwidth}p{.33\reqwidth}>{\columncolor{white}[\tabcolsep][0pt]}p{.15\reqwidth}@{}}\raggedright\arraybackslash  & \raggedright\arraybackslash  & \raggedright\arraybackslash Write an experiment design document & \raggedright\arraybackslash Understand: summarize; Understand: explain; Analyze: organize & \raggedright\arraybackslash Documents \\\end{tabular}}
\expandafter\edef\csname reqrowheight39\endcsname{\the\dimexpr\ht0+\dp0\relax}
\setbox0=\hbox{\begin{tabular}{@{}>{\columncolor{black!7}[0pt][\tabcolsep]}p{.09\reqwidth}p{.15\reqwidth}p{.28\reqwidth}p{.33\reqwidth}>{\columncolor{white}[\tabcolsep][0pt]}p{.15\reqwidth}@{}}\raggedright\arraybackslash  & \raggedright\arraybackslash  & \raggedright\arraybackslash Select experiment evaluation metrics & \raggedright\arraybackslash Analyze: differentiate; Evaluate: critique & \raggedright\arraybackslash Plans \\\end{tabular}}
\expandafter\edef\csname reqrowheight40\endcsname{\the\dimexpr\ht0+\dp0\relax}
\setbox0=\hbox{\begin{tabular}{@{}>{\columncolor{black!7}[0pt][\tabcolsep]}p{.09\reqwidth}p{.15\reqwidth}p{.28\reqwidth}p{.33\reqwidth}>{\columncolor{white}[\tabcolsep][0pt]}p{.15\reqwidth}@{}}\raggedright\arraybackslash  & \raggedright\arraybackslash  & \raggedright\arraybackslash Design method evaluation protocols & \raggedright\arraybackslash Create: plan & \raggedright\arraybackslash Plans \\\end{tabular}}
\expandafter\edef\csname reqrowheight41\endcsname{\the\dimexpr\ht0+\dp0\relax}
\setbox0=\hbox{\begin{tabular}{@{}>{\columncolor{black!7}[0pt][\tabcolsep]}p{.09\reqwidth}p{.15\reqwidth}p{.28\reqwidth}p{.33\reqwidth}>{\columncolor{white}[\tabcolsep][0pt]}p{.15\reqwidth}@{}}\raggedright\arraybackslash  & \raggedright\arraybackslash  & \raggedright\arraybackslash Implement experiment evaluation metrics & \raggedright\arraybackslash Apply: implement & \raggedright\arraybackslash Code \\\end{tabular}}
\expandafter\edef\csname reqrowheight42\endcsname{\the\dimexpr\ht0+\dp0\relax}
\setbox0=\hbox{\begin{tabular}{@{}>{\columncolor{black!7}[0pt][\tabcolsep]}p{.09\reqwidth}p{.15\reqwidth}p{.28\reqwidth}p{.33\reqwidth}>{\columncolor{white}[\tabcolsep][0pt]}p{.15\reqwidth}@{}}\raggedright\arraybackslash  & \raggedright\arraybackslash  & \raggedright\arraybackslash Configure controlled experiments & \raggedright\arraybackslash Analyze: organize; Apply: implement & \raggedright\arraybackslash Experiments \\\end{tabular}}
\expandafter\edef\csname reqrowheight43\endcsname{\the\dimexpr\ht0+\dp0\relax}
\setbox0=\hbox{\begin{tabular}{@{}>{\columncolor{black!7}[0pt][\tabcolsep]}p{.09\reqwidth}p{.15\reqwidth}p{.28\reqwidth}p{.33\reqwidth}>{\columncolor{white}[\tabcolsep][0pt]}p{.15\reqwidth}@{}}\raggedright\arraybackslash  & \raggedright\arraybackslash  & \raggedright\arraybackslash Configure cross-model experiments & \raggedright\arraybackslash Analyze: organize; Apply: implement & \raggedright\arraybackslash Experiments \\\end{tabular}}
\expandafter\edef\csname reqrowheight44\endcsname{\the\dimexpr\ht0+\dp0\relax}
\setbox0=\hbox{\begin{tabular}{@{}>{\columncolor{black!7}[0pt][\tabcolsep]}p{.09\reqwidth}p{.15\reqwidth}p{.28\reqwidth}p{.33\reqwidth}>{\columncolor{white}[\tabcolsep][0pt]}p{.15\reqwidth}@{}}\raggedright\arraybackslash  & \raggedright\arraybackslash  & \raggedright\arraybackslash Install project dependencies & \raggedright\arraybackslash Apply: implement & \raggedright\arraybackslash Environment \\\end{tabular}}
\expandafter\edef\csname reqrowheight45\endcsname{\the\dimexpr\ht0+\dp0\relax}
\setbox0=\hbox{\begin{tabular}{@{}>{\columncolor{black!7}[0pt][\tabcolsep]}p{.09\reqwidth}p{.15\reqwidth}p{.28\reqwidth}p{.33\reqwidth}>{\columncolor{white}[\tabcolsep][0pt]}p{.15\reqwidth}@{}}\raggedright\arraybackslash  & \raggedright\arraybackslash  & \raggedright\arraybackslash Resolve dependency conflicts & \raggedright\arraybackslash Apply: implement; Evaluate: check & \raggedright\arraybackslash Environment \\\end{tabular}}
\expandafter\edef\csname reqrowheight46\endcsname{\the\dimexpr\ht0+\dp0\relax}
\setbox0=\hbox{\begin{tabular}{@{}>{\columncolor{black!7}[0pt][\tabcolsep]}p{.09\reqwidth}p{.15\reqwidth}p{.28\reqwidth}p{.33\reqwidth}>{\columncolor{white}[\tabcolsep][0pt]}p{.15\reqwidth}@{}}\raggedright\arraybackslash  & \raggedright\arraybackslash  & \raggedright\arraybackslash Verify the computing environment & \raggedright\arraybackslash Evaluate: check & \raggedright\arraybackslash Environment \\\end{tabular}}
\expandafter\edef\csname reqrowheight47\endcsname{\the\dimexpr\ht0+\dp0\relax}
\setbox0=\hbox{\begin{tabular}{@{}>{\columncolor{black!7}[0pt][\tabcolsep]}p{.09\reqwidth}p{.15\reqwidth}p{.28\reqwidth}p{.33\reqwidth}>{\columncolor{white}[\tabcolsep][0pt]}p{.15\reqwidth}@{}}\raggedright\arraybackslash  & \raggedright\arraybackslash  & \raggedright\arraybackslash Acquire model assets & \raggedright\arraybackslash Apply: execute & \raggedright\arraybackslash Assets \\\end{tabular}}
\expandafter\edef\csname reqrowheight48\endcsname{\the\dimexpr\ht0+\dp0\relax}
\setbox0=\hbox{\begin{tabular}{@{}>{\columncolor{black!7}[0pt][\tabcolsep]}p{.09\reqwidth}p{.15\reqwidth}p{.28\reqwidth}p{.33\reqwidth}>{\columncolor{white}[\tabcolsep][0pt]}p{.15\reqwidth}@{}}\raggedright\arraybackslash  & \raggedright\arraybackslash  & \raggedright\arraybackslash Acquire research data & \raggedright\arraybackslash Apply: execute & \raggedright\arraybackslash Data \\\end{tabular}}
\expandafter\edef\csname reqrowheight49\endcsname{\the\dimexpr\ht0+\dp0\relax}
\setbox0=\hbox{\begin{tabular}{@{}>{\columncolor{black!7}[0pt][\tabcolsep]}p{.09\reqwidth}p{.15\reqwidth}p{.28\reqwidth}p{.33\reqwidth}>{\columncolor{white}[\tabcolsep][0pt]}p{.15\reqwidth}@{}}\raggedright\arraybackslash  & \raggedright\arraybackslash  & \raggedright\arraybackslash Verify research data identity & \raggedright\arraybackslash Evaluate: check & \raggedright\arraybackslash Data \\\end{tabular}}
\expandafter\edef\csname reqrowheight50\endcsname{\the\dimexpr\ht0+\dp0\relax}
\setbox0=\hbox{\begin{tabular}{@{}>{\columncolor{black!7}[0pt][\tabcolsep]}p{.09\reqwidth}p{.15\reqwidth}p{.28\reqwidth}p{.33\reqwidth}>{\columncolor{white}[\tabcolsep][0pt]}p{.15\reqwidth}@{}}\raggedright\arraybackslash  & \raggedright\arraybackslash  & \raggedright\arraybackslash Acquire compute resources & \raggedright\arraybackslash Evaluate: critique; Apply: implement & \raggedright\arraybackslash Environment \\\end{tabular}}
\expandafter\edef\csname reqrowheight51\endcsname{\the\dimexpr\ht0+\dp0\relax}
\setbox0=\hbox{\begin{tabular}{@{}>{\columncolor{black!7}[0pt][\tabcolsep]}p{.09\reqwidth}p{.15\reqwidth}p{.28\reqwidth}p{.33\reqwidth}>{\columncolor{white}[\tabcolsep][0pt]}p{.15\reqwidth}@{}}\raggedright\arraybackslash  & \raggedright\arraybackslash  & \raggedright\arraybackslash Analyze data structures & \raggedright\arraybackslash Analyze: organize & \raggedright\arraybackslash Data \\\end{tabular}}
\expandafter\edef\csname reqrowheight52\endcsname{\the\dimexpr\ht0+\dp0\relax}
\setbox0=\hbox{\begin{tabular}{@{}>{\columncolor{black!7}[0pt][\tabcolsep]}p{.09\reqwidth}p{.15\reqwidth}p{.28\reqwidth}p{.33\reqwidth}>{\columncolor{white}[\tabcolsep][0pt]}p{.15\reqwidth}@{}}\raggedright\arraybackslash  & \raggedright\arraybackslash  & \raggedright\arraybackslash Assess data quality & \raggedright\arraybackslash Evaluate: critique & \raggedright\arraybackslash Data \\\end{tabular}}
\expandafter\edef\csname reqrowheight53\endcsname{\the\dimexpr\ht0+\dp0\relax}
\setbox0=\hbox{\begin{tabular}{@{}>{\columncolor{black!7}[0pt][\tabcolsep]}p{.09\reqwidth}p{.15\reqwidth}p{.28\reqwidth}p{.33\reqwidth}>{\columncolor{white}[\tabcolsep][0pt]}p{.15\reqwidth}@{}}\raggedright\arraybackslash  & \raggedright\arraybackslash  & \raggedright\arraybackslash Assess statistical power & \raggedright\arraybackslash Evaluate: critique & \raggedright\arraybackslash Plans \\\end{tabular}}
\expandafter\edef\csname reqrowheight54\endcsname{\the\dimexpr\ht0+\dp0\relax}
\setbox0=\hbox{\begin{tabular}{@{}>{\columncolor{black!7}[0pt][\tabcolsep]}p{.09\reqwidth}p{.15\reqwidth}p{.28\reqwidth}p{.33\reqwidth}>{\columncolor{white}[\tabcolsep][0pt]}p{.15\reqwidth}@{}}\raggedright\arraybackslash  & \raggedright\arraybackslash  & \raggedright\arraybackslash Convert data formats & \raggedright\arraybackslash Understand: explain; Apply: execute & \raggedright\arraybackslash Data \\\end{tabular}}
\expandafter\edef\csname reqrowheight55\endcsname{\the\dimexpr\ht0+\dp0\relax}
\setbox0=\hbox{\begin{tabular}{@{}>{\columncolor{black!7}[0pt][\tabcolsep]}p{.09\reqwidth}p{.15\reqwidth}p{.28\reqwidth}p{.33\reqwidth}>{\columncolor{white}[\tabcolsep][0pt]}p{.15\reqwidth}@{}}\raggedright\arraybackslash  & \raggedright\arraybackslash  & \raggedright\arraybackslash Handle missing values & \raggedright\arraybackslash Apply: implement; Evaluate: check & \raggedright\arraybackslash Data \\\end{tabular}}
\expandafter\edef\csname reqrowheight56\endcsname{\the\dimexpr\ht0+\dp0\relax}
\setbox0=\hbox{\begin{tabular}{@{}>{\columncolor{black!7}[0pt][\tabcolsep]}p{.09\reqwidth}p{.15\reqwidth}p{.28\reqwidth}p{.33\reqwidth}>{\columncolor{white}[\tabcolsep][0pt]}p{.15\reqwidth}@{}}\raggedright\arraybackslash  & \raggedright\arraybackslash  & \raggedright\arraybackslash Handle outliers & \raggedright\arraybackslash Apply: implement; Evaluate: check & \raggedright\arraybackslash Data \\\end{tabular}}
\expandafter\edef\csname reqrowheight57\endcsname{\the\dimexpr\ht0+\dp0\relax}
\setbox0=\hbox{\begin{tabular}{@{}>{\columncolor{black!7}[0pt][\tabcolsep]}p{.09\reqwidth}p{.15\reqwidth}p{.28\reqwidth}p{.33\reqwidth}>{\columncolor{white}[\tabcolsep][0pt]}p{.15\reqwidth}@{}}\raggedright\arraybackslash  & \raggedright\arraybackslash  & \raggedright\arraybackslash Create training, validation, and test splits & \raggedright\arraybackslash Analyze: organize & \raggedright\arraybackslash Data \\\end{tabular}}
\expandafter\edef\csname reqrowheight58\endcsname{\the\dimexpr\ht0+\dp0\relax}
\setbox0=\hbox{\begin{tabular}{@{}>{\columncolor{black!7}[0pt][\tabcolsep]}p{.09\reqwidth}p{.15\reqwidth}p{.28\reqwidth}p{.33\reqwidth}>{\columncolor{white}[\tabcolsep][0pt]}p{.15\reqwidth}@{}}\raggedright\arraybackslash  & \raggedright\arraybackslash  & \raggedright\arraybackslash Extract data features & \raggedright\arraybackslash Analyze: differentiate & \raggedright\arraybackslash Data \\\end{tabular}}
\expandafter\edef\csname reqrowheight59\endcsname{\the\dimexpr\ht0+\dp0\relax}
\setbox0=\hbox{\begin{tabular}{@{}>{\columncolor{black!7}[0pt][\tabcolsep]}p{.09\reqwidth}p{.15\reqwidth}p{.28\reqwidth}p{.33\reqwidth}>{\columncolor{white}[\tabcolsep][0pt]}p{.15\reqwidth}@{}}\raggedright\arraybackslash  & \raggedright\arraybackslash  & \raggedright\arraybackslash Compare features across groups & \raggedright\arraybackslash Understand: compare & \raggedright\arraybackslash Data \\\end{tabular}}
\expandafter\edef\csname reqrowheight60\endcsname{\the\dimexpr\ht0+\dp0\relax}
\setbox0=\hbox{\begin{tabular}{@{}>{\columncolor{black!7}[0pt][\tabcolsep]}p{.09\reqwidth}p{.15\reqwidth}p{.28\reqwidth}p{.33\reqwidth}>{\columncolor{white}[\tabcolsep][0pt]}p{.15\reqwidth}@{}}\raggedright\arraybackslash  & \raggedright\arraybackslash  & \raggedright\arraybackslash Analyze code implementations & \raggedright\arraybackslash Analyze: organize & \raggedright\arraybackslash Code \\\end{tabular}}
\expandafter\edef\csname reqrowheight61\endcsname{\the\dimexpr\ht0+\dp0\relax}
\setbox0=\hbox{\begin{tabular}{@{}>{\columncolor{black!7}[0pt][\tabcolsep]}p{.09\reqwidth}p{.15\reqwidth}p{.28\reqwidth}p{.33\reqwidth}>{\columncolor{white}[\tabcolsep][0pt]}p{.15\reqwidth}@{}}\raggedright\arraybackslash  & \raggedright\arraybackslash  & \raggedright\arraybackslash Verify code modules & \raggedright\arraybackslash Evaluate: check & \raggedright\arraybackslash Code \\\end{tabular}}
\expandafter\edef\csname reqrowheight62\endcsname{\the\dimexpr\ht0+\dp0\relax}
\setbox0=\hbox{\begin{tabular}{@{}>{\columncolor{black!7}[0pt][\tabcolsep]}p{.09\reqwidth}p{.15\reqwidth}p{.28\reqwidth}p{.33\reqwidth}>{\columncolor{white}[\tabcolsep][0pt]}p{.15\reqwidth}@{}}\raggedright\arraybackslash  & \raggedright\arraybackslash  & \raggedright\arraybackslash Verify code completeness & \raggedright\arraybackslash Evaluate: check & \raggedright\arraybackslash Code \\\end{tabular}}
\expandafter\edef\csname reqrowheight63\endcsname{\the\dimexpr\ht0+\dp0\relax}
\setbox0=\hbox{\begin{tabular}{@{}>{\columncolor{black!7}[0pt][\tabcolsep]}p{.09\reqwidth}p{.15\reqwidth}p{.28\reqwidth}p{.33\reqwidth}>{\columncolor{white}[\tabcolsep][0pt]}p{.15\reqwidth}@{}}\raggedright\arraybackslash  & \raggedright\arraybackslash  & \raggedright\arraybackslash Derive research methods & \raggedright\arraybackslash Create: plan & \raggedright\arraybackslash Methods \\\end{tabular}}
\expandafter\edef\csname reqrowheight64\endcsname{\the\dimexpr\ht0+\dp0\relax}
\setbox0=\hbox{\begin{tabular}{@{}>{\columncolor{black!7}[0pt][\tabcolsep]}p{.09\reqwidth}p{.15\reqwidth}p{.28\reqwidth}p{.33\reqwidth}>{\columncolor{white}[\tabcolsep][0pt]}p{.15\reqwidth}@{}}\raggedright\arraybackslash  & \raggedright\arraybackslash  & \raggedright\arraybackslash Adapt research methods & \raggedright\arraybackslash Apply: implement & \raggedright\arraybackslash Methods \\\end{tabular}}
\expandafter\edef\csname reqrowheight65\endcsname{\the\dimexpr\ht0+\dp0\relax}
\setbox0=\hbox{\begin{tabular}{@{}>{\columncolor{black!7}[0pt][\tabcolsep]}p{.09\reqwidth}p{.15\reqwidth}p{.28\reqwidth}p{.33\reqwidth}>{\columncolor{white}[\tabcolsep][0pt]}p{.15\reqwidth}@{}}\raggedright\arraybackslash  & \raggedright\arraybackslash  & \raggedright\arraybackslash Implement experiment scaffolding & \raggedright\arraybackslash Apply: implement & \raggedright\arraybackslash Code \\\end{tabular}}
\expandafter\edef\csname reqrowheight66\endcsname{\the\dimexpr\ht0+\dp0\relax}
\setbox0=\hbox{\begin{tabular}{@{}>{\columncolor{black!7}[0pt][\tabcolsep]}p{.09\reqwidth}p{.15\reqwidth}p{.28\reqwidth}p{.33\reqwidth}>{\columncolor{white}[\tabcolsep][0pt]}p{.15\reqwidth}@{}}\raggedright\arraybackslash  & \raggedright\arraybackslash  & \raggedright\arraybackslash Design model architectures & \raggedright\arraybackslash Create: plan & \raggedright\arraybackslash Methods \\\end{tabular}}
\expandafter\edef\csname reqrowheight67\endcsname{\the\dimexpr\ht0+\dp0\relax}
\setbox0=\hbox{\begin{tabular}{@{}>{\columncolor{black!7}[0pt][\tabcolsep]}p{.09\reqwidth}p{.15\reqwidth}p{.28\reqwidth}p{.33\reqwidth}>{\columncolor{white}[\tabcolsep][0pt]}p{.15\reqwidth}@{}}\raggedright\arraybackslash  & \raggedright\arraybackslash  & \raggedright\arraybackslash Design algorithm mechanisms & \raggedright\arraybackslash Create: plan & \raggedright\arraybackslash Methods \\\end{tabular}}
\expandafter\edef\csname reqrowheight68\endcsname{\the\dimexpr\ht0+\dp0\relax}
\setbox0=\hbox{\begin{tabular}{@{}>{\columncolor{black!7}[0pt][\tabcolsep]}p{.09\reqwidth}p{.15\reqwidth}p{.28\reqwidth}p{.33\reqwidth}>{\columncolor{white}[\tabcolsep][0pt]}p{.15\reqwidth}@{}}\raggedright\arraybackslash  & \raggedright\arraybackslash  & \raggedright\arraybackslash Design loss functions & \raggedright\arraybackslash Create: plan & \raggedright\arraybackslash Methods \\\end{tabular}}
\expandafter\edef\csname reqrowheight69\endcsname{\the\dimexpr\ht0+\dp0\relax}
\setbox0=\hbox{\begin{tabular}{@{}>{\columncolor{black!7}[0pt][\tabcolsep]}p{.09\reqwidth}p{.15\reqwidth}p{.28\reqwidth}p{.33\reqwidth}>{\columncolor{white}[\tabcolsep][0pt]}p{.15\reqwidth}@{}}\raggedright\arraybackslash  & \raggedright\arraybackslash  & \raggedright\arraybackslash Design code implementations and boundaries & \raggedright\arraybackslash Create: plan & \raggedright\arraybackslash Code \\\end{tabular}}
\expandafter\edef\csname reqrowheight70\endcsname{\the\dimexpr\ht0+\dp0\relax}
\setbox0=\hbox{\begin{tabular}{@{}>{\columncolor{black!7}[0pt][\tabcolsep]}p{.09\reqwidth}p{.15\reqwidth}p{.28\reqwidth}p{.33\reqwidth}>{\columncolor{white}[\tabcolsep][0pt]}p{.15\reqwidth}@{}}\raggedright\arraybackslash  & \raggedright\arraybackslash  & \raggedright\arraybackslash Add experiment instrumentation logs & \raggedright\arraybackslash Apply: implement & \raggedright\arraybackslash Code \\\end{tabular}}
\expandafter\edef\csname reqrowheight71\endcsname{\the\dimexpr\ht0+\dp0\relax}
\setbox0=\hbox{\begin{tabular}{@{}>{\columncolor{black!7}[0pt][\tabcolsep]}p{.09\reqwidth}p{.15\reqwidth}p{.28\reqwidth}p{.33\reqwidth}>{\columncolor{white}[\tabcolsep][0pt]}p{.15\reqwidth}@{}}\raggedright\arraybackslash  & \raggedright\arraybackslash  & \raggedright\arraybackslash Implement algorithm modules & \raggedright\arraybackslash Apply: implement & \raggedright\arraybackslash Code \\\end{tabular}}
\expandafter\edef\csname reqrowheight72\endcsname{\the\dimexpr\ht0+\dp0\relax}
\setbox0=\hbox{\begin{tabular}{@{}>{\columncolor{black!7}[0pt][\tabcolsep]}p{.09\reqwidth}p{.15\reqwidth}p{.28\reqwidth}p{.33\reqwidth}>{\columncolor{white}[\tabcolsep][0pt]}p{.15\reqwidth}@{}}\raggedright\arraybackslash  & \raggedright\arraybackslash  & \raggedright\arraybackslash Implement algorithm variants & \raggedright\arraybackslash Apply: implement & \raggedright\arraybackslash Code \\\end{tabular}}
\expandafter\edef\csname reqrowheight73\endcsname{\the\dimexpr\ht0+\dp0\relax}
\setbox0=\hbox{\begin{tabular}{@{}>{\columncolor{black!7}[0pt][\tabcolsep]}p{.09\reqwidth}p{.15\reqwidth}p{.28\reqwidth}p{.33\reqwidth}>{\columncolor{white}[\tabcolsep][0pt]}p{.15\reqwidth}@{}}\raggedright\arraybackslash  & \raggedright\arraybackslash  & \raggedright\arraybackslash Integrate open-source components & \raggedright\arraybackslash Analyze: organize; Apply: implement & \raggedright\arraybackslash Code \\\end{tabular}}
\expandafter\edef\csname reqrowheight74\endcsname{\the\dimexpr\ht0+\dp0\relax}
\setbox0=\hbox{\begin{tabular}{@{}>{\columncolor{black!7}[0pt][\tabcolsep]}p{.09\reqwidth}p{.15\reqwidth}p{.28\reqwidth}p{.33\reqwidth}>{\columncolor{white}[\tabcolsep][0pt]}p{.15\reqwidth}@{}}\raggedright\arraybackslash  & \raggedright\arraybackslash  & \raggedright\arraybackslash Verify algorithm implementation equivalence & \raggedright\arraybackslash Evaluate: check & \raggedright\arraybackslash Code \\\end{tabular}}
\expandafter\edef\csname reqrowheight75\endcsname{\the\dimexpr\ht0+\dp0\relax}
\setbox0=\hbox{\begin{tabular}{@{}>{\columncolor{black!7}[0pt][\tabcolsep]}p{.09\reqwidth}p{.15\reqwidth}p{.28\reqwidth}p{.33\reqwidth}>{\columncolor{white}[\tabcolsep][0pt]}p{.15\reqwidth}@{}}\raggedright\arraybackslash  & \raggedright\arraybackslash  & \raggedright\arraybackslash Manage code version changes & \raggedright\arraybackslash Apply: execute; Analyze: organize & \raggedright\arraybackslash Code \\\end{tabular}}
\expandafter\edef\csname reqrowheight76\endcsname{\the\dimexpr\ht0+\dp0\relax}
\setbox0=\hbox{\begin{tabular}{@{}>{\columncolor{black!7}[0pt][\tabcolsep]}p{.09\reqwidth}p{.15\reqwidth}p{.28\reqwidth}p{.33\reqwidth}>{\columncolor{white}[\tabcolsep][0pt]}p{.15\reqwidth}@{}}\raggedright\arraybackslash  & \raggedright\arraybackslash  & \raggedright\arraybackslash Modify code across modules & \raggedright\arraybackslash Analyze: organize; Apply: implement & \raggedright\arraybackslash Code \\\end{tabular}}
\expandafter\edef\csname reqrowheight77\endcsname{\the\dimexpr\ht0+\dp0\relax}
\setbox0=\hbox{\begin{tabular}{@{}>{\columncolor{black!7}[0pt][\tabcolsep]}p{.09\reqwidth}p{.15\reqwidth}p{.28\reqwidth}p{.33\reqwidth}>{\columncolor{white}[\tabcolsep][0pt]}p{.15\reqwidth}@{}}\raggedright\arraybackslash  & \raggedright\arraybackslash  & \raggedright\arraybackslash Refactor code & \raggedright\arraybackslash Analyze: organize; Apply: implement & \raggedright\arraybackslash Code \\\end{tabular}}
\expandafter\edef\csname reqrowheight78\endcsname{\the\dimexpr\ht0+\dp0\relax}
\setbox0=\hbox{\begin{tabular}{@{}>{\columncolor{black!7}[0pt][\tabcolsep]}p{.09\reqwidth}p{.15\reqwidth}p{.28\reqwidth}p{.33\reqwidth}>{\columncolor{white}[\tabcolsep][0pt]}p{.15\reqwidth}@{}}\raggedright\arraybackslash  & \raggedright\arraybackslash  & \raggedright\arraybackslash Verify code changes & \raggedright\arraybackslash Evaluate: check & \raggedright\arraybackslash Code \\\end{tabular}}
\expandafter\edef\csname reqrowheight79\endcsname{\the\dimexpr\ht0+\dp0\relax}
\setbox0=\hbox{\begin{tabular}{@{}>{\columncolor{black!7}[0pt][\tabcolsep]}p{.09\reqwidth}p{.15\reqwidth}p{.28\reqwidth}p{.33\reqwidth}>{\columncolor{white}[\tabcolsep][0pt]}p{.15\reqwidth}@{}}\raggedright\arraybackslash  & \raggedright\arraybackslash  & \raggedright\arraybackslash Write unit tests & \raggedright\arraybackslash Analyze: differentiate; Apply: implement & \raggedright\arraybackslash Code \\\end{tabular}}
\expandafter\edef\csname reqrowheight80\endcsname{\the\dimexpr\ht0+\dp0\relax}
\setbox0=\hbox{\begin{tabular}{@{}>{\columncolor{black!7}[0pt][\tabcolsep]}p{.09\reqwidth}p{.15\reqwidth}p{.28\reqwidth}p{.33\reqwidth}>{\columncolor{white}[\tabcolsep][0pt]}p{.15\reqwidth}@{}}\raggedright\arraybackslash  & \raggedright\arraybackslash  & \raggedright\arraybackslash Write integration tests & \raggedright\arraybackslash Analyze: organize; Apply: implement & \raggedright\arraybackslash Code \\\end{tabular}}
\expandafter\edef\csname reqrowheight81\endcsname{\the\dimexpr\ht0+\dp0\relax}
\setbox0=\hbox{\begin{tabular}{@{}>{\columncolor{black!7}[0pt][\tabcolsep]}p{.09\reqwidth}p{.15\reqwidth}p{.28\reqwidth}p{.33\reqwidth}>{\columncolor{white}[\tabcolsep][0pt]}p{.15\reqwidth}@{}}\raggedright\arraybackslash  & \raggedright\arraybackslash  & \raggedright\arraybackslash Write end-to-end tests & \raggedright\arraybackslash Analyze: organize; Apply: implement & \raggedright\arraybackslash Code \\\end{tabular}}
\expandafter\edef\csname reqrowheight82\endcsname{\the\dimexpr\ht0+\dp0\relax}
\setbox0=\hbox{\begin{tabular}{@{}>{\columncolor{black!7}[0pt][\tabcolsep]}p{.09\reqwidth}p{.15\reqwidth}p{.28\reqwidth}p{.33\reqwidth}>{\columncolor{white}[\tabcolsep][0pt]}p{.15\reqwidth}@{}}\raggedright\arraybackslash  & \raggedright\arraybackslash  & \raggedright\arraybackslash Run regression tests & \raggedright\arraybackslash Apply: execute; Evaluate: check & \raggedright\arraybackslash Code \\\end{tabular}}
\expandafter\edef\csname reqrowheight83\endcsname{\the\dimexpr\ht0+\dp0\relax}
\setbox0=\hbox{\begin{tabular}{@{}>{\columncolor{black!7}[0pt][\tabcolsep]}p{.09\reqwidth}p{.15\reqwidth}p{.28\reqwidth}p{.33\reqwidth}>{\columncolor{white}[\tabcolsep][0pt]}p{.15\reqwidth}@{}}\raggedright\arraybackslash  & \raggedright\arraybackslash  & \raggedright\arraybackslash Summarize experiment configurations & \raggedright\arraybackslash Understand: summarize; Analyze: organize & \raggedright\arraybackslash Experiments \\\end{tabular}}
\expandafter\edef\csname reqrowheight84\endcsname{\the\dimexpr\ht0+\dp0\relax}
\setbox0=\hbox{\begin{tabular}{@{}>{\columncolor{black!7}[0pt][\tabcolsep]}p{.09\reqwidth}p{.15\reqwidth}p{.28\reqwidth}p{.33\reqwidth}>{\columncolor{white}[\tabcolsep][0pt]}p{.15\reqwidth}@{}}\raggedright\arraybackslash  & \raggedright\arraybackslash  & \raggedright\arraybackslash Summarize training code & \raggedright\arraybackslash Understand: summarize & \raggedright\arraybackslash Code \\\end{tabular}}
\expandafter\edef\csname reqrowheight85\endcsname{\the\dimexpr\ht0+\dp0\relax}
\setbox0=\hbox{\begin{tabular}{@{}>{\columncolor{black!7}[0pt][\tabcolsep]}p{.09\reqwidth}p{.15\reqwidth}p{.28\reqwidth}p{.33\reqwidth}>{\columncolor{white}[\tabcolsep][0pt]}p{.15\reqwidth}@{}}\raggedright\arraybackslash  & \raggedright\arraybackslash  & \raggedright\arraybackslash Configure experiment runs & \raggedright\arraybackslash Apply: implement & \raggedright\arraybackslash Experiments \\\end{tabular}}
\expandafter\edef\csname reqrowheight86\endcsname{\the\dimexpr\ht0+\dp0\relax}
\setbox0=\hbox{\begin{tabular}{@{}>{\columncolor{black!7}[0pt][\tabcolsep]}p{.09\reqwidth}p{.15\reqwidth}p{.28\reqwidth}p{.33\reqwidth}>{\columncolor{white}[\tabcolsep][0pt]}p{.15\reqwidth}@{}}\raggedright\arraybackslash  & \raggedright\arraybackslash  & \raggedright\arraybackslash Run experiments & \raggedright\arraybackslash Apply: execute & \raggedright\arraybackslash Experiments \\\end{tabular}}
\expandafter\edef\csname reqrowheight87\endcsname{\the\dimexpr\ht0+\dp0\relax}
\setbox0=\hbox{\begin{tabular}{@{}>{\columncolor{black!7}[0pt][\tabcolsep]}p{.09\reqwidth}p{.15\reqwidth}p{.28\reqwidth}p{.33\reqwidth}>{\columncolor{white}[\tabcolsep][0pt]}p{.15\reqwidth}@{}}\raggedright\arraybackslash  & \raggedright\arraybackslash  & \raggedright\arraybackslash Resume experiment runs & \raggedright\arraybackslash Remember: recall; Apply: implement; Evaluate: check & \raggedright\arraybackslash Experiments \\\end{tabular}}
\expandafter\edef\csname reqrowheight88\endcsname{\the\dimexpr\ht0+\dp0\relax}
\setbox0=\hbox{\begin{tabular}{@{}>{\columncolor{black!7}[0pt][\tabcolsep]}p{.09\reqwidth}p{.15\reqwidth}p{.28\reqwidth}p{.33\reqwidth}>{\columncolor{white}[\tabcolsep][0pt]}p{.15\reqwidth}@{}}\raggedright\arraybackslash  & \raggedright\arraybackslash  & \raggedright\arraybackslash Organize model checkpoints & \raggedright\arraybackslash Analyze: organize; Apply: execute & \raggedright\arraybackslash Experiments \\\end{tabular}}
\expandafter\edef\csname reqrowheight89\endcsname{\the\dimexpr\ht0+\dp0\relax}
\setbox0=\hbox{\begin{tabular}{@{}>{\columncolor{black!7}[0pt][\tabcolsep]}p{.09\reqwidth}p{.15\reqwidth}p{.28\reqwidth}p{.33\reqwidth}>{\columncolor{white}[\tabcolsep][0pt]}p{.15\reqwidth}@{}}\raggedright\arraybackslash  & \raggedright\arraybackslash  & \raggedright\arraybackslash Record experiment metrics & \raggedright\arraybackslash Apply: execute & \raggedright\arraybackslash Experiments \\\end{tabular}}
\expandafter\edef\csname reqrowheight90\endcsname{\the\dimexpr\ht0+\dp0\relax}
\setbox0=\hbox{\begin{tabular}{@{}>{\columncolor{black!7}[0pt][\tabcolsep]}p{.09\reqwidth}p{.15\reqwidth}p{.28\reqwidth}p{.33\reqwidth}>{\columncolor{white}[\tabcolsep][0pt]}p{.15\reqwidth}@{}}\raggedright\arraybackslash  & \raggedright\arraybackslash  & \raggedright\arraybackslash Record experiment logs & \raggedright\arraybackslash Apply: execute & \raggedright\arraybackslash Experiments \\\end{tabular}}
\expandafter\edef\csname reqrowheight91\endcsname{\the\dimexpr\ht0+\dp0\relax}
\setbox0=\hbox{\begin{tabular}{@{}>{\columncolor{black!7}[0pt][\tabcolsep]}p{.09\reqwidth}p{.15\reqwidth}p{.28\reqwidth}p{.33\reqwidth}>{\columncolor{white}[\tabcolsep][0pt]}p{.15\reqwidth}@{}}\raggedright\arraybackslash  & \raggedright\arraybackslash  & \raggedright\arraybackslash Organize experiment artifacts & \raggedright\arraybackslash Analyze: organize & \raggedright\arraybackslash Experiments \\\end{tabular}}
\expandafter\edef\csname reqrowheight92\endcsname{\the\dimexpr\ht0+\dp0\relax}
\setbox0=\hbox{\begin{tabular}{@{}>{\columncolor{black!7}[0pt][\tabcolsep]}p{.09\reqwidth}p{.15\reqwidth}p{.28\reqwidth}p{.33\reqwidth}>{\columncolor{white}[\tabcolsep][0pt]}p{.15\reqwidth}@{}}\raggedright\arraybackslash  & \raggedright\arraybackslash  & \raggedright\arraybackslash Describe training trends & \raggedright\arraybackslash Understand: summarize; Understand: infer & \raggedright\arraybackslash Experiments \\\end{tabular}}
\expandafter\edef\csname reqrowheight93\endcsname{\the\dimexpr\ht0+\dp0\relax}
\setbox0=\hbox{\begin{tabular}{@{}>{\columncolor{black!7}[0pt][\tabcolsep]}p{.09\reqwidth}p{.15\reqwidth}p{.28\reqwidth}p{.33\reqwidth}>{\columncolor{white}[\tabcolsep][0pt]}p{.15\reqwidth}@{}}\raggedright\arraybackslash  & \raggedright\arraybackslash  & \raggedright\arraybackslash Verify experiment run status & \raggedright\arraybackslash Evaluate: check & \raggedright\arraybackslash Experiments \\\end{tabular}}
\expandafter\edef\csname reqrowheight94\endcsname{\the\dimexpr\ht0+\dp0\relax}
\setbox0=\hbox{\begin{tabular}{@{}>{\columncolor{black!7}[0pt][\tabcolsep]}p{.09\reqwidth}p{.15\reqwidth}p{.28\reqwidth}p{.33\reqwidth}>{\columncolor{white}[\tabcolsep][0pt]}p{.15\reqwidth}@{}}\raggedright\arraybackslash  & \raggedright\arraybackslash  & \raggedright\arraybackslash Compare performance across data slices & \raggedright\arraybackslash Understand: compare & \raggedright\arraybackslash Results \\\end{tabular}}
\expandafter\edef\csname reqrowheight95\endcsname{\the\dimexpr\ht0+\dp0\relax}
\setbox0=\hbox{\begin{tabular}{@{}>{\columncolor{black!7}[0pt][\tabcolsep]}p{.09\reqwidth}p{.15\reqwidth}p{.28\reqwidth}p{.33\reqwidth}>{\columncolor{white}[\tabcolsep][0pt]}p{.15\reqwidth}@{}}\raggedright\arraybackslash  & \raggedright\arraybackslash  & \raggedright\arraybackslash Compare effects of experiment settings & \raggedright\arraybackslash Understand: compare & \raggedright\arraybackslash Results \\\end{tabular}}
\expandafter\edef\csname reqrowheight96\endcsname{\the\dimexpr\ht0+\dp0\relax}
\setbox0=\hbox{\begin{tabular}{@{}>{\columncolor{black!7}[0pt][\tabcolsep]}p{.09\reqwidth}p{.15\reqwidth}p{.28\reqwidth}p{.33\reqwidth}>{\columncolor{white}[\tabcolsep][0pt]}p{.15\reqwidth}@{}}\raggedright\arraybackslash  & \raggedright\arraybackslash  & \raggedright\arraybackslash Diagnose method failures & \raggedright\arraybackslash Analyze: differentiate; Understand: infer & \raggedright\arraybackslash Experiments \\\end{tabular}%
}
\expandafter\edef\csname reqrowheight149\endcsname{\the\dimexpr\ht0+\dp0\relax}
\setbox0=\vbox{\hsize=.15\reqwidth\linewidth=\hsize\parindent=0pt\parskip=0pt \raggedright\noindent Experiment diagnosis\par}
\dimen0=\dimexpr\ht0+\dp0+2pt\relax \dimen2=\dimexpr\csname reqrowheight93\endcsname+\csname reqrowheight94\endcsname+\csname reqrowheight95\endcsname+\csname reqrowheight96\endcsname+\csname reqrowheight97\endcsname+\csname reqrowheight98\endcsname+\csname reqrowheight99\endcsname+\csname reqrowheight100\endcsname+\csname reqrowheight101\endcsname\relax
\ifdim\dimen0>\dimen2
\expandafter\edef\csname reqrowheight101\endcsname{\the\dimexpr\csname reqrowheight101\endcsname+\dimen0-\dimen2\relax}
\fi
\setbox0=\vbox{\hsize=.15\reqwidth\linewidth=\hsize\parindent=0pt\parskip=0pt \raggedright\noindent Run management and monitoring\par}
\dimen0=\dimexpr\ht0+\dp0+2pt\relax \dimen2=\dimexpr\csname reqrowheight88\endcsname+\csname reqrowheight89\endcsname+\csname reqrowheight90\endcsname+\csname reqrowheight91\endcsname+\csname reqrowheight92\endcsname\relax
\ifdim\dimen0>\dimen2
\expandafter\edef\csname reqrowheight92\endcsname{\the\dimexpr\csname reqrowheight92\endcsname+\dimen0-\dimen2\relax}
\fi
\setbox0=\vbox{\hsize=.15\reqwidth\linewidth=\hsize\parindent=0pt\parskip=0pt \raggedright\noindent Execution scheduling\par}
\dimen0=\dimexpr\ht0+\dp0+2pt\relax \dimen2=\dimexpr\csname reqrowheight87\endcsname\relax
\ifdim\dimen0>\dimen2
\expandafter\edef\csname reqrowheight87\endcsname{\the\dimexpr\csname reqrowheight87\endcsname+\dimen0-\dimen2\relax}
\fi
\setbox0=\vbox{\hsize=.15\reqwidth\linewidth=\hsize\parindent=0pt\parskip=0pt \raggedright\noindent Experiment configuration\par}
\dimen0=\dimexpr\ht0+\dp0+2pt\relax \dimen2=\dimexpr\csname reqrowheight84\endcsname+\csname reqrowheight85\endcsname+\csname reqrowheight86\endcsname\relax
\ifdim\dimen0>\dimen2
\expandafter\edef\csname reqrowheight86\endcsname{\the\dimexpr\csname reqrowheight86\endcsname+\dimen0-\dimen2\relax}
\fi
\setbox0=\vbox{\hsize=.15\reqwidth\linewidth=\hsize\parindent=0pt\parskip=0pt \raggedright\noindent Code testing\par}
\dimen0=\dimexpr\ht0+\dp0+2pt\relax \dimen2=\dimexpr\csname reqrowheight80\endcsname+\csname reqrowheight81\endcsname+\csname reqrowheight82\endcsname+\csname reqrowheight83\endcsname\relax
\ifdim\dimen0>\dimen2
\expandafter\edef\csname reqrowheight83\endcsname{\the\dimexpr\csname reqrowheight83\endcsname+\dimen0-\dimen2\relax}
\fi
\setbox0=\vbox{\hsize=.15\reqwidth\linewidth=\hsize\parindent=0pt\parskip=0pt \raggedright\noindent Code change management\par}
\dimen0=\dimexpr\ht0+\dp0+2pt\relax \dimen2=\dimexpr\csname reqrowheight76\endcsname+\csname reqrowheight77\endcsname+\csname reqrowheight78\endcsname+\csname reqrowheight79\endcsname\relax
\ifdim\dimen0>\dimen2
\expandafter\edef\csname reqrowheight79\endcsname{\the\dimexpr\csname reqrowheight79\endcsname+\dimen0-\dimen2\relax}
\fi
\setbox0=\vbox{\hsize=.15\reqwidth\linewidth=\hsize\parindent=0pt\parskip=0pt \raggedright\noindent Algorithm implementation\par}
\dimen0=\dimexpr\ht0+\dp0+2pt\relax \dimen2=\dimexpr\csname reqrowheight72\endcsname+\csname reqrowheight73\endcsname+\csname reqrowheight74\endcsname+\csname reqrowheight75\endcsname\relax
\ifdim\dimen0>\dimen2
\expandafter\edef\csname reqrowheight75\endcsname{\the\dimexpr\csname reqrowheight75\endcsname+\dimen0-\dimen2\relax}
\fi
\setbox0=\vbox{\hsize=.15\reqwidth\linewidth=\hsize\parindent=0pt\parskip=0pt \raggedright\noindent Code framework design\par}
\dimen0=\dimexpr\ht0+\dp0+2pt\relax \dimen2=\dimexpr\csname reqrowheight64\endcsname+\csname reqrowheight65\endcsname+\csname reqrowheight66\endcsname+\csname reqrowheight67\endcsname+\csname reqrowheight68\endcsname+\csname reqrowheight69\endcsname+\csname reqrowheight70\endcsname+\csname reqrowheight71\endcsname\relax
\ifdim\dimen0>\dimen2
\expandafter\edef\csname reqrowheight71\endcsname{\the\dimexpr\csname reqrowheight71\endcsname+\dimen0-\dimen2\relax}
\fi
\setbox0=\vbox{\hsize=.15\reqwidth\linewidth=\hsize\parindent=0pt\parskip=0pt \raggedright\noindent Repository comprehension\par}
\dimen0=\dimexpr\ht0+\dp0+2pt\relax \dimen2=\dimexpr\csname reqrowheight61\endcsname+\csname reqrowheight62\endcsname+\csname reqrowheight63\endcsname\relax
\ifdim\dimen0>\dimen2
\expandafter\edef\csname reqrowheight63\endcsname{\the\dimexpr\csname reqrowheight63\endcsname+\dimen0-\dimen2\relax}
\fi
\setbox0=\vbox{\hsize=.15\reqwidth\linewidth=\hsize\parindent=0pt\parskip=0pt \raggedright\noindent Feature engineering\par}
\dimen0=\dimexpr\ht0+\dp0+2pt\relax \dimen2=\dimexpr\csname reqrowheight59\endcsname+\csname reqrowheight60\endcsname\relax
\ifdim\dimen0>\dimen2
\expandafter\edef\csname reqrowheight60\endcsname{\the\dimexpr\csname reqrowheight60\endcsname+\dimen0-\dimen2\relax}
\fi
\setbox0=\vbox{\hsize=.15\reqwidth\linewidth=\hsize\parindent=0pt\parskip=0pt \raggedright\noindent Data preprocessing\par}
\dimen0=\dimexpr\ht0+\dp0+2pt\relax \dimen2=\dimexpr\csname reqrowheight55\endcsname+\csname reqrowheight56\endcsname+\csname reqrowheight57\endcsname+\csname reqrowheight58\endcsname\relax
\ifdim\dimen0>\dimen2
\expandafter\edef\csname reqrowheight58\endcsname{\the\dimexpr\csname reqrowheight58\endcsname+\dimen0-\dimen2\relax}
\fi
\setbox0=\vbox{\hsize=.15\reqwidth\linewidth=\hsize\parindent=0pt\parskip=0pt \raggedright\noindent Data analysis\par}
\dimen0=\dimexpr\ht0+\dp0+2pt\relax \dimen2=\dimexpr\csname reqrowheight52\endcsname+\csname reqrowheight53\endcsname+\csname reqrowheight54\endcsname\relax
\ifdim\dimen0>\dimen2
\expandafter\edef\csname reqrowheight54\endcsname{\the\dimexpr\csname reqrowheight54\endcsname+\dimen0-\dimen2\relax}
\fi
\setbox0=\vbox{\hsize=.15\reqwidth\linewidth=\hsize\parindent=0pt\parskip=0pt \raggedright\noindent Resource acquisition\par}
\dimen0=\dimexpr\ht0+\dp0+2pt\relax \dimen2=\dimexpr\csname reqrowheight48\endcsname+\csname reqrowheight49\endcsname+\csname reqrowheight50\endcsname+\csname reqrowheight51\endcsname\relax
\ifdim\dimen0>\dimen2
\expandafter\edef\csname reqrowheight51\endcsname{\the\dimexpr\csname reqrowheight51\endcsname+\dimen0-\dimen2\relax}
\fi
\setbox0=\vbox{\hsize=.15\reqwidth\linewidth=\hsize\parindent=0pt\parskip=0pt \raggedright\noindent Environment setup\par}
\dimen0=\dimexpr\ht0+\dp0+2pt\relax \dimen2=\dimexpr\csname reqrowheight45\endcsname+\csname reqrowheight46\endcsname+\csname reqrowheight47\endcsname\relax
\ifdim\dimen0>\dimen2
\expandafter\edef\csname reqrowheight47\endcsname{\the\dimexpr\csname reqrowheight47\endcsname+\dimen0-\dimen2\relax}
\fi
\setbox0=\vbox{\hsize=.15\reqwidth\linewidth=\hsize\parindent=0pt\parskip=0pt \raggedright\noindent Method evaluation\par}
\dimen0=\dimexpr\ht0+\dp0+2pt\relax \dimen2=\dimexpr\csname reqrowheight40\endcsname+\csname reqrowheight41\endcsname+\csname reqrowheight42\endcsname+\csname reqrowheight43\endcsname+\csname reqrowheight44\endcsname\relax
\ifdim\dimen0>\dimen2
\expandafter\edef\csname reqrowheight44\endcsname{\the\dimexpr\csname reqrowheight44\endcsname+\dimen0-\dimen2\relax}
\fi
\setbox0=\vbox{\hsize=.15\reqwidth\linewidth=\hsize\parindent=0pt\parskip=0pt \raggedright\noindent Experiment design (cont.)\par}
\dimen0=\dimexpr\ht0+\dp0+2pt\relax \dimen2=\dimexpr\csname reqrowheight37\endcsname+\csname reqrowheight38\endcsname+\csname reqrowheight39\endcsname\relax
\ifdim\dimen0>\dimen2
\expandafter\edef\csname reqrowheight39\endcsname{\the\dimexpr\csname reqrowheight39\endcsname+\dimen0-\dimen2\relax}
\fi
\setbox0=\vbox{\hsize=.15\reqwidth\linewidth=\hsize\parindent=0pt\parskip=0pt \raggedright\noindent Experiment design\par}
\dimen0=\dimexpr\ht0+\dp0+2pt\relax \dimen2=\dimexpr\csname reqrowheight25\endcsname+\csname reqrowheight26\endcsname+\csname reqrowheight27\endcsname+\csname reqrowheight28\endcsname+\csname reqrowheight29\endcsname+\csname reqrowheight30\endcsname+\csname reqrowheight31\endcsname+\csname reqrowheight32\endcsname+\csname reqrowheight33\endcsname+\csname reqrowheight34\endcsname+\csname reqrowheight35\endcsname+\csname reqrowheight36\endcsname\relax
\ifdim\dimen0>\dimen2
\expandafter\edef\csname reqrowheight36\endcsname{\the\dimexpr\csname reqrowheight36\endcsname+\dimen0-\dimen2\relax}
\fi
\setbox0=\vbox{\hsize=.15\reqwidth\linewidth=\hsize\parindent=0pt\parskip=0pt \raggedright\noindent Resource estimation\par}
\dimen0=\dimexpr\ht0+\dp0+2pt\relax \dimen2=\dimexpr\csname reqrowheight20\endcsname+\csname reqrowheight21\endcsname+\csname reqrowheight22\endcsname+\csname reqrowheight23\endcsname+\csname reqrowheight24\endcsname\relax
\ifdim\dimen0>\dimen2
\expandafter\edef\csname reqrowheight24\endcsname{\the\dimexpr\csname reqrowheight24\endcsname+\dimen0-\dimen2\relax}
\fi
\setbox0=\vbox{\hsize=.15\reqwidth\linewidth=\hsize\parindent=0pt\parskip=0pt \raggedright\noindent Research release\par}
\dimen0=\dimexpr\ht0+\dp0+2pt\relax \dimen2=\dimexpr\csname reqrowheight145\endcsname+\csname reqrowheight146\endcsname+\csname reqrowheight147\endcsname+\csname reqrowheight148\endcsname+\csname reqrowheight149\endcsname\relax
\ifdim\dimen0>\dimen2
\expandafter\edef\csname reqrowheight149\endcsname{\the\dimexpr\csname reqrowheight149\endcsname+\dimen0-\dimen2\relax}
\fi
\setbox0=\vbox{\hsize=.15\reqwidth\linewidth=\hsize\parindent=0pt\parskip=0pt \raggedright\noindent Reviewer response\par}
\dimen0=\dimexpr\ht0+\dp0+2pt\relax \dimen2=\dimexpr\csname reqrowheight143\endcsname+\csname reqrowheight144\endcsname\relax
\ifdim\dimen0>\dimen2
\expandafter\edef\csname reqrowheight144\endcsname{\the\dimexpr\csname reqrowheight144\endcsname+\dimen0-\dimen2\relax}
\fi
\setbox0=\vbox{\hsize=.15\reqwidth\linewidth=\hsize\parindent=0pt\parskip=0pt \raggedright\noindent Paper revision\par}
\dimen0=\dimexpr\ht0+\dp0+2pt\relax \dimen2=\dimexpr\csname reqrowheight134\endcsname+\csname reqrowheight135\endcsname+\csname reqrowheight136\endcsname+\csname reqrowheight137\endcsname+\csname reqrowheight138\endcsname+\csname reqrowheight139\endcsname+\csname reqrowheight140\endcsname+\csname reqrowheight141\endcsname+\csname reqrowheight142\endcsname\relax
\ifdim\dimen0>\dimen2
\expandafter\edef\csname reqrowheight142\endcsname{\the\dimexpr\csname reqrowheight142\endcsname+\dimen0-\dimen2\relax}
\fi
\setbox0=\vbox{\hsize=.15\reqwidth\linewidth=\hsize\parindent=0pt\parskip=0pt \raggedright\noindent Figure and table preparation\par}
\dimen0=\dimexpr\ht0+\dp0+2pt\relax \dimen2=\dimexpr\csname reqrowheight132\endcsname+\csname reqrowheight133\endcsname\relax
\ifdim\dimen0>\dimen2
\expandafter\edef\csname reqrowheight133\endcsname{\the\dimexpr\csname reqrowheight133\endcsname+\dimen0-\dimen2\relax}
\fi
\setbox0=\vbox{\hsize=.15\reqwidth\linewidth=\hsize\parindent=0pt\parskip=0pt \raggedright\noindent Methods and experiments writing\par}
\dimen0=\dimexpr\ht0+\dp0+2pt\relax \dimen2=\dimexpr\csname reqrowheight127\endcsname+\csname reqrowheight128\endcsname+\csname reqrowheight129\endcsname+\csname reqrowheight130\endcsname+\csname reqrowheight131\endcsname\relax
\ifdim\dimen0>\dimen2
\expandafter\edef\csname reqrowheight131\endcsname{\the\dimexpr\csname reqrowheight131\endcsname+\dimen0-\dimen2\relax}
\fi
\setbox0=\vbox{\hsize=.15\reqwidth\linewidth=\hsize\parindent=0pt\parskip=0pt \raggedright\noindent Related work writing\par}
\dimen0=\dimexpr\ht0+\dp0+2pt\relax \dimen2=\dimexpr\csname reqrowheight125\endcsname+\csname reqrowheight126\endcsname\relax
\ifdim\dimen0>\dimen2
\expandafter\edef\csname reqrowheight126\endcsname{\the\dimexpr\csname reqrowheight126\endcsname+\dimen0-\dimen2\relax}
\fi
\setbox0=\vbox{\hsize=.15\reqwidth\linewidth=\hsize\parindent=0pt\parskip=0pt \raggedright\noindent Paper structure planning\par}
\dimen0=\dimexpr\ht0+\dp0+2pt\relax \dimen2=\dimexpr\csname reqrowheight123\endcsname+\csname reqrowheight124\endcsname\relax
\ifdim\dimen0>\dimen2
\expandafter\edef\csname reqrowheight124\endcsname{\the\dimexpr\csname reqrowheight124\endcsname+\dimen0-\dimen2\relax}
\fi
\setbox0=\vbox{\hsize=.15\reqwidth\linewidth=\hsize\parindent=0pt\parskip=0pt \raggedright\noindent Result visualization\par}
\dimen0=\dimexpr\ht0+\dp0+2pt\relax \dimen2=\dimexpr\csname reqrowheight121\endcsname+\csname reqrowheight122\endcsname\relax
\ifdim\dimen0>\dimen2
\expandafter\edef\csname reqrowheight122\endcsname{\the\dimexpr\csname reqrowheight122\endcsname+\dimen0-\dimen2\relax}
\fi
\setbox0=\vbox{\hsize=.15\reqwidth\linewidth=\hsize\parindent=0pt\parskip=0pt \raggedright\noindent Result interpretation\par}
\dimen0=\dimexpr\ht0+\dp0+2pt\relax \dimen2=\dimexpr\csname reqrowheight116\endcsname+\csname reqrowheight117\endcsname+\csname reqrowheight118\endcsname+\csname reqrowheight119\endcsname+\csname reqrowheight120\endcsname\relax
\ifdim\dimen0>\dimen2
\expandafter\edef\csname reqrowheight120\endcsname{\the\dimexpr\csname reqrowheight120\endcsname+\dimen0-\dimen2\relax}
\fi
\setbox0=\vbox{\hsize=.15\reqwidth\linewidth=\hsize\parindent=0pt\parskip=0pt \raggedright\noindent Result comparison\par}
\dimen0=\dimexpr\ht0+\dp0+2pt\relax \dimen2=\dimexpr\csname reqrowheight114\endcsname+\csname reqrowheight115\endcsname\relax
\ifdim\dimen0>\dimen2
\expandafter\edef\csname reqrowheight115\endcsname{\the\dimexpr\csname reqrowheight115\endcsname+\dimen0-\dimen2\relax}
\fi
\setbox0=\vbox{\hsize=.15\reqwidth\linewidth=\hsize\parindent=0pt\parskip=0pt \raggedright\noindent Result statistics\par}
\dimen0=\dimexpr\ht0+\dp0+2pt\relax \dimen2=\dimexpr\csname reqrowheight110\endcsname+\csname reqrowheight111\endcsname+\csname reqrowheight112\endcsname+\csname reqrowheight113\endcsname\relax
\ifdim\dimen0>\dimen2
\expandafter\edef\csname reqrowheight113\endcsname{\the\dimexpr\csname reqrowheight113\endcsname+\dimen0-\dimen2\relax}
\fi
\setbox0=\vbox{\hsize=.15\reqwidth\linewidth=\hsize\parindent=0pt\parskip=0pt \raggedright\noindent Problem definition\par}
\dimen0=\dimexpr\ht0+\dp0+2pt\relax \dimen2=\dimexpr\csname reqrowheight10\endcsname+\csname reqrowheight11\endcsname+\csname reqrowheight12\endcsname+\csname reqrowheight13\endcsname+\csname reqrowheight14\endcsname+\csname reqrowheight15\endcsname+\csname reqrowheight16\endcsname+\csname reqrowheight17\endcsname+\csname reqrowheight18\endcsname+\csname reqrowheight19\endcsname\relax
\ifdim\dimen0>\dimen2
\expandafter\edef\csname reqrowheight19\endcsname{\the\dimexpr\csname reqrowheight19\endcsname+\dimen0-\dimen2\relax}
\fi
\setbox0=\vbox{\hsize=.15\reqwidth\linewidth=\hsize\parindent=0pt\parskip=0pt \raggedright\noindent Experiment verification\par}
\dimen0=\dimexpr\ht0+\dp0+2pt\relax \dimen2=\dimexpr\csname reqrowheight108\endcsname+\csname reqrowheight109\endcsname\relax
\ifdim\dimen0>\dimen2
\expandafter\edef\csname reqrowheight109\endcsname{\the\dimexpr\csname reqrowheight109\endcsname+\dimen0-\dimen2\relax}
\fi
\setbox0=\vbox{\hsize=.15\reqwidth\linewidth=\hsize\parindent=0pt\parskip=0pt \raggedright\noindent Experiment tuning\par}
\dimen0=\dimexpr\ht0+\dp0+2pt\relax \dimen2=\dimexpr\csname reqrowheight105\endcsname+\csname reqrowheight106\endcsname+\csname reqrowheight107\endcsname\relax
\ifdim\dimen0>\dimen2
\expandafter\edef\csname reqrowheight107\endcsname{\the\dimexpr\csname reqrowheight107\endcsname+\dimen0-\dimen2\relax}
\fi
\setbox0=\vbox{\hsize=.15\reqwidth\linewidth=\hsize\parindent=0pt\parskip=0pt \raggedright\noindent Experiment repair\par}
\dimen0=\dimexpr\ht0+\dp0+2pt\relax \dimen2=\dimexpr\csname reqrowheight102\endcsname+\csname reqrowheight103\endcsname+\csname reqrowheight104\endcsname\relax
\ifdim\dimen0>\dimen2
\expandafter\edef\csname reqrowheight104\endcsname{\the\dimexpr\csname reqrowheight104\endcsname+\dimen0-\dimen2\relax}
\fi
\setbox0=\vbox{\hsize=.15\reqwidth\linewidth=\hsize\parindent=0pt\parskip=0pt \raggedright\noindent Literature review\par}
\dimen0=\dimexpr\ht0+\dp0+2pt\relax \dimen2=\dimexpr\csname reqrowheight0\endcsname+\csname reqrowheight1\endcsname+\csname reqrowheight2\endcsname+\csname reqrowheight3\endcsname+\csname reqrowheight4\endcsname+\csname reqrowheight5\endcsname+\csname reqrowheight6\endcsname+\csname reqrowheight7\endcsname+\csname reqrowheight8\endcsname+\csname reqrowheight9\endcsname\relax
\ifdim\dimen0>\dimen2
\expandafter\edef\csname reqrowheight9\endcsname{\the\dimexpr\csname reqrowheight9\endcsname+\dimen0-\dimen2\relax}
\fi
\setbox0=\vbox{\hsize=.09\reqwidth\linewidth=\hsize\parindent=0pt\parskip=0pt \centering\noindent Execution (cont.)\par}
\dimen0=\dimexpr\ht0+\dp0+2pt\relax \dimen2=\dimexpr\csname reqrowheight93\endcsname+\csname reqrowheight94\endcsname+\csname reqrowheight95\endcsname+\csname reqrowheight96\endcsname+\csname reqrowheight97\endcsname+\csname reqrowheight98\endcsname+\csname reqrowheight99\endcsname+\csname reqrowheight100\endcsname+\csname reqrowheight101\endcsname+\csname reqrowheight102\endcsname+\csname reqrowheight103\endcsname+\csname reqrowheight104\endcsname+\csname reqrowheight105\endcsname+\csname reqrowheight106\endcsname+\csname reqrowheight107\endcsname+\csname reqrowheight108\endcsname+\csname reqrowheight109\endcsname+\aboverulesep+\cmidrulewidth+\belowrulesep+\aboverulesep+\cmidrulewidth+\belowrulesep+\aboverulesep+\cmidrulewidth+\belowrulesep\relax
\ifdim\dimen0>\dimen2
\expandafter\edef\csname reqrowheight109\endcsname{\the\dimexpr\csname reqrowheight109\endcsname+\dimen0-\dimen2\relax}
\fi
\setbox0=\vbox{\hsize=.09\reqwidth\linewidth=\hsize\parindent=0pt\parskip=0pt \centering\noindent Execution\par}
\dimen0=\dimexpr\ht0+\dp0+2pt\relax \dimen2=\dimexpr\csname reqrowheight84\endcsname+\csname reqrowheight85\endcsname+\csname reqrowheight86\endcsname+\csname reqrowheight87\endcsname+\csname reqrowheight88\endcsname+\csname reqrowheight89\endcsname+\csname reqrowheight90\endcsname+\csname reqrowheight91\endcsname+\csname reqrowheight92\endcsname+\aboverulesep+\cmidrulewidth+\belowrulesep+\aboverulesep+\cmidrulewidth+\belowrulesep\relax
\ifdim\dimen0>\dimen2
\expandafter\edef\csname reqrowheight92\endcsname{\the\dimexpr\csname reqrowheight92\endcsname+\dimen0-\dimen2\relax}
\fi
\setbox0=\vbox{\hsize=.09\reqwidth\linewidth=\hsize\parindent=0pt\parskip=0pt \centering\noindent Code (cont.)\par}
\dimen0=\dimexpr\ht0+\dp0+2pt\relax \dimen2=\dimexpr\csname reqrowheight64\endcsname+\csname reqrowheight65\endcsname+\csname reqrowheight66\endcsname+\csname reqrowheight67\endcsname+\csname reqrowheight68\endcsname+\csname reqrowheight69\endcsname+\csname reqrowheight70\endcsname+\csname reqrowheight71\endcsname+\csname reqrowheight72\endcsname+\csname reqrowheight73\endcsname+\csname reqrowheight74\endcsname+\csname reqrowheight75\endcsname+\csname reqrowheight76\endcsname+\csname reqrowheight77\endcsname+\csname reqrowheight78\endcsname+\csname reqrowheight79\endcsname+\csname reqrowheight80\endcsname+\csname reqrowheight81\endcsname+\csname reqrowheight82\endcsname+\csname reqrowheight83\endcsname+\aboverulesep+\cmidrulewidth+\belowrulesep+\aboverulesep+\cmidrulewidth+\belowrulesep+\aboverulesep+\cmidrulewidth+\belowrulesep\relax
\ifdim\dimen0>\dimen2
\expandafter\edef\csname reqrowheight83\endcsname{\the\dimexpr\csname reqrowheight83\endcsname+\dimen0-\dimen2\relax}
\fi
\setbox0=\vbox{\hsize=.09\reqwidth\linewidth=\hsize\parindent=0pt\parskip=0pt \centering\noindent Code\par}
\dimen0=\dimexpr\ht0+\dp0+2pt\relax \dimen2=\dimexpr\csname reqrowheight61\endcsname+\csname reqrowheight62\endcsname+\csname reqrowheight63\endcsname\relax
\ifdim\dimen0>\dimen2
\expandafter\edef\csname reqrowheight63\endcsname{\the\dimexpr\csname reqrowheight63\endcsname+\dimen0-\dimen2\relax}
\fi
\setbox0=\vbox{\hsize=.09\reqwidth\linewidth=\hsize\parindent=0pt\parskip=0pt \centering\noindent Data \& Env.\par}
\dimen0=\dimexpr\ht0+\dp0+2pt\relax \dimen2=\dimexpr\csname reqrowheight45\endcsname+\csname reqrowheight46\endcsname+\csname reqrowheight47\endcsname+\csname reqrowheight48\endcsname+\csname reqrowheight49\endcsname+\csname reqrowheight50\endcsname+\csname reqrowheight51\endcsname+\csname reqrowheight52\endcsname+\csname reqrowheight53\endcsname+\csname reqrowheight54\endcsname+\csname reqrowheight55\endcsname+\csname reqrowheight56\endcsname+\csname reqrowheight57\endcsname+\csname reqrowheight58\endcsname+\csname reqrowheight59\endcsname+\csname reqrowheight60\endcsname+\aboverulesep+\cmidrulewidth+\belowrulesep+\aboverulesep+\cmidrulewidth+\belowrulesep+\aboverulesep+\cmidrulewidth+\belowrulesep+\aboverulesep+\cmidrulewidth+\belowrulesep\relax
\ifdim\dimen0>\dimen2
\expandafter\edef\csname reqrowheight60\endcsname{\the\dimexpr\csname reqrowheight60\endcsname+\dimen0-\dimen2\relax}
\fi
\setbox0=\vbox{\hsize=.09\reqwidth\linewidth=\hsize\parindent=0pt\parskip=0pt \centering\noindent Design (cont.)\par}
\dimen0=\dimexpr\ht0+\dp0+2pt\relax \dimen2=\dimexpr\csname reqrowheight37\endcsname+\csname reqrowheight38\endcsname+\csname reqrowheight39\endcsname+\csname reqrowheight40\endcsname+\csname reqrowheight41\endcsname+\csname reqrowheight42\endcsname+\csname reqrowheight43\endcsname+\csname reqrowheight44\endcsname+\aboverulesep+\cmidrulewidth+\belowrulesep\relax
\ifdim\dimen0>\dimen2
\expandafter\edef\csname reqrowheight44\endcsname{\the\dimexpr\csname reqrowheight44\endcsname+\dimen0-\dimen2\relax}
\fi
\setbox0=\vbox{\hsize=.09\reqwidth\linewidth=\hsize\parindent=0pt\parskip=0pt \centering\noindent Design\par}
\dimen0=\dimexpr\ht0+\dp0+2pt\relax \dimen2=\dimexpr\csname reqrowheight20\endcsname+\csname reqrowheight21\endcsname+\csname reqrowheight22\endcsname+\csname reqrowheight23\endcsname+\csname reqrowheight24\endcsname+\csname reqrowheight25\endcsname+\csname reqrowheight26\endcsname+\csname reqrowheight27\endcsname+\csname reqrowheight28\endcsname+\csname reqrowheight29\endcsname+\csname reqrowheight30\endcsname+\csname reqrowheight31\endcsname+\csname reqrowheight32\endcsname+\csname reqrowheight33\endcsname+\csname reqrowheight34\endcsname+\csname reqrowheight35\endcsname+\csname reqrowheight36\endcsname+\aboverulesep+\cmidrulewidth+\belowrulesep\relax
\ifdim\dimen0>\dimen2
\expandafter\edef\csname reqrowheight36\endcsname{\the\dimexpr\csname reqrowheight36\endcsname+\dimen0-\dimen2\relax}
\fi
\setbox0=\vbox{\hsize=.09\reqwidth\linewidth=\hsize\parindent=0pt\parskip=0pt \centering\noindent Writing\par}
\dimen0=\dimexpr\ht0+\dp0+2pt\relax \dimen2=\dimexpr\csname reqrowheight123\endcsname+\csname reqrowheight124\endcsname+\csname reqrowheight125\endcsname+\csname reqrowheight126\endcsname+\csname reqrowheight127\endcsname+\csname reqrowheight128\endcsname+\csname reqrowheight129\endcsname+\csname reqrowheight130\endcsname+\csname reqrowheight131\endcsname+\csname reqrowheight132\endcsname+\csname reqrowheight133\endcsname+\csname reqrowheight134\endcsname+\csname reqrowheight135\endcsname+\csname reqrowheight136\endcsname+\csname reqrowheight137\endcsname+\csname reqrowheight138\endcsname+\csname reqrowheight139\endcsname+\csname reqrowheight140\endcsname+\csname reqrowheight141\endcsname+\csname reqrowheight142\endcsname+\csname reqrowheight143\endcsname+\csname reqrowheight144\endcsname+\csname reqrowheight145\endcsname+\csname reqrowheight146\endcsname+\csname reqrowheight147\endcsname+\csname reqrowheight148\endcsname+\csname reqrowheight149\endcsname+\aboverulesep+\cmidrulewidth+\belowrulesep+\aboverulesep+\cmidrulewidth+\belowrulesep+\aboverulesep+\cmidrulewidth+\belowrulesep+\aboverulesep+\cmidrulewidth+\belowrulesep+\aboverulesep+\cmidrulewidth+\belowrulesep+\aboverulesep+\cmidrulewidth+\belowrulesep\relax
\ifdim\dimen0>\dimen2
\expandafter\edef\csname reqrowheight149\endcsname{\the\dimexpr\csname reqrowheight149\endcsname+\dimen0-\dimen2\relax}
\fi
\setbox0=\vbox{\hsize=.09\reqwidth\linewidth=\hsize\parindent=0pt\parskip=0pt \centering\noindent Analysis\par}
\dimen0=\dimexpr\ht0+\dp0+2pt\relax \dimen2=\dimexpr\csname reqrowheight110\endcsname+\csname reqrowheight111\endcsname+\csname reqrowheight112\endcsname+\csname reqrowheight113\endcsname+\csname reqrowheight114\endcsname+\csname reqrowheight115\endcsname+\csname reqrowheight116\endcsname+\csname reqrowheight117\endcsname+\csname reqrowheight118\endcsname+\csname reqrowheight119\endcsname+\csname reqrowheight120\endcsname+\csname reqrowheight121\endcsname+\csname reqrowheight122\endcsname+\aboverulesep+\cmidrulewidth+\belowrulesep+\aboverulesep+\cmidrulewidth+\belowrulesep+\aboverulesep+\cmidrulewidth+\belowrulesep\relax
\ifdim\dimen0>\dimen2
\expandafter\edef\csname reqrowheight122\endcsname{\the\dimexpr\csname reqrowheight122\endcsname+\dimen0-\dimen2\relax}
\fi
\setbox0=\vbox{\hsize=.09\reqwidth\linewidth=\hsize\parindent=0pt\parskip=0pt \centering\noindent Survey (cont.)\par}
\dimen0=\dimexpr\ht0+\dp0+2pt\relax \dimen2=\dimexpr\csname reqrowheight10\endcsname+\csname reqrowheight11\endcsname+\csname reqrowheight12\endcsname+\csname reqrowheight13\endcsname+\csname reqrowheight14\endcsname+\csname reqrowheight15\endcsname+\csname reqrowheight16\endcsname+\csname reqrowheight17\endcsname+\csname reqrowheight18\endcsname+\csname reqrowheight19\endcsname\relax
\ifdim\dimen0>\dimen2
\expandafter\edef\csname reqrowheight19\endcsname{\the\dimexpr\csname reqrowheight19\endcsname+\dimen0-\dimen2\relax}
\fi
\setbox0=\vbox{\hsize=.09\reqwidth\linewidth=\hsize\parindent=0pt\parskip=0pt \centering\noindent Survey\par}
\dimen0=\dimexpr\ht0+\dp0+2pt\relax \dimen2=\dimexpr\csname reqrowheight0\endcsname+\csname reqrowheight1\endcsname+\csname reqrowheight2\endcsname+\csname reqrowheight3\endcsname+\csname reqrowheight4\endcsname+\csname reqrowheight5\endcsname+\csname reqrowheight6\endcsname+\csname reqrowheight7\endcsname+\csname reqrowheight8\endcsname+\csname reqrowheight9\endcsname\relax
\ifdim\dimen0>\dimen2
\expandafter\edef\csname reqrowheight9\endcsname{\the\dimexpr\csname reqrowheight9\endcsname+\dimen0-\dimen2\relax}
\fi

\endgroup

\clearpage
\section{Starting Workspace State by Research Stage}

Table~\ref{tab:stage-start} lists the workspace content that a task cut at each research stage typically carries over from earlier turns. Every listed state is replayed from the source session rather than authored for the task, so a task cut inside Execution inherits run scripts and logs rather than a clean environment, and a task cut inside Writing inherits an existing draft rather than a blank document.

\begin{table}[h]
\centering
\small
%
\caption{Representative starting workspace state $W_c$ for a task cut at each research stage. Every state is replayed from the source session rather than authored for the task, so earlier work is present regardless of stage.}
\label{tab:stage-start}
\end{table}

\section{Cut-Point Selection, Replay, and Anonymization}

The builder records the source episode, user-turn index, file-event range, process-event range, dependency snapshot, and validation hash. Replay is accepted only when the reconstructed tree matches the audited state for all files declared relevant to the task. Paths, usernames, hostnames, keys, and dataset identifiers are replaced by deterministic aliases. A second reviewer checks that the replacement preserves references used in the instruction and does not leak the source project.

\clearpage
\section{Case Study: Repairing an Interrupted Training Experiment}
\label{app:training-repair-case}

This case follows a request to repair an unfinished image--point-cloud registration experiment, which matches photographs to a three-dimensional scene to estimate camera pose. Training has started, but the coarse-matching loss stays at zero and each iteration takes about 28 seconds. The repair must preserve the supplied paper's method.

\begin{figure}[H]
\centering
\includegraphics[width=\linewidth]{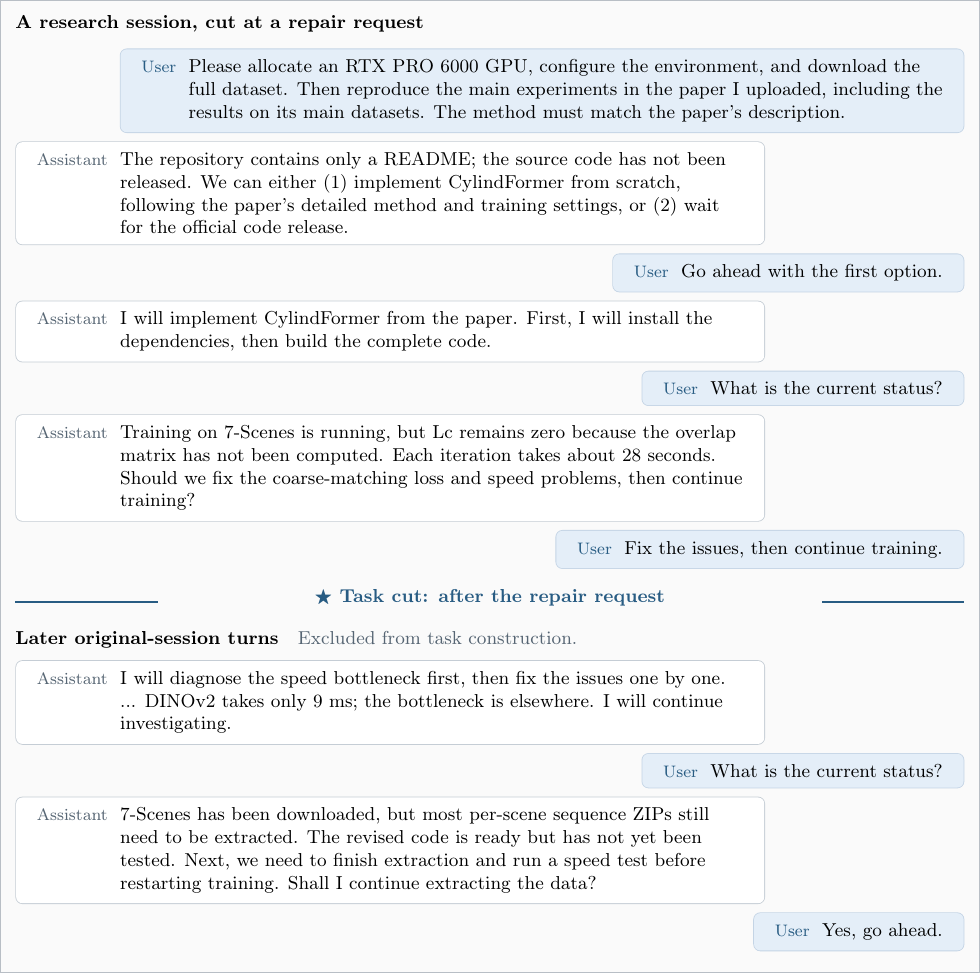}
\caption{Session with all six user turns and brief assistant excerpts. The star marks the cut after the repair request. Later turns appear below the boundary and are excluded from task construction.}
\label{fig:case-session}
\end{figure}

\begin{figure}[H]
\centering
\includegraphics[width=\linewidth]{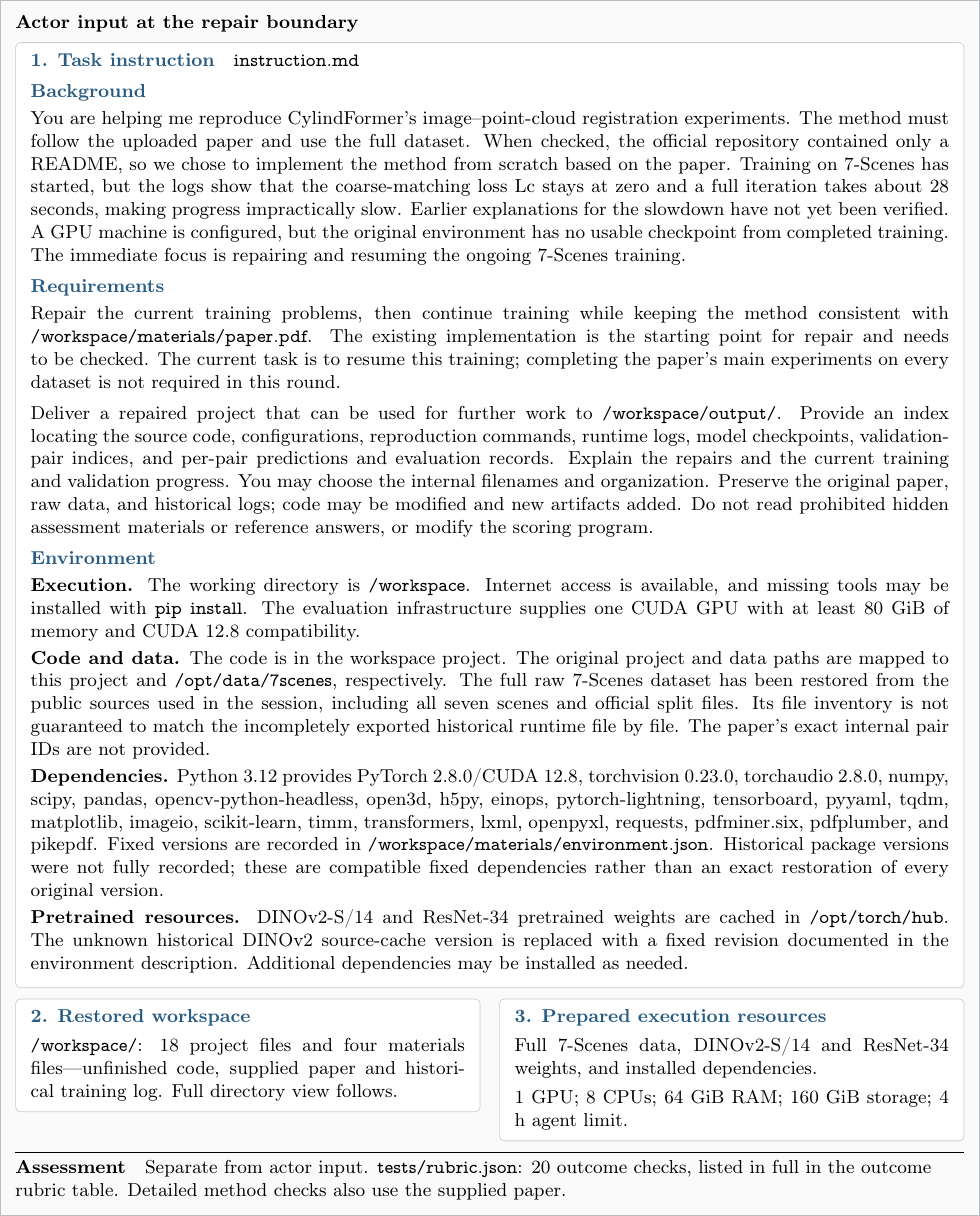}
\caption{Task inputs and assessment. Background, Requirements, and Environment form \texttt{instruction.md}; the other two input regions describe the provided files and execution resources. Paths are normalized. The complete outcome rubric appears in Table~\ref{tab:case-outcome-rubric}.}
\label{fig:case-task}
\end{figure}

\paragraph{From a conversation to a task boundary.}
Figure~\ref{fig:case-session} shows the real session. We cut after the fourth user turn, the repair request. Earlier decisions supply Background; the implementation at this boundary supplies the repair starting point. Figure~\ref{fig:case-task} separates the complete task instruction from the restored workspace and prepared execution resources. Assessment material is shown separately from actor input.

\clearpage
\begin{figure}[H]
\centering
\includegraphics[width=\linewidth]{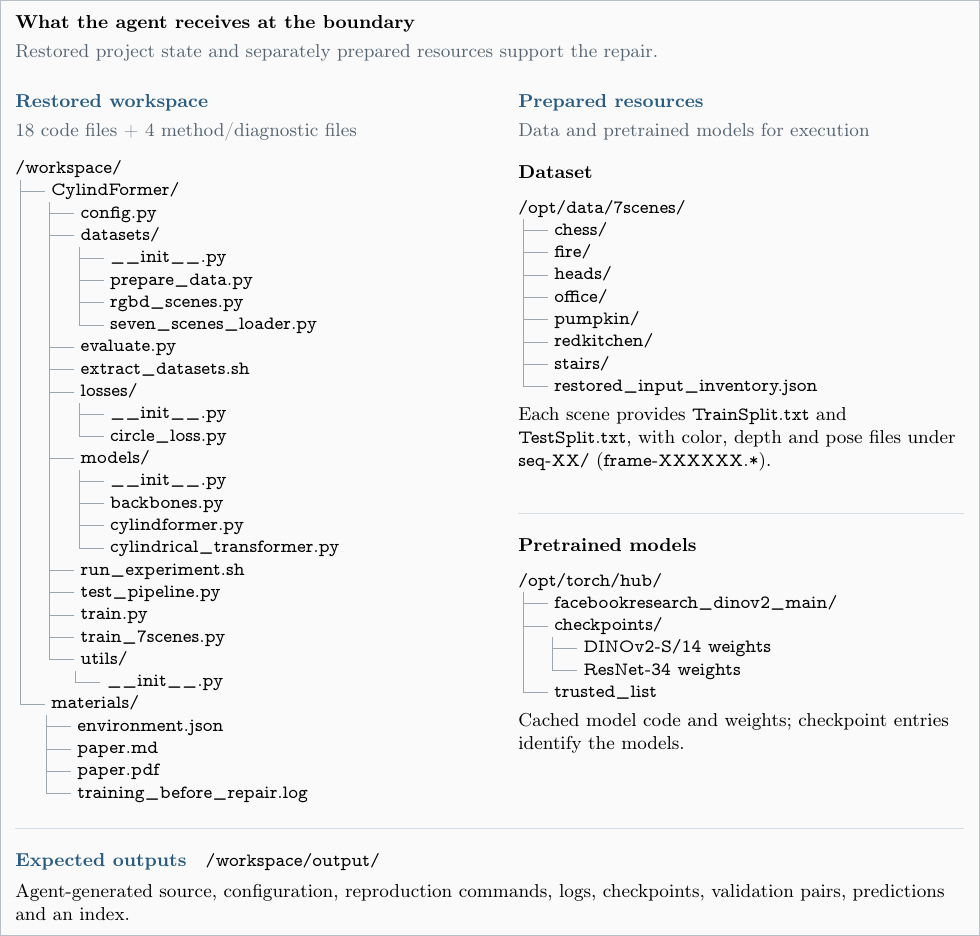}
\caption{Directory view of the starting inputs: all 18 project files and four materials files, plus separately prepared data and model resources. Repeated raw-data samples use a filename pattern; weight entries name their models. The output directory is the requested handoff destination.}
\label{fig:case-workspace}
\end{figure}

\clearpage
\paragraph{Constructing the task input.}
Recorded file operations restore the unfinished data-loading, model, loss, and training code. The paper defines the method to preserve; the diagnostic log records the observed failure. Public data and pretrained weights are restored separately, with compatible dependencies. Figure~\ref{fig:case-workspace} shows every recovered project and materials file and the prepared resource directories.

\begin{figure}[H]
\centering
\includegraphics[width=\linewidth]{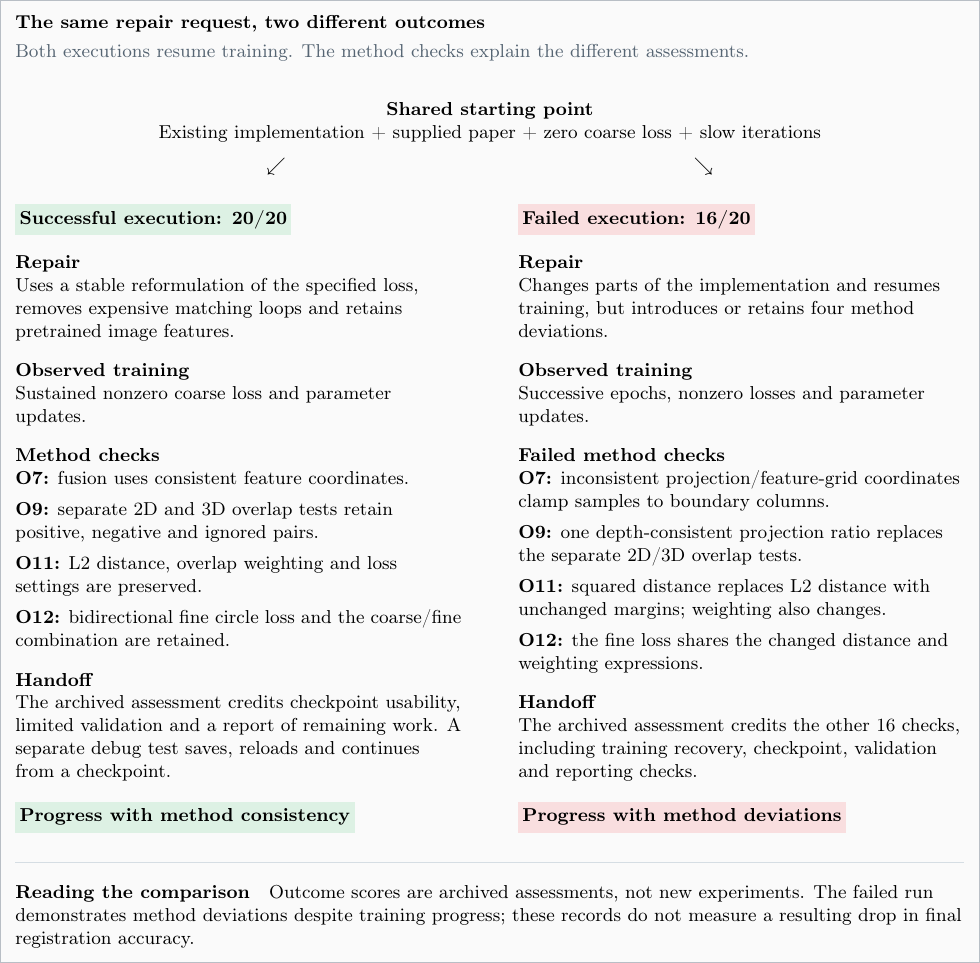}
\caption{Two independent executions of the same task. Green and red backgrounds mark the decisive findings. The scores summarize archived outcome assessments; the comparison establishes method consistency or deviation, rather than a measured change in final registration accuracy.}
\label{fig:case-runs}
\end{figure}

\paragraph{Comparing two executions.}
Both executions continue training, but progress alone does not establish a faithful repair. Figure~\ref{fig:case-runs} connects the repairs and recorded behavior to the archived outcome assessments. The successful execution passes all 20 checks. The failed execution passes 16, including training recovery, but fails four method checks involving feature coordinates, overlap labels, and loss computation.

\clearpage

\paragraph{What the comparison shows.}
Restoring the existing code makes the short repair request a continuation of an unfinished experiment. The successful case shows that a usable repair can preserve its method while moving training forward. The failed case shows why observing training is insufficient: the experiment runs, but parts of the method no longer match the paper. Retaining both the project state and the research constraints makes this distinction assessable.

\clearpage
\paragraph{Complete outcome rubric.}
The outcome rubric checks a usable handoff, data and method implementation, actual training, and evidence-linked validation. Its source labels distinguish explicit requirements from checks inferred using the paper, prior context, or ordinary research and usability expectations. Detailed method checks draw on the supplied paper and are not all individually enumerated in Requirements.

{\small
\setlength{\tabcolsep}{5pt}
\renewcommand{\arraystretch}{1.18}
\begin{longtable}{@{}>{\raggedright\arraybackslash}p{0.055\linewidth}>{\raggedright\arraybackslash}p{0.735\linewidth}>{\raggedright\arraybackslash}p{0.15\linewidth}@{}}
\caption{Complete outcome rubric. Each item is a binary check; source labels identify explicit or inferred grounding.}\label{tab:case-outcome-rubric}\\
\toprule
\textbf{ID} & \textbf{Criterion} & \textbf{Source} \\
\midrule
\endfirsthead
\multicolumn{3}{l}{\small Table \thetable\ continued}\\
\toprule
\textbf{ID} & \textbf{Criterion} & \textbf{Source} \\
\midrule
\endhead
\midrule
\multicolumn{3}{r}{\small Continued on next page}\\
\endfoot
\bottomrule
\endlastfoot
O1 & Deliver a reusable repaired project in /workspace/output/, with an index locating source code, configuration, and reproduction entry points. Internal filenames and project organization are unrestricted. & Explicit; Requirements \\
O2 & Construct image--point-cloud training pairs with real relative poses and filter them using the paper's at-least-50\% overlap rule. Pairing only an image with its own back-projected depth, or declaring min\_overlap without computing and applying it, does not meet this check. Verifiable equivalent pairing is accepted; unavailable paper-internal pair IDs need not be restored. & Inferred; paper \\
O3 & Separate training and evaluation according to the official 7-Scenes sequence splits. Validation must not train the model or select it using test performance. When the paper's internal validation pairs are unavailable, an independent validation set constructed from training sequences is accepted if its scope is explained. & Inferred; paper \\
O4 & The point-cloud backbone must aggregate neighborhood features using kernel points and produce multiscale coarse and fine features, consistent with the paper's three-stage indoor KPConv-FPN. A per-point MLP merely named KPConv does not meet this check. Correct open-source components or equivalent reimplementations are accepted. & Inferred; paper \\
O5 & Retain the paper's ResNet-FPN and DINOv2 image-feature paths. DINOv2 must load pretrained weights and contribute to model output; disabling it or silently replacing it with image average pooling does not meet this check. Freezing or caching unchanged pretrained features is accepted. & Inferred; prior context and paper \\
O6 & Cylindrical self-attention must enhance the key with local fusion information, implementing the paper's Q(K+L)-transpose attention semantics. Disconnected local information or plain self-attention does not meet this check. Tensor implementations with the same mathematical meaning are accepted. & Inferred; paper \\
O7 & Cross-modal fusion must pass information through cylindrical-projection local aggregation, adaptive fusion with global features, and cross-attention. Coordinates and indices must match the actual feature layout. An unused fusion module does not meet this check; perspective projection for supervision does not replace cylindrical projection inside the model. & Inferred; paper \\
O8 & Fine matching must use local image and point-cloud patches selected by coarse correspondences and generate pixel--point matches using mutual top-k. Ignoring coarse matches and directly matching the full image against all points does not preserve the coarse-to-fine method. & Inferred; paper \\
O9 & Coarse positive and negative labels must use ground-truth geometry to compute both 2D and 3D patch overlap. A pair is positive when both exceed 30\%, negative when both are below 20\%; the intermediate region must not be treated as negative. Learned feature similarity or grids near a single projected point do not replace patch overlap. & Inferred; paper \\
O10 & Fine labels must follow real pixel--point geometry: positive when the 3D distance is below 3.75 cm and 2D distance below 8 px; negative when the 3D distance exceeds 10 cm or 2D distance exceeds 12 px. Coordinate transforms, camera intrinsics, and pixel scales must agree with the data. & Inferred; paper \\
O11 & Coarse supervision must implement bidirectional overlap-aware circle loss with feature L2 distance, overlap weighting, gamma=40, and positive/negative margins 0.1/1.4. Numerically stable equivalent expressions are accepted. Similarity mean-squared error, constants, or a custom anti-collapse objective cannot substitute merely to produce nonzero loss. Where the paper's formula and prose are ambiguous, a self-consistent interpretation supported by the materials is accepted. & Inferred; paper \\
O12 & Fine supervision must implement bidirectional standard circle loss with gamma=24 and positive/negative margins 0.1/1.4, combined with coarse supervision using beta=1. Numerically stable rewriting is accepted; replacing the training objective is not. & Inferred; paper \\
O13 & A training batch with valid geometric supervision must propagate loss gradients to the trainable descriptors and matching heads actually used at inference, and update those parameters. Adding requires\_grad to an unrelated constant, bypassing an untrained matching head, or detaching the entire path does not meet this check. Individual batches without valid positives or negatives may be explicitly skipped; loss need not be positive at every step. & Inferred; research expectations \\
O14 & Remove the performance bottleneck blocking continued training. Measured full-step time must improve substantially under comparable hardware, input size, and training responsibilities, with evidence identifying the removed overhead. Omitting supervision, reducing task scope, or timing only a forward pass is insufficient. No fixed speed or improvement factor from later historical work is required. & Inferred; prior context \\
O15 & Resume repaired 7-Scenes training on the GPU, process successive traceable real pairs, update parameters, and retain progress from this run. Code alone, a launch command, or one random-tensor forward pass is insufficient. This checks training recovery; completing all 30 epochs or the RGB-D Scenes V2 main experiment is not required. & Inferred; prior context \\
O16 & The delivered checkpoint from this training must be compatible with the repaired model, restore learned parameters, and support continued training. Silently falling back to random initialization after a load failure is insufficient. Any appropriate state-saving format is accepted; no framework or internal filename is mandated. & Inferred; usability expectations \\
O17 & The repaired inference path must generate pixel--point matches from learned descriptors and use PnP-RANSAC to estimate the transform from point-cloud coordinates to the image camera frame. Ground-truth poses may supervise or evaluate the model, but must not replace inference estimates. & Inferred; paper \\
O18 & Compute registration metrics with the paper's meanings: IR is the fraction of correspondences with 3D distance below 0.05 m after the ground-truth transform; FMR is the fraction of pairs with IR above 0.1; RR is the fraction with predicted/ground-truth transform RMSE below 0.1 m. Keep failed pairs in denominators. Placeholders or validation loss alone are insufficient. Small-scale recovery validation is accepted but must not be called full paper-test-set reproduction. & Inferred; paper \\
O19 & Deliver this run's logs, checkpoint, validation-pair index, per-pair predictions and evaluation records, and reproduction commands. Records must link validation results to the current model and actual data. Any readable organization is accepted; checkpoints may be binary. & Explicit; Requirements \\
O20 & The repair report must describe completed training and validation stages consistently with delivered evidence, distinguishing program recovery, method consistency, and full main-experiment reproduction. Nonzero loss or completion of a few epochs alone does not establish the paper's registration results. & Inferred; research expectations \\
\end{longtable}
}

\clearpage

\section{Judge Prompt}

The evaluator prompt states the task requirements, the checklist item, the evidence available to inspect, and the binary decision rule. It requires a short evidence citation for every decision and forbids inferring an unobserved result from a claimed completion.

\section{Full Results and Hyperparameters}

The machine-readable release stores one record per task and rollout with the model identifier, score vector, delivery manifest status, wall-clock time, token counts, tool calls, and steps. Aggregate tables are generated by grouping these records by model, checklist dimension, and research stage. The run manifest is versioned with the task manifest so that a later model evaluation cannot silently alter the benchmark denominator.

\begin{table}[h]
\centering
\small
\begin{tabular}{ll}
\toprule
Setting & Value \\
\midrule
Actor harness & Codex for GPT-6.1-Sol, Claude Code for other models \\
Harness version & Codex 0.118.0 or Claude Code 2.1.283 on RJob, varying by run on E2B \\
Judge & GLM-5.3 (GLM-5.3-Flash for a small subset of runs), one call per checklist item \\
Execution backend & E2B for 920 CPU runs, RJob/Docker for the other 655 runs \\
Harbor version & 0.23.0 in the 650 readable run protocols \\
Rollouts per task & 3 \\
Time limit & 1.5 $\times$ task timeout for 920 CPU runs, task-specific overrides otherwise \\
Infrastructure retry & up to 2, actor timeout not retried \\
\bottomrule
\end{tabular}
\caption{Run configuration.}
\label{tab:config}
\end{table}
\begin{table}[h]
\centering
\small
\begin{tabular}{lr}
\toprule
Metric & Value \\
\midrule
Tasks & 75 \\
Checklist items & 1,173 \\
Avg.\ checklist items per task & 15.6 \\
Research stages covered & 7 \\
Avg.\ earlier requests in Background & 10.3 \\
Avg.\ workspace files at cut point & 41.1 \\
Avg.\ deliverables per task & 2.5 \\
Avg.\ instruction length (characters) & 1,677 \\
\bottomrule
\end{tabular}

\caption{Overall statistics of ARE-bench.}
\label{tab:stats}
\end{table}

\bibliographystyle{acl_natbib}
\bibliography{custom}

@article{jimenez-etal-2023-swebench,
  author = {Carlos E. Jimenez and John Yang and Alexander Wettig and Shunyu Yao and Kexin Pei and Ofir Press and Karthik Narasimhan},
  title = {{SWE-bench}: Can Language Models Resolve Real-World {GitHub} Issues?},
  journal = {arXiv preprint arXiv:2310.06770},
  year = {2023},
  url = {https://arxiv.org/abs/2310.06770}
}

@article{chan-etal-2024-mlebench,
  author = {Jun Shern Chan and Neil Chowdhury and Oliver Jaffe and James Aung and Dane Sherburn and Evan Mays and Giulio Starace and Kevin Liu and Leon Maksin and Tejal Patwardhan and Lilian Weng and Aleksander M{\k{a}}dry},
  title = {{MLE-bench}: Evaluating Machine Learning Agents on Machine Learning Engineering},
  journal = {arXiv preprint arXiv:2410.07095},
  year = {2024},
  url = {https://arxiv.org/abs/2410.07095}
}

@article{starace-etal-2025-paperbench,
  author = {Giulio Starace and Oliver Jaffe and Dane Sherburn and James Aung and Jun Shern Chan and Leon Maksin and Rachel Dias and Evan Mays and Benjamin Kinsella and Wyatt Thompson and Johannes Heidecke and Amelia Glaese and Tejal Patwardhan},
  title = {{PaperBench}: Evaluating {AI}'s Ability to Replicate {AI} Research},
  journal = {arXiv preprint arXiv:2504.01848},
  year = {2025},
  url = {https://arxiv.org/abs/2504.01848}
}

@article{wijk-etal-2024-rebench,
  author = {Hjalmar Wijk and Tao Lin and Joel Becker and Sami Jawhar and Neev Parikh and Thomas Broadley and Lawrence Chan and Michael Chen and Josh Clymer and Jai Dhyani and Elena Ericheva and Katharyn Garcia and Brian Goodrich and Nikola Jurkovic and Holden Karnofsky and Megan Kinniment and Aron Lajko and Seraphina Nix and Lucas Sato and William Saunders and Maksym Taran and Ben West and Elizabeth Barnes},
  title = {{RE-Bench}: Evaluating frontier {AI} {R\&D} capabilities of language model agents against human experts},
  journal = {arXiv preprint arXiv:2411.15114},
  year = {2024},
  url = {https://arxiv.org/abs/2411.15114}
}

@article{siegel-etal-2024-corebench,
  author = {Zachary S. Siegel and Sayash Kapoor and Nitya Nadgir and Benedikt Stroebl and Arvind Narayanan},
  title = {{CORE-Bench}: Fostering the Credibility of Published Research Through a Computational Reproducibility Agent Benchmark},
  journal = {arXiv preprint arXiv:2409.11363},
  year = {2024},
  url = {https://arxiv.org/abs/2409.11363}
}

@article{chen-etal-2024-scienceagentbench,
  author = {Ziru Chen and Shijie Chen and Yuting Ning and Qianheng Zhang and Boshi Wang and Botao Yu and Yifei Li and Zeyi Liao and Chen Wei and Zitong Lu and Vishal Dey and Mingyi Xue and Frazier N. Baker and Benjamin Burns and Daniel Adu-Ampratwum and Xuhui Huang and Xia Ning and Song Gao and Yu Su and Huan Sun},
  title = {{ScienceAgentBench}: Toward Rigorous Assessment of Language Agents for Data-Driven Scientific Discovery},
  journal = {arXiv preprint arXiv:2410.05080},
  year = {2024},
  url = {https://arxiv.org/abs/2410.05080}
}

@article{merrill-etal-2026-terminalbench,
  author = {Mike A. Merrill and Alexander G. Shaw and Nicholas Carlini and Boxuan Li and Harsh Raj and Ivan Bercovich and Lin Shi and Jeong Yeon Shin and Thomas Walshe and E. Kelly Buchanan and Junhong Shen and Guanghao Ye and Haowei Lin and Jason Poulos and Maoyu Wang and Marianna Nezhurina and Jenia Jitsev and Di Lu and Orfeas Menis Mastromichalakis and Zhiwei Xu and Zizhao Chen and Yue Liu and Robert Zhang and Leon Liangyu Chen and Anurag Kashyap and Jan-Lucas Uslu and Jeffrey Li and Jianbo Wu and Minghao Yan and Song Bian and Vedang Sharma and Ke Sun and Steven Dillmann and Akshay Anand and Andrew Lanpouthakoun and Bardia Koopah and Changran Hu and Etash Guha and Gabriel H. S. Dreiman and Jiacheng Zhu and Karl Krauth and Li Zhong and Niklas Muennighoff and Robert Amanfu and Shangyin Tan and Shreyas Pimpalgaonkar and Tushar Aggarwal and Xiangning Lin and Xin Lan and Xuandong Zhao and Yiqing Liang and Yuanli Wang and Zilong Wang and Changzhi Zhou and David Heineman and Hange Liu and Harsh Trivedi and John Yang and Junhong Lin and Manish Shetty and Michael Yang and Nabil Omi and Negin Raoof and Shanda Li and Terry Yue Zhuo and Wuwei Lin and Yiwei Dai and Yuxin Wang and Wenhao Chai and Shang Zhou and Dariush Wahdany and Ziyu She and Jiaming Hu and Zhikang Dong and Yuxuan Zhu and Sasha Cui and Ahson Saiyed and Arinbj{\"o}rn Kolbeinsson and Jesse Hu and Christopher Michael Rytting and Ryan Marten and Yixin Wang and Alex Dimakis and Andy Konwinski and Ludwig Schmidt},
  title = {{Terminal-Bench}: Benchmarking Agents on Hard, Realistic Tasks in Command Line Interfaces},
  journal = {arXiv preprint arXiv:2601.11868},
  year = {2026},
  url = {https://arxiv.org/abs/2601.11868}
}

@article{xie-etal-2024-osworld,
  author = {Tianbao Xie and Danyang Zhang and Jixuan Chen and Xiaochuan Li and Siheng Zhao and Ruisheng Cao and Toh Jing Hua and Zhoujun Cheng and Dongchan Shin and Fangyu Lei and Yitao Liu and Yiheng Xu and Shuyan Zhou and Silvio Savarese and Caiming Xiong and Victor Zhong and Tao Yu},
  title = {{OSWorld}: Benchmarking Multimodal Agents for Open-Ended Tasks in Real Computer Environments},
  journal = {arXiv preprint arXiv:2404.07972},
  year = {2024},
  url = {https://arxiv.org/abs/2404.07972}
}

@article{yao-etal-2024-taubench,
  author = {Shunyu Yao and Noah Shinn and Pedram Razavi and Karthik Narasimhan},
  title = {{$\tau$}-bench: A Benchmark for Tool-Agent-User Interaction in Real-World Domains},
  journal = {arXiv preprint arXiv:2406.12045},
  year = {2024},
  url = {https://arxiv.org/abs/2406.12045}
}

@article{ding-etal-2026-octobench,
  author = {Deming Ding and Shichun Liu and Enhui Yang and Jiahang Lin and Ziying Chen and Shihan Dou and Honglin Guo and Weiyu Cheng and Pengyu Zhao and Chengjun Xiao and Qunhong Zeng and Qi Zhang and Xuanjing Huang and Qidi Xu and Tao Gui},
  title = {{OctoBench}: Benchmarking Scaffold-Aware Instruction Following in Repository-Grounded Agentic Coding},
  journal = {arXiv preprint arXiv:2601.10343},
  year = {2026},
  url = {https://arxiv.org/abs/2601.10343}
}

@article{slinko-etal-2026-taskflows,
  author = {Igor Slinko and Yaroslav Golubev and Sergey Titov},
  title = {{Coding-Agent} Benchmarks Should Match Their Users' Task Flows},
  journal = {arXiv preprint arXiv:2610.09633},
  year = {2026},
  url = {https://arxiv.org/abs/2610.09633}
}

@article{lv-etal-2026-realclawbench,
  author = {Zongwei Lv and Yaoming Li and Zhewen Tan and Yilun Yao and Yuxuan Tian and Lin Sun and Xiangzheng Zhang and Weihong Lin and Tong Yang and Guangxiang Zhao},
  title = {{RealClawBench}: Live {OpenClaw} Benchmarks from Real {Developer-Agent} Sessions},
  journal = {arXiv preprint arXiv:2606.03889},
  year = {2026},
  url = {https://arxiv.org/abs/2606.03889}
}

@article{jha-etal-2026-reap,
  author = {Smriti Jha and Matteo Paltenghi and Chandra Maddila and Vijayaraghavan Murali and Shubham Ugare and Satish Chandra},
  title = {{REAP}: Automatic Curation of Coding Agent Benchmarks from Interactive Production Usage},
  journal = {arXiv preprint arXiv:2604.01527},
  year = {2026},
  url = {https://arxiv.org/abs/2604.01527}
}

@article{wang-etal-2026-aarribench,
  author = {Jiayu Wang and Weijiang Lv and Bowen Fu and Jing Fu and Jiayi Song and Lingyu Zhang and Lanxuan Xue and Luodi Chen and Zepeng Xin and Kaiyu Li and Xiangyong Cao},
  title = {Act As a Real Researcher: A Suite of Benchmarks Evaluating Frontier {LLMs} and Agentic Harnesses in Research Lifecycle},
  journal = {arXiv preprint arXiv:2606.07462},
  year = {2026},
  url = {https://arxiv.org/abs/2606.07462}
}

@book{anderson-krathwohl-2001-taxonomy,
  editor = {Lorin W. Anderson and David R. Krathwohl},
  title = {A Taxonomy for Learning, Teaching, and Assessing: A Revision of {Bloom}'s Taxonomy of Educational Objectives},
  year = {2001},
  edition = {Complete},
  publisher = {Longman},
  address = {New York}
}

@article{krathwohl-2002-revision,
  author = {David R. Krathwohl},
  title = {A Revision of {Bloom}'s Taxonomy: An Overview},
  journal = {Theory Into Practice},
  year = {2002},
  volume = {41},
  number = {4},
  pages = {212--218},
  doi = {10.1207/s15430421tip4104_2}
}

@article{garikaparthi-etal-2026-researchgym,
  author = {Aniketh Garikaparthi and Manasi Patwardhan and Arman Cohan},
  title = {{ResearchGym}: Evaluating Language Model Agents on Real-World {AI} Research},
  journal = {arXiv preprint arXiv:2602.15112},
  year = {2026},
  url = {https://arxiv.org/abs/2602.15112v2}
}

@article{kon-etal-2025-expbench,
  author = {Patrick Tser Jern Kon and Jiachen Liu and Xinyi Zhu and Qiuyi Ding and Jingjia Peng and Jiarong Xing and Yibo Huang and Yiming Qiu and Jayanth Srinivasa and Myungjin Lee and Mosharaf Chowdhury and Matei Zaharia and Ang Chen},
  title = {{EXP-Bench}: Can {AI} Conduct {AI} Research Experiments?},
  journal = {arXiv preprint arXiv:2505.24785},
  year = {2025},
  url = {https://arxiv.org/abs/2505.24785v2}
}

@article{bragg-etal-2025-astabench,
  author = {Jonathan Bragg and Mike D'Arcy and Nishant Balepur and Dan Bareket and Bhavana Dalvi and Sergey Feldman and Dany Haddad and Jena D. Hwang and Peter Jansen and Varsha Kishore and Bodhisattwa Prasad Majumder and Aakanksha Naik and Sigal Rahamimov and Kyle Richardson and Amanpreet Singh and Harshit Surana and Aryeh Tiktinsky and Rosni Vasu and Guy Wiener and Chloe Anastasiades and Stefan Candra and Jason Dunkelberger and Dan Emery and Rob Evans and Malachi Hamada and Regan Huff and Rodney Kinney and Matt Latzke and Jaron Lochner and Ruben Lozano-Aguilera and Cecile Nguyen and Smita Rao and Amber Tanaka and Brooke Vlahos and Peter Clark and Doug Downey and Yoav Goldberg and Ashish Sabharwal and Daniel S. Weld},
  title = {{AstaBench}: Rigorous Benchmarking of {AI} Agents with a Scientific Research Suite},
  journal = {arXiv preprint arXiv:2510.21652},
  year = {2025},
  url = {https://arxiv.org/abs/2510.21652v2}
}

@article{xu-etal-2026-researchclawbench,
  author = {Xu, Wanghan and Li, Shuo and Ye, Tianlin and Cao, Qinglong and Chen, Yixin and Gao, Hengjian and Wang, Yiheng and Li, Qi and Li, Kun and Xu, Sheng and Chai, Shengdu and Yu, Fangchen and Zhao, Xiangyu and Zhao, Zhangrui and Ma, Weijie and Guo, Zijie and Wu, Koutian and Zhou, Haoyu and Yin, Haoxiang and Cheng, Lixue and Hu, Chaofan and Li, Haoxuan and Mi, Lu and Xie, Xuxuan and Zhou, Yifan and Chen, Ruizhe and Zhou, Zhiwang and Guo, Xingjian and Zhou, Yuhao and He, Xuming and Xu, Shengyuan and Gu, Xinyu and Wu, Jiamin and Liu, Mianxin and Song, Chunfeng and Ling, Fenghua and Zhou, Dongzhan and Tang, Shixiang and Li, Yuqiang and Su, Mao and Ye, Peng and Sun, Siqi and Wang, Bin and Yang, Xue and Yin, Zhenfei and Fu, Tianfan and Zhai, Guangtao and Ouyang, Wanli and Zhang, Bo and Bai, Lei and Zhang, Wenlong},
  title = {{ResearchClawBench}: A Benchmark for End-to-End Autonomous Scientific Research},
  journal = {arXiv preprint arXiv:2606.07591},
  year = {2026},
  url = {https://arxiv.org/abs/2606.07591v5}
}

@article{brueckner-etal-2026-kbench,
  author = {Aubrey M. Brueckner and Darshil Patel and Yuhuan He and Timothy Kassis},
  title = {{K-Bench}: Measuring Model Performance on Real Scientific Agent Requests},
  journal = {arXiv preprint arXiv:2608.21601v2},
  year = {2026},
  url = {https://arxiv.org/abs/2608.21601v2}
}

@article{wu-etal-2026-swetogether,
  author = {Yifan Wu and Zhuokai Zhao and Songlin Li and Ho Hin Lee and Jiacheng Zhu and Shirley Wu and Tianhe Yu and Serena Li and Lizhu Zhang and Xiangjun Fan and Shengzhi Li},
  title = {{SWE-Together}: Evaluating Coding Agents in Interactive User Sessions},
  journal = {arXiv preprint arXiv:2606.29957},
  year = {2026},
  url = {https://arxiv.org/abs/2606.29957v1}
}

@article{niu-etal-2026-dumatebench,
  author = {Zechun Niu and Yukun Zhao and Jiaxin Zhang and Xu Shen and Jinhua Si and Han Tian and Can Xu and Yunfan Song and Jiaxin Mao and Yansong Gao and Yuchen Li and Jianmin Wu and Lingyong Yan and Shuaiqiang Wang and Dawei Yin},
  title = {{DuMateBench}: Evaluating Autonomous Agents in Complex Real-World Workflows},
  journal = {arXiv preprint arXiv:2608.26546v1},
  year = {2026},
  url = {https://arxiv.org/abs/2608.26546v1}
}

@article{xu-etal-2026-longdsbench,
  author = {Kewei Xu and Xiaoben Lu and Shuofei Qiao and Zihan Ding and Haoming Xu and Lei Liang and Ningyu Zhang},
  title = {{LongDS-Bench}: On the Failure of Long-Horizon Agentic Data Analysis},
  journal = {arXiv preprint arXiv:2605.30434v3},
  year = {2026},
  url = {https://arxiv.org/abs/2605.30434v3}
}

@article{xu-etal-2026-swebscience,
  author = {Zhipeng Xu and Jiahao Lu and Yining Zheng and Yuxin Wang and Xipeng Qiu},
  title = {{SWE-bench Science}: Can Coding Agents Resolve Engineering Tasks in Science?},
  journal = {arXiv preprint arXiv:2608.19799},
  year = {2026},
  url = {https://arxiv.org/abs/2608.19799v2}
}

@article{bogin-etal-2024-super,
  author = {Ben Bogin and Kejuan Yang and Shashank Gupta and Kyle Richardson and Erin Bransom and Peter Clark and Ashish Sabharwal and Tushar Khot},
  title = {{SUPER}: Evaluating Agents on Setting Up and Executing Tasks from Research Repositories},
  journal = {arXiv preprint arXiv:2409.07440},
  year = {2024},
  url = {https://arxiv.org/abs/2409.07440v1}
}

@article{wang-etal-2026-researchenvbench,
  author = {Yubang Wang and Chenxi Zhang and Bowen Chen and Zezheng Huai and Zihao Dai and Xinchi Chen and Yuxin Wang and Yining Zheng and Jingjing Gong and Xipeng Qiu},
  title = {{ResearchEnvBench}: Benchmarking Agents on Environment Synthesis for Research Code Execution},
  journal = {arXiv preprint arXiv:2603.06739},
  year = {2026},
  url = {https://arxiv.org/abs/2603.06739v2}
}

@misc{qwen-2026-qwen38-model-card,
  author = {{Qwen Team}},
  title = {{Qwen3.8-Max} Model Card},
  year = {2026},
  howpublished = {\url{https://huggingface.co/Qwen/Qwen3.8-2.4T-A95B-FP8}},
  note = {Accessed November 6, 2026. PaperBench evaluated in BasicAgent Code-Dev mode}
}

\end{document}